\documentclass{article}

\usepackage{arxiv}

\usepackage[utf8]{inputenc} %
\usepackage[T1]{fontenc}    %
\usepackage[dvipsnames]{xcolor}
\definecolor{mycolor}{RGB}{14,3,123}
\usepackage[
    colorlinks=true,
    citecolor=blue,      %
    linkcolor=red,       %
    urlcolor=magenta,    %
    filecolor=cyan,      %
    menucolor=red,       %
    runcolor=filecolor,  %
    allcolors=mycolor      %
]{hyperref}
\usepackage{url}            %
\usepackage{booktabs}       %
\usepackage{amsfonts}       %
\usepackage{nicefrac}       %
\usepackage{microtype}      %
\usepackage{lipsum}         %
\usepackage{graphicx}
\usepackage{natbib}
\usepackage{doi}

\usepackage{lipsum}
\usepackage{amsmath}
\usepackage{amsthm}
\usepackage{amssymb}
\usepackage{cleveref}       %
\usepackage{graphicx}
\usepackage{epstopdf}
\usepackage{soul}
\usepackage{caption}
\usepackage{subcaption}
\usepackage{algorithm}
\usepackage{algorithmic}
\usepackage{enumitem}
\usepackage{comment}
\usepackage{bm}
\usepackage{etoc}
\usepackage{multirow}
\usepackage{tikz}
\usetikzlibrary{shapes,arrows}
\usepackage{wasysym}
\usepackage{mwe}

\newcommand{\EE}{\mathbb{E}}
\newcommand{\RR}{\mathbb{R}}
\newcommand{\CN}{\mathcal{N}}
\newcommand{\CF}{\mathcal{F}}

\newcommand{\CS}{\mathcal{S}}

\title{Bifidelity Karhunen--Lo\`{e}ve Expansion Surrogate with Active Learning for Random Fields}

\newif\ifuniqueAffiliation
\uniqueAffiliationtrue

\author{\href{https://orcid.org/0000-0003-3379-7514}{\includegraphics[scale=0.06]{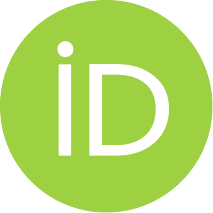}\hspace{1mm}Aniket Jivani}\\
	University of Southern California\\
	Los Angeles, CA 90089 \\
	\texttt{ajivani@usc.edu} \\
	\And
	\href{https://orcid.org/0000-0001-7219-7736}{\includegraphics[scale=0.06]{orcid.pdf}\hspace{1mm}Cosmin Safta} \\
    Sandia National Laboratories \\
    Livermore, CA 94550 \\
    \texttt{csafta@sandia.gov}
    \And
	\href{https://orcid.org/0000-0003-4062-4734}{\includegraphics[scale=0.06]{orcid.pdf}\hspace{1mm}Beckett Y. Zhou} \\
    Guggenheim School of Aerospace Engineering\\
	Georgia Institute of Technology\\
	Atlanta, GA 30332 \\
    \texttt{beckett.zhou@gatech.edu}
    \And
	\href{https://orcid.org/0000-0001-6544-2764}{\includegraphics[scale=0.06]{orcid.pdf}\hspace{1mm}Xun Huan} \\
    Department of Mechanical Engineering\\
	University of Michigan\\
	Ann Arbor, MI 48109 \\
    \texttt{xhuan@umich.edu}
}

\definecolor{mymagenta}{HTML}{F901FE}
\definecolor{matplotlibcyan}{RGB}{0,255,255}
\definecolor{tabcyan}{RGB}{23,190,207}
\def\norm#1{||#1||}

\newcommand{\CG}{\mathcal{G}}

\newcommand{\CI}{\mathcal{I}}

\newcommand{\CU}{\mathcal{U}}

\newcommand{\CX}{\mathcal{X}}

\newcommand{\NN}{\mathbb{N}}

\newcommand{\PP}{\mathbb{P}}

\newcommand{\argmax}{\operatornamewithlimits{arg\,max}}

\renewcommand{\shorttitle}{Bifidelity KLEs with Active Learning for Random Fields}

\hypersetup{
pdftitle={Bifidelity KLEs with Active Learning for Random Fields},
pdfauthor={Aniket~Jivani,Cosmin~Safta,Beckett Y.~Zhou,Xun~Huan},
pdfkeywords={multifidelity, uncertainty quantification, stochastic process, active learning, Gaussian process},
}

\begin{document}

\maketitle
\footnote{Address all correspondence to: Aniket Jivani, \href{mailto:ajivani@usc.edu}{\texttt{ajivani@usc.edu}}. This work was conducted while at the Department of Mechanical Engineering, University of Michigan, Ann Arbor, MI 48109.}
\begin{abstract}
We present a bifidelity Karhunen--Lo\`{e}ve expansion (KLE) surrogate model for field-valued quantities of interest (QoIs) under uncertain inputs. The QoIs considered here are scalar fields. The approach combines the spectral efficiency of the KLE with polynomial chaos expansions (PCEs) to preserve an explicit mapping between input uncertainties and output fields. By coupling inexpensive low-fidelity (LF) simulations that capture dominant response trends with a limited number of high-fidelity (HF) simulations that correct for systematic bias, the proposed method can enable accurate and computationally affordable surrogate construction.
To further improve surrogate accuracy, we develop an active learning strategy that adaptively selects new HF evaluations based on the surrogate's generalization error, estimated via cross-validation and modeled using Gaussian process regression. New HF samples are then acquired by maximizing an expected improvement criterion, targeting regions of high surrogate error.
The resulting BF-KLE-AL framework is demonstrated on three examples of increasing complexity: a one-dimensional analytical benchmark, a two-dimensional convection-diffusion system, and a three-dimensional turbulent round jet simulation based on Reynolds-averaged Navier--Stokes (RANS) and enhanced delayed detached-eddy simulations (EDDES). The experiments show that bifidelity gains depend on LF accuracy, discrepancy approximation, and the allocation of simulation cost. Active learning improves prediction over random sampling in several settings, while the cost-matched comparisons identify both favorable regimes and cases where an HF-only surrogate is more accurate.
\end{abstract}

\keywords{multifidelity, uncertainty quantification, stochastic process, active learning, Gaussian process}

\section{Introduction}
\label{s:introduction}

Computational models are now indispensable tools for studying complex physical phenomena and engineering systems. Such models often involve solving systems of governing equations (e.g., partial differential equations) that capture detailed physics at high spatial and temporal resolutions. When these models resolve the underlying processes with high accuracy, they are referred to as high-fidelity (HF) models. HF simulations can provide predictive accuracy suitable for decision-making and design; however, they are often prohibitively expensive for many-query tasks such as design optimization, inverse problems, or uncertainty quantification (UQ) under random inputs.

In contrast, low-fidelity (LF) models are computationally inexpensive and approximate the true system behavior, typically through simplifying assumptions or reduced numerical resolution. Relying solely on LF models can lead to systematic biases in the quantities of interest (QoIs), while exclusive reliance on HF models is computationally infeasible. Bifidelity modeling seeks a compromise by leveraging both: inexpensive LF evaluations to capture the bulk structure of the response and a limited number of HF simulations to correct the residual bias. 
This idea extends naturally to multifidelity hierarchies, but we focus here on the bifidelity case. 
Examples of multifidelity surrogate modeling based on linear and nonlinear surrogate representations include 
\cite{kennedy_predicting_2000,ng_multifidelity_2012,narayan_stochastic_2014,ghanem_multifidelity_2017,zhang_multifidelity_2018,howard_multifidelity_2023}
with broader surveys available in~\cite{peherstorfer_survey_2018,fernandez-godino_review_2023}.

In this work, we propose a bifidelity surrogate modeling framework for scalar-field quantities of interest (e.g., spatially and/or temporally varying outputs such as temperature fields or individual velocity components) based on the Karhunen--Lo\`{e}ve expansion (KLE) (see, e.g.,~\citep[Chapter 2]{LeMaitre2010}, ~\citep{betz_numerical_2014,safta_numerical_2026}). 
The KLE provides a spectral decomposition of stochastic processes into orthogonal spatial modes and random coefficients, allowing a compact representation of correlated fields. To explicitly represent inputs in terms of their physical sources of uncertainty, we couple the KLE with polynomial chaos expansions (PCEs) (see, e.g.,~\citep{Ghanem1991, Najm2009, Xiu2009, LeMaitre2010, Ernst2012, Mueller:2025}), which express the random coefficients as polynomial functions of independent random variables. The combined KLE-PCE structure represents the dependence of output fields on uncertain inputs and propagates their distribution across fidelity levels. The discrepancy surrogate corrects the modeled response, which can change both its mean and its variability. We note that reducing physical output variance is not the objective.

We then introduce an active learning (AL) framework~\cite{Cohn1996,Settles2012} that automatically identifies where new HF simulations should be conducted to best improve the surrogate. Specifically, we estimate the surrogate's generalization error through cross-validation and fit these errors using Gaussian process (GP) regression~\citep{Rasmussen2006,gramacy2020surrogates}. The GP provides a probabilistic prediction of the cross-validation error over the input space, from which new HF samples are selected by maximizing an expected improvement acquisition function~\citep{Jones1998,Mockus1974}. This approach enables data-efficient surrogate refinement in a greedy manner, but does not perform resource allocation between the HF and LF models that is done in variance-minimization approaches such as multi-level, multifidelity Monte Carlo and approximate control variate methods~\cite{Giles2008,geraci2015multifidelity,peherstorfer_optimal_2016,Gorodetsky2020, bomarito_optimization_2022}. AL also differs from data augmentation techniques used for bifidelity surrogates  \citep{sella_projection-based_2023}, which focus on efficient correction of trained LF surrogate predictions to generate additional synthetic HF evaluations. These are common in settings where growing a pre-existing simulation dataset is not feasible.

The resulting framework, BF-KLE-AL (Bifidelity Karhunen--Lo\`{e}ve Expansion with Active Learning), is demonstrated on three progressively complex examples:
(a) a one-dimensional (1D) analytical benchmark with known truth;
(b) a two-dimensional (2D) convection-diffusion problem; and
(c) a three-dimensional (3D) turbulent jet flow simulation using Reynolds-averaged Navier--Stokes and enhanced delayed detached-eddy simulation solvers.
Across these examples, we evaluate the framework's accuracy, efficiency, and capability for forward UQ.

Our main contributions are as follows.
\begin{itemize}
\item We design a new bifidelity surrogate model based on the combination of KLE and PCE for representing spatially or temporally distributed QoIs.

\item We develop an active learning algorithm that iteratively refines the bifidelity surrogate by selecting informative HF samples, guided by cross-validated error modeling through GP regression.

\item We evaluate predictive accuracy and simulation cost on synthetic benchmarks and turbulent flow simulations, identifying both favorable bifidelity regimes and cases where single-fidelity construction is more accurate.
\end{itemize}
A public implementation and supporting materials are available at \url{https://github.com/aniketjivani/KLE_UQ_Final}.

The paper is organized as follows. 
Section \ref{s:bgrd} reviews the theoretical background on KLEs and PCEs. Section \ref{s:methods} presents the proposed bifidelity KLE formulation and active learning strategy. Section \ref{s:results} reports numerical experiments and analyses on the benchmark and jet-flow test cases. Section \ref{s:conclusions} concludes with key findings and discusses directions for future work.

\section{Background}
\label{s:bgrd}

\subsection{Karhunen\textendash Lo\`{e}ve Expansions}
\label{s:KLEs}

Consider a probability space $(\Omega,\CF,\PP)$, where $\Omega$ is the sample space,
$\CF$ is a $\sigma$-algebra,
and $\PP$ is a probability measure.
Let $\theta : \Omega \to \RR^{n_{\theta}}$ be a random vector representing all uncertain input parameters of a model.
Suppose the model output is a real-valued stochastic process $y : \CX \times \RR^{n_\theta} \to \RR$ indexed by ${x}\in\CX \subseteq \RR^{n}$ (e.g., spatial or temporal location). For each input realization $\theta$, the field-valued QoI is the function $y(\cdot,\theta)$, with a scalar value at each location.
Then,
\begin{align}
y(x,\theta) = \mu(x) + y_0(x,\theta), \label{e:centering}
\end{align}
where $\mu(x)=\EE[y(x,\cdot)]$ is the mean function, and $y_0(x,\theta)$ is the zero-mean component after centering. 
In practice, we estimate and subtract the empirical mean from data when modeling the process, and add it back when making predictions. All theoretical development below pertains to the centered process $y_0$, which we focus on without loss of generality.

If $\CX$ is bounded and $y_0$ is square-integrable 
and mean-square continuous, it admits a spectral expansion known as the KLE (see, e.g.,~\citep[Chapter 2]{LeMaitre2010}):
\begin{align}\label{e:inf_sum_kle}
  y_0(x, \theta) = \sum_{k=1}^{\infty} \sqrt{\lambda_k}
  q_k(x) \zeta_k(\theta),
\end{align}
where $\zeta_k(\theta)$ are mutually uncorrelated random variables with zero mean and unit variance:
\begin{align}
    \EE[\zeta_k] = 0, \qquad \EE[\zeta_k \zeta_l] = \delta_{kl}.
\end{align}
The eigenvalues $\lambda_1\geq \lambda_2 \geq \ldots \rightarrow 0$ and orthonormal eigenfunctions $q_k(x)$ are obtained by solving the homogeneous Fredholm equation of the second kind:
\begin{align}
  \int_{\mathcal{X}}C(x,x') q_k(x')
  \,\mathrm{d}x' = \lambda_k q_k(x),\label{e:eigenvalue}
\end{align}
with 
\begin{align}
C(x,x') =\mathbb{E}\left[y_0(x, \cdot)
y_0(x', \cdot)\right]
\end{align}
as the covariance function.
The modal coefficients $\zeta_k(\theta)$ are obtained by projection, leveraging the orthonormality of $q_k(x)$:
\begin{align}
\zeta_k(\theta) = \frac{1}{\sqrt{\lambda_k}} \int_{\mathcal{X}}
  y_0(x,\theta) q_k(x)\,\mathrm{d}x.
  \label{e:zeta_exact}
\end{align}
The KLE is optimal in the mean-squared error sense:
truncating after $k_t$ terms yields the best possible rank-$k_t$ approximation.

\subsubsection{Numerical Implementation}

To compute a KLE numerically, we start with a training set $\{y(x_j,\theta^{(n)})\}_{j=1,\ldots,n_g;\,n=1,\ldots,N}$ of model simulations,
where $n=1,\ldots,N$ indexes the sampled parameter realizations, and $j=1,\ldots,n_g$ indexes spatial or temporal grid points. Thus the observations are indexed by all pairs $(j,n)$ in the Cartesian product of the grid and sample index sets. We assume a common grid across all samples (or otherwise interpolate onto a common grid). 

First, we estimate the mean from the training set:
\begin{align}
    \tilde{\mu}(x_j)  =  
    \frac{1}{N} \sum_{n=1}^{N}
  y(x_j,\theta^{(n)}),
\end{align}
and center the dataset to obtain $y_0(x_j,\theta^{(n)})=y(x_j,\theta^{(n)})-\tilde{\mu}({x_j})$.
The covariance is empirically estimated using a Monte Carlo approximation:
\begin{align}
C(x_i,x_j) \approx  C_{ij} = \frac{1}{N-1} \sum_{n=1}^{N}
  y_0(x_i,\theta^{(n)})y_0(x_j,\theta^{(n)}).
  \label{eq: K_approx}
\end{align}

To accommodate non-uniform grids, we introduce integration weights $w_j>0$ reflecting the local volume or area associated with grid point $x_j$. 
Based on the Nystr\"{o}m quadrature scheme \citep{betz_numerical_2014}, the discrete eigenproblem becomes:
\begin{align}
  \sum_{j=1}^{n_{g}} w_j C_{ij} q_{k,j} = \lambda_k q_{k,i}.
  \label{e:discrete_eigenvalue}
\end{align}
Letting $S=[y_0(x, \theta^{(1)}), \cdots, y_0(x, \theta^{(N)})]$ be the $n_g \times N$ snapshot matrix of centered data, we have $C=\frac{1}{N - 1} SS^{\top}$.
In matrix form, \eqref{e:discrete_eigenvalue} becomes:
\begin{align}
  (CW)Q = Q\Lambda,\label{e:eig_matrix}
\end{align}
where 
$W=\mathrm{diag}(w_1,\ldots,w_{n_g})$, $Q$ contains the right eigenvectors, and $\Lambda$ contains the eigenvalues. 
Empirical centering implies $S\mathbf{1}_N=0$, where $\mathbf{1}_N$ is the vector of $N$ ones. Hence $\operatorname{rank}(S)\leq\min(n_g,N-1)$, and the covariance has at most this many positive eigenvalues.

To reduce memory and computational cost when $n_g$ is large, we perform singular value decomposition (SVD) on
\begin{align}
B = \frac{1}{\sqrt{N-1}}W^{\frac{1}{2}}S,
\end{align}
yielding $B=U \Sigma V^\top$.
Since $BB^\top=W^{1/2}CW^{1/2}$, the squared singular values in $\Sigma^2$ are the eigenvalues of this symmetric weighted covariance matrix and, equivalently, of $CW$. The corresponding right eigenvectors of $CW$ are $Q=W^{-1/2}U$, with $Q^\top WQ=I$ for the retained modes. This recovers the quadrature-discretized covariance eigenproblem without forming the full covariance matrix.  
This way, only the $n_g\times N$ matrix $B$ needs to be formed instead of the $n_g \times n_g$ matrix $C$. 

We truncate the expansion 
to the first $k_t$ terms:
\begin{align}
  y_0(x, \theta) \approx \tilde{y}_0(x, \theta) = \sum_{k=1}^{k_t} \sqrt{\lambda_k}
  q_k(x) \zeta_k(\theta).\label{e:KLE_truncate}
\end{align}
The truncation level $k_t$  is typically selected based on the decay of $\lambda_{k}$; for example, one may truncate when $\lambda_{k}/\lambda_{1}\leq 0.1$. Alternative criteria include cumulative energy thresholds (e.g., retaining 99\% of total variance) or cross-validation-based model selection.

Realizations of the KLE modes' coefficients are computed as:
\begin{align}
  \zeta_k^{(n)} = \frac{1}{\sqrt{\lambda_k}}
  \sum_{j=1}^{n_{g}} w_j y_0(x_j,\theta^{(n)}) q_{k,j}.
  \label{e:zeta_samples}
\end{align}
One may model each $\zeta_k$ from these samples using a Gaussian approximation or kernel density estimation. The KLE construction yields uncorrelated coefficients that may still be statistically dependent. Sampling each fitted marginal distribution independently imposes the additional assumption that the coefficients are mutually independent, as holds for jointly Gaussian KLE coefficients.
However, this severs the explicit dependence on $\theta$, making it difficult to condition on a specific parameter value or to merge KLEs across fidelity levels.
To overcome this, we will introduce polynomial chaos expansion (PCE) to construct a surrogate mapping from $\theta$ to $\zeta_k$, enabling conditional inference and multifidelity fusion.

\subsection{Polynomial Chaos Expansions}
\label{ss: PCE}

Each modal coefficient $\zeta_k$ is a scalar random variable induced by uncertainty in the model parameters $\theta$. We treat $\zeta_k$ as a deterministic function of $\theta$, thereby preserving its dependence on the original physical sources of uncertainty. In other words, instead of modeling $\zeta_k$ as an unstructured random output, we seek to represent it explicitly as a function of $\theta$. One standard approach is to express each $\zeta_k$ using a truncated PCE (see, e.g.,~\citep{Ghanem1991, Najm2009, Xiu2009, LeMaitre2010, Ernst2012}):
\begin{align}
  \zeta_k(\theta) \mathrel{\approx} \tilde{\zeta}_k(\theta) = \sum_{\beta \in \CI} b_{k,\beta} \Psi_{\beta}(\xi_1(\theta),
  \ldots, \xi_{n_s}(\theta)),\label{e:zeta_PCE}
\end{align}
where $b_{k,\beta}$ are the expansion coefficients,
$\beta=(\beta_1,\ldots,\beta_{n_s})$, with $\beta_j\in\NN_0$ for all $j$, is a multi-index, $\CI$ is a finite index set, $n_s$ is the number of stochastic input dimensions, $\xi_i$ are reference random variables, and $\Psi_{\beta}(\xi_1, \ldots, \xi_{n_s})$ are multivariate orthonormal basis polynomials.
Each multivariate basis function $\Psi_{\beta}$ is expressed as a product of univariate polynomials:
\begin{align}
    \Psi_{\beta}(\xi_1(\theta), \ldots, \xi_{n_s}(\theta)) = \prod_{i=1}^{n_s}
    \psi_{\beta_i}(\xi_i(\theta)),
\end{align}
where $\psi_{\beta_i}$ denotes the degree-$\beta_i$ univariate orthonormal polynomial corresponding to the distribution of $\xi_i$.

\subsubsection{Numerical Implementation}

In our implementation, we adopt several common design choices.
\begin{itemize}
\item We set $n_s$ equal to the number of uncertain input parameters $\theta$ and associate each $\xi_i$ with the $i$th $\theta$ dimension. 
This provides a natural indexing and simplifies interpretation~\citep{Ernst2012}.

\item For our applications, where each uncertain input $\theta_i$ follows a uniform distribution, we use Legendre polynomials with $\xi_i\sim\CU(-1,1)$. 
The transformation from each physical parameter $\theta_i$ to the reference variable $\xi_i$ is then a simple affine scaling. 
Other options of $\xi_i$ and $\psi_{\beta_i}$ are also possible~\citep{Xiu2002}; for example, Gaussian-distributed inputs naturally pair with Hermite polynomials.

\item We adopt a total-order expansion of polynomial degree $p$, yielding the index set $\CI=\{\beta : \norm{\beta}_{1} \leq p \}$, with total number of terms $n_t = (n_s+p)!\big/(n_s!p!)$. 
The convergence rate depends on the regularity of $\zeta_k(\theta)$; smoother functions admit more rapid decay in expansion coefficients and hence require fewer terms.

\item For simplicity, we use the same basis for all $\zeta_k$, though it is possible to vary the basis or polynomial degree across modes.
\end{itemize}

Given the $N$ realizations of each $\zeta_k$, we solve for the coefficients $b_{k,\beta}$ using non-intrusive regression:
\begin{align}
  \underbrace{\begin{bmatrix} \Psi_{\beta^{(1)}}(\xi(\theta^{(1)})) & \cdots &
      \Psi_{\beta^{(n_t)}}(\xi(\theta^{(1)})) \\ \vdots & & \vdots
      \\ \Psi_{\beta^{(1)}}(\xi(\theta^{(N)})) & \cdots &
      \Psi_{\beta^{(n_t)}}(\xi(\theta^{(N)})) \end{bmatrix}}_{A_k}
  \underbrace{\begin{bmatrix} b_{k,\beta^{(1)}} \\ \vdots
      \\ b_{k,\beta^{(n_t)}} \end{bmatrix}}_{b_k} \mathrel{\approx}
  \underbrace{\begin{bmatrix} \zeta_k^{(1)} \\ \vdots
      \\ \zeta_k^{(N)} \end{bmatrix}}_{c_k},\label{e:sparse_regression}
\end{align}
where $\beta^{(j)}$ denotes the $j$th basis term following a particular ordering in the set $\mathcal{I}$,
$A_k \in \RR^{N\times n_t}$ is the feature matrix,
$b_k \in \RR^{n_t}$ is the vector of unknown coefficients, and $c_k \in\RR^{N}$ contains the known values $\zeta_k^{(n)}$.

Since the number of available simulations $N$ can be small relative to the number of basis terms $n_t$, a suitable choice of regularization is useful to prevent overfitting. We use Tikhonov ($\ell_2$) regularization in all examples:
\begin{align}
     b_k^{\ast} = \arg \min_{b_k} \|A_k b_k - c_k\|_2^2 + \tau \|\Gamma b_k\|^2_2,
  \label{e:Tikhonov}
\end{align}
where $\tau \geq 0$ is a regularization parameter 
and $\Gamma$ is chosen to be the first-order differencing weight matrix. For the chosen basis ordering, this quadratic penalty discourages differences between adjacent coefficients. It promotes smoothness of the coefficient sequence without imposing sparsity. The parameter $\tau$ typically can be selected via cross-validation~\citep{Huan2018b}. 

This procedure enables fast evaluation of $\zeta_k(\theta)$ for new inputs $\theta$, thereby allowing efficient reconstruction of $y_0(x, \theta)$ via the truncated KLE.

\section{Methodology}
\label{s:methods}

In this section, we first describe the construction of the additive bifidelity KLE surrogate with the incorporation of PCEs, followed by the active learning procedure used to improve the surrogate model under budget constraints. 
These steps are then summarized in an overall algorithm for BF-KLE-AL.

\subsection{Bifidelity Karhunen\textendash Lo\`{e}ve Expansions}
\label{ss: MethodsMFUQ}

Suppose we have access to both an HF model and an LF model, each taking in the same input variables and predicting the same QoIs. Then we can write:
\begin{align}
    y_{\mathrm{HF}}({x}, \theta) = y_{\mathrm{LF}}({x}, \theta) + \lbrack y_{\mathrm{HF}}({x}, \theta)-y_{\mathrm{LF}}({x}, \theta)\rbrack .
\end{align}
We approximate the right-hand side using surrogates:
\begin{align}
{y}_{\mathrm{HF}}({x}, \theta) \approx \underbrace{\tilde{y}_{\mathrm{LF}}({x}, \theta)}_{\text{LF surrogate}} + \underbrace{\tilde{y}_{\mathrm{\Delta}}({x}, \theta)}_{\text{Discrepancy surrogate}},
\label{eq:hf_approx_lf_disc}
\end{align}
where 
$\tilde{y}_{\Delta}({x}, \theta)$ 
approximates the discrepancy between the HF and LF models.
This approach allows us to leverage the efficiency of the LF model while correcting for its bias using a small number of HF evaluations, yielding a bifidelity approximation of the HF model at reduced cost.
Expanding both terms into their respective mean and zero-mean components:
\begin{align}
{y}_{\mathrm{HF}}({x}, \theta)    \approx \lbrack \tilde{\mu}_{\mathrm{LF}}({x}) + \tilde{y}_{0,\mathrm{LF}}({x}, \theta)\rbrack +
    \lbrack 
    \tilde{\mu}_{\mathrm{\Delta}}({x}) + \tilde{y}_{0,\Delta}({x}, \theta)\rbrack,
\end{align}
we define the right-hand side to be the bifidelity surrogate model:
\begin{align}
    \tilde{y}_{\mathrm{BF}}({x}, \theta) = \tilde{\mu}_{\mathrm{BF}}({x})+
     \tilde{y}_{0,\mathrm{BF}}({x}, \theta), \label{e:MF_KLE_sum}
\end{align}
where $\tilde{\mu}_{\mathrm{BF}}({x}) = \lbrack \tilde{\mu}_{\mathrm{LF}}({x}) + \tilde{\mu}_{\mathrm{\Delta}}({x})\rbrack$ and $\tilde{y}_{0,\mathrm{BF}}({x}, \theta) = \lbrack 
     \tilde{y}_{0,\mathrm{LF}}({x}, \theta) + \tilde{y}_{0,\Delta}({x}, \theta)\rbrack$. 

To construct 
$\tilde{y}_{0,\Delta}({x}, \theta)$, we require paired HF and LF simulation evaluations at the {same} $\theta$ values, and ideally on a common spatial grid. If the HF and LF solvers use different meshes, interpolation or restriction operations may be needed to align them.
If $N_1$ LF runs construct $\tilde{y}_{0,\mathrm{LF}}$ and $N_{\Delta}$ HF-LF pairs construct $\tilde{y}_{0,\Delta}$, LF evaluations shared by the two datasets are counted only once. Let $N_{\cap}$ denote their overlap. The total cost is $(N_1+N_{\Delta}-N_{\cap})c_{\mathrm{LF}}+N_{\Delta}c_{\mathrm{HF}}$, where $c_{\mathrm{LF}}$ and $c_{\mathrm{HF}}$ are per-evaluation costs. For a nested discrepancy set, $N_{\cap}=N_{\Delta}$ and no extra LF evaluations are needed at those pilot points. Subsequent acquisitions outside the existing LF set require both evaluations.

A possible extension for nonnested pre-existing data is to train a residual surrogate on $y_{\mathrm{HF}}(x,\theta)-\tilde{y}_{\mathrm{LF}}(x,\theta)$ at the available HF inputs, avoiding additional LF simulator evaluations there. This residual includes LF-surrogate error as well as the HF-LF model discrepancy. Prediction must use the same LF surrogate that defined the residual targets; changing that predictor requires recomputing the residuals and retraining their surrogate. This extension has not been tested here.

Substituting the truncated KLE and the PCE representations for the zero-mean components gives:
\begin{align}
    \tilde{y}_{0,\mathrm{BF}}({x}, \theta) &=\tilde{y}_{0,\mathrm{LF}}({x}, \theta) + \tilde{y}_{0,\Delta}({x}, \theta) \nonumber\\
    &=\sum_{k=1}^{k_{t}}\sqrt{\lambda_k}q_k({x})
    \left \lbrack \sum_{j=1}^{n_{t}}b_{k,{\beta^{(j)}}}\Psi_{\beta^{(j)}}(\xi(\theta))\right \rbrack \nonumber\\
    &\quad + \sum_{k'=1}^{k'_{t}}\sqrt{\lambda'_{k'}}q'_{k'}({x})
    \left \lbrack \sum_{j=1}^{n_{t}}b'_{k',{\beta^{(j)}}}\Psi_{\beta^{(j)}}(\xi(\theta))\right \rbrack \nonumber\\
    &= \sum_{j=1}^{n_{t}} \left \lbrack\sum_{k=1}^{k_{t}}\sqrt{\lambda_k}q_k({x})b_{k,{\beta^{(j)}}} + \sum_{k'=1}^{k'_{t}}\sqrt{\lambda'_{k'}}q'_{k'}({x})b'_{k',{\beta^{(j)}}} \right \rbrack\Psi_{\beta^{(j)}}(\xi(\theta)).\label{e:MF_KLE}
\end{align}
The coefficients $b_{k,\beta}$ and $b'_{k',\beta}$ are fitted separately for the LF and discrepancy expansions, respectively. Here, we have substituted the
PCEs for each modal coefficient $\zeta_k$ and $\zeta'_{k'}$, interchanged the order of summations, and 
combined the two components under the assumption that the same PCE basis is used for both LF and discrepancy terms. 
Finally, we obtain the full bifidelity prediction $\tilde{y}_{\mathrm{BF}}$ by adding $\tilde{\mu}_{\mathrm{BF}}$ back to $\tilde{y}_{0,\mathrm{BF}}$ via \eqref{e:MF_KLE_sum}.

We make a few observations about \eqref{e:MF_KLE}. First, although we assume a common set of PCE basis functions for all $\zeta_k$ and $\zeta'_{k'}$, this assumption can be relaxed. In more general settings, different polynomial families and degrees can be used for different components, and the resulting expansions can still be combined by taking the union of all basis terms. Second, the eigenvalues and eigenvectors associated with the LF and discrepancy terms are distinct: $(\lambda_k, q_k)$ from $\tilde{y}_{0,\mathrm{LF}}$ are not the same as $(\lambda'_{k'}, q'_{k'})$  from $\tilde{y}_{0,\Delta}$. Their eigenbases are not generally mutually orthogonal, so simply adding the two expansions does not produce a KLE of the HF field. A KLE of the resulting surrogate would require constructing and diagonalizing its covariance.
Rather, it represents a bifidelity surrogate: a global PCE of the bifidelity field, with spatially varying coefficients constructed from two independently built surrogates. 
In our numerical experiments, we compare this bifidelity model 
against a KLE built solely from HF data and, for simpler synthetic cases, against reference ensembles of true HF simulations, evaluating both its predictive accuracy and computational efficiency. 
Finally, the implementation requires a common spatial representation for HF and LF solution fields.

The above derivation assumes that both models share the same input parameterization. When this assumption does not hold, alternative strategies are needed. 
One possible approach is to map the distinct parameter spaces to a common latent space.
For example, PCEs can be constructed not only for outputs but also for inputs, allowing samples drawn in the latent space
to be mapped to both models' inputs and outputs. Multifidelity methods that address dissimilar parameterizations
can be found in \citet{geraci_2018_asd,geraci_leveraging_2018,gorodetsky_mfnets:_2020}.

\subsection{Active Learning for Bifidelity KLE}
\label{ss: al-bf-kle}

Having established the bifidelity KLE framework, we now 
focus on actively selecting new HF samples to improve the surrogate model. 
This is achieved through an active learning strategy~\citep{Settles2012} that iteratively refines the surrogate in regions where its predictions are expected to be least accurate.  
To quantify and guide improvement, we employ cross-validation to estimate the surrogate's generalization error.

At the initial pilot stage, the training dataset consists of $N_{\mathrm{LF}}^P$ LF samples used to construct the LF term, and a smaller subset of $N_{\Delta}^P$ paired HF-LF samples used to model the discrepancy term. 
The initial sample design for both levels can be generated using space-filling strategies (e.g., Latin hypercube or maximin designs) or pseudo-/quasi-random sequences.

We adopt a $k$-fold cross-validation procedure to estimate the local surrogate error. At each stage, the currently available paired samples are partitioned into $k$ folds; one fold provides validation points and the remaining folds provide discrepancy-training points. In the standard construction that updates both terms, held-out HF evaluations and their LF partners are excluded from fitting both surrogate terms, while other LF-only training points remain available. 
For each sample $\theta_i$, we compute the relative prediction error between the HF output and the surrogate prediction constructed when that sample is part of the held-out fold. Since the prediction QoIs are field-valued, this comparison yields an error field over the spatial or temporal domain, which is then summarized into a scalar error metric (relative $\ell_2$ error):
\begin{align}
    \varepsilon^{(i)} = \frac{\|y_{\mathrm{HF}}({x}, \theta_i) -\tilde{y}_{\mathrm{BF}}^{-\mathtt{s}(i)}({x}, \theta_i)\|_{2}}{\|y_{\mathrm{HF}}({x}, \theta_i)\|_{2}},
    \label{eq:kfold_err}
\end{align}
where $\mathtt{s}(i)$ denotes the fold index containing sample $\theta_i$, and $\tilde{y}_{\text{BF}}^{-\mathtt{s}(i)}$ is the surrogate trained without that fold's data.
The norm is the Euclidean norm of the discrete field vector on the common evaluation grid: $\|v\|_2=(\sum_{j=1}^{n_g}v_j^2)^{1/2}$. It weights grid entries equally, whereas the quadrature weights in the KLE eigenproblem approximate spatial integration. Relative errors require a nonzero reference norm.

The key idea is to use these cross-validation errors as proxies for the surrogate's generalization performance. Regions exhibiting large errors are prioritized for additional HF sampling to improve model accuracy. 
However, since cross-validation errors are only available at evaluated points, we model them using a GP~\citep{Rasmussen2006,gramacy2020surrogates} to infer a smooth approximation of the generalization error across the parameter space. 

We define a GP prior over the scalar-valued cross-validation error as
\begin{align}
{\varepsilon}
 \sim \mathcal{GP}( {m_0}(\cdot), \mathtt{k}(\cdot, \cdot)),
\label{e: gp1}
\end{align}
where $m_0(\cdot)=0$ is the mean function and $\mathtt{k}(\cdot,\cdot)$ is the covariance kernel. The kernel choice controls the GP's expressivity. Here we employ a stationary Mat\'{e}rn kernel with $\nu=5/2$ for moderate smoothness; its other hyperparameters (length scale and variance) are optimized through maximum likelihood estimation. While other kernels such as a squared-exponential kernel may also be used in GP-based active learning, the Mat\'{e}rn kernel can accommodate localized variation in the estimated error landscape through less restrictive smoothness assumptions.

At active learning stage $\ell$, let $\varepsilon_{\ell} = [\varepsilon^{(1)}_\ell, \ldots, \varepsilon^{(M)}_\ell ]^\top$ denote the $M$ cross-validation errors, with associated parameter samples
$\Theta_{\ell}=[\theta_{\ell}^{(1)}, \ldots, \theta_{\ell}^{(M)}]^\top$.
The GP posterior mean and variance at a new point $\theta^\ast$ are given by:
\begin{align}
m_{{\varepsilon}}(\theta^\ast) 
&= 
\mathtt{k}^{\top}\left(\Theta_{\ell},\theta^\ast\right)K\left(\Theta_{\ell}, \Theta_{\ell}\right)^{-1} \varepsilon_\ell, \label{e: gpr_eqns1}
\\
\sigma_{{\varepsilon}}^{2}(\theta^\ast)&= \mathtt{k}(\theta^\ast, \theta^\ast)-\mathtt{k}^{\top}\left(\Theta_{\ell},\theta^{\ast}\right)K\left(\Theta_{\ell}, \Theta_{\ell}\right)^{-1}\mathtt{k}\left(\Theta_{\ell},\theta^\ast\right).
\label{e: gpr_eqns2}
\end{align}
To determine the next HF sample, we maximize expected improvement (EI) in the modeled error~\citep{Jones1998,Mockus1974}. This criterion seeks locations where the error is expected to exceed the largest observed cross-validation error, balancing the GP mean and uncertainty. A high-error prediction identifies a candidate for surrogate refinement; it does not guarantee that retraining will reduce the integrated prediction error. For a GP, EI of a single point is defined as:
\begin{align}
    \textrm{EI}(\theta) = \EE[\max(\varepsilon - \varepsilon^{\ast}, 0)\; | \;\varepsilon \sim \CN(m_{\varepsilon}(\theta), \sigma_{\varepsilon}^2(\theta))],
    \label{eq:EI_defn}
\end{align}
where $\varepsilon^{\ast}$ denotes the largest observed value thus far. The EI 
objective admits an analytic expression in terms of the GP posterior mean and variance at a candidate point. Batch EI can also admit analytic formulas in special cases, while sampling approximations are common for larger batches. 

\begin{align} 
\theta_{\textrm{new}} &= \argmax_{\theta} \textrm{EI}(\theta) \nonumber\\
&= \argmax_{\theta} \left\{\left(m_{{\varepsilon}}(\theta)-\varepsilon^{\ast}\right) {\Phi}\left(\frac{m_{{\varepsilon}}(\theta)-\varepsilon^{\ast}}{\sigma_{{\varepsilon}}(\theta)}\right)
+\sigma_{{\varepsilon}}(\theta)\phi\left(\frac{m_{{\varepsilon}}(\theta)-\varepsilon^{\ast}}{\sigma_{\varepsilon}(\theta)}\right)\right\},\label{eq:al_ei_analytic}
\end{align}
where $\Phi(\cdot)$ and $\mathit{\phi}(\cdot)$ are the standard normal cumulative distribution and probability density functions, respectively. When the posterior variance is zero, EI is defined by its continuous limit $\max(m_{\varepsilon}(\theta)-\varepsilon^\ast,0)$. Maximizing downward EI, defined from a decrease in modeled error, targets a different objective; it is generally not equivalent to minimizing the upward EI used here.

Each time a new HF sample is selected, the LF model is also evaluated at the same input. These paired evaluations are immediately incorporated into the training set, and either the discrepancy term alone or both components of the bifidelity surrogate are retrained. Keeping the LF surrogate fixed is a reasonable simplification when the pilot LF dataset already provides an accurate representation. It saves LF-surrogate retraining; paired LF simulator evaluations are still required at new HF inputs unless existing evaluations or the untested residual extension in Section~\ref{ss: MethodsMFUQ} are used.
Cross-validation errors are recomputed at the available paired points, and the GP is fitted to these updated targets. Rebuilding the surrogate changes the error landscape even at previously evaluated points, so the GP posterior must account for the revised errors. The experiments stop at their prescribed sampling budgets. Neither EI maximization nor additional training data guarantees monotone improvement in held-out error, and the algorithm does not retrospectively discard a worsening stage. A separate validation set could support stopping or selection among saved surrogates, but that strategy would consume validation resources and is not evaluated here.

Examples 1a and 2 express the acquisition budget as a number of HF evaluations and neglect LF evaluation costs. Example 1b includes synthetic HF-LF cost ratios, and Example 3 uses simulation costs in CPU hours. These comparisons count reused LF evaluations only once.

Future extensions will also explore cost-aware acquisition strategies that explicitly balance performance gain against the relative costs of LF and HF evaluations, potentially enabling adaptive fidelity selection during learning.

Nonuniform input distributions and unbounded acquisition domains raise distinct questions. On a bounded domain, the PCE basis or reference-variable transformation, initial sampling design, and integrated error measure should reflect the input law. The EI formula itself does not require uniformly distributed inputs. However, maximizing error-based EI emphasizes locations with large predicted error, and it does not directly minimize error integrated against the input density. The experiments here use uniform input laws; performance under nonuniform laws has not been assessed.

Common off-the-shelf implementations of Bayesian optimization algorithms typically use bounded inputs. For unbounded inputs, the implementation used in this study would require an explicit acquisition-domain strategy. Possible extensions include a bounded region selected by probability mass with a tail-error assessment, or adaptive domain expansion \citep{pmlr-v51-shahriari16,berk_bayesian_2020}. Prior means and acquisition criteria can also be tailored to control search behavior on unbounded domains \citep{Rainforth2016}.

A further practical consideration concerns the number of HF samples acquired per stage. Sequential (single-point) acquisition may be inefficient in high-dimensional problems or when parallel computational resources are available. In such cases, batch acquisition offers a more scalable alternative.
While analytical formulations of batch EI ($q$-EI) exist for small batch sizes~\citep{ginsbourger_kriging_2010}, the optimization rapidly becomes intractable as batch size grows. Consequently, several approximate methods have been proposed~\citep{ginsbourger_kriging_2010,azimi_hybrid_2012,gonzalez_batch_2016,joy_batch_2020}. In this work, we adopt the ``Kriging Believer'' heuristic \citep{ginsbourger_kriging_2010} for batch selection. Once the first acquisition point is chosen, its predicted mean value is treated as a pseudo-observation, and the GP is refitted iteratively until a batch of $q$ points is identified. This strategy enables efficient parallel HF sampling while preserving the exploratory behavior of single-point acquisition.

\subsection{Overall Algorithm for BF-KLE-AL}\label{ss: kle_alg}

The complete procedure for constructing the bifidelity KLE surrogate with active learning-based selection of new HF evaluations is summarized in Algorithm~\ref{alg: KLEAL}. The algorithm uses an HF-count budget $B_{\mathrm{HF}}$ and a fixed positive batch size $q$. 
The simulation-cost budgets in the experiments additionally account for LF cost as described above. The bifidelity surrogate is initially built from pilot datasets (line~\ref{l: initialize bifidelity KLE}) and the relative $k$-fold cross-validation errors are computed. Lines~\ref{l: begin_while}--\ref{l: update_al_stage} implement the AL procedure for a new batch of $q$ points until the HF-evaluation budget $B_{\mathrm{HF}}$ permits no further complete batch: in particular, GP regression and EI maximization (lines~\ref{l: init_gp_prior}\textendash\ref{l: ei_maximization}), surrogate model reconstruction (line~\ref{l: rebuild_BF}), and cross-validation error update (line~\ref{l: update_k_fold}).

\begin{algorithm}[htbp]
\caption{Construction of bifidelity KLE surrogate with active learning (BF-KLE-AL).}\label{alg: KLEAL}
\setlength{\linewidth}{\hsize}
\begin{algorithmic}[1]
\STATE \textbf{Inputs:} HF and LF models $\{y_\mathrm{HF}, y_{\mathrm{LF}}\}$, 
batch size $q$ for active learning,
pilot designs $\Theta_{\mathrm{LF}} = \{\theta_i\}_{i=1}^{N_\mathrm{LF}^P}$ and $\Theta_{\Delta} = \{\theta_i\}_{i=1}^{N_{\Delta}^P}$, 
corresponding datasets $Y_{\mathrm{LF}} = y_{\mathrm{LF}}({x}, \Theta_{\mathrm{LF}})$, $Y_{\Delta} = y_{\mathrm{HF}}({x}, \Theta_{\Delta}) - y_{\mathrm{LF}}({x}, \Theta_{\Delta})$, maximum number of HF evaluations $B_{\mathrm{HF}}\geq N_\Delta^P$ (including the pilot)\;
\STATE \textbf{Output:} Final bifidelity KLE surrogate model\;
\vspace{0.5em}
\STATE Initialize active learning stage $\ell = 0$\;
\STATE Build initial bifidelity KLE surrogate using pilot $(\Theta_{\mathrm{LF}}, Y_{\mathrm{LF}})$ and $(\Theta_{\Delta}, Y_{\Delta})$ following Section~\ref{ss: MethodsMFUQ}\; \label{l: initialize bifidelity KLE}
\STATE Compute relative $k$-fold errors $\varepsilon_{\ell}$ at $\Theta_{\Delta}$ via \eqref{eq:kfold_err}; in the standard construction, exclude each held-out paired fold from both surrogate fits\;
\STATE 
$N_{\mathrm{HF}} \gets N_\Delta^P$\;
\WHILE{$N_{\mathrm{HF}}+q \leq B_{\mathrm{HF}}$} \label{l: begin_while}
\STATE Initialize GP prior for $\varepsilon_\ell$\; \label{l: init_gp_prior}
\STATE Establish GP posterior mean and variance using \eqref{e: gpr_eqns1}--\eqref{e: gpr_eqns2}\;
\STATE Maximize EI acquisition function in \eqref{eq:al_ei_analytic} to select the next point, or employ Kriging Believer to select the next batch of points: $\Theta_{\mathrm{new}}$ where $|\Theta_{\mathrm{new}}| = q$\; \label{l: ei_maximization}
\STATE 
$\Theta_{\Delta} \gets \; \Theta_{\Delta} \cup \Theta_{\mathrm{new}}$, optionally: 
$\Theta_{\mathrm{LF}} \gets \; \Theta_{\mathrm{LF}} \cup \Theta_{\mathrm{new}}$\;
\STATE $Y_{\Delta} \gets \begin{bmatrix} Y_{\Delta} & y_{\mathrm{HF}}({x}, \Theta_{\mathrm{new}}) - y_{\mathrm{LF}}({x}, \Theta_{\mathrm{new}})\end{bmatrix}$, optionally:  \par\hspace{\algorithmicindent}%
$Y_{\mathrm{LF}} \gets \begin{bmatrix}Y_{\mathrm{LF}} & y_{\mathrm{LF}}({x}, \Theta_{\mathrm{new}})\end{bmatrix}$
\;
\STATE $N_{\mathrm{HF}}\gets N_{\mathrm{HF}} + q$\;
\STATE Rebuild bifidelity KLE surrogate using  $(\Theta_{\Delta}, Y_{\Delta})$ and optionally $(\Theta_{\mathrm{LF}}, Y_{\mathrm{LF}})$ following Section~\ref{ss: MethodsMFUQ}\; \label{l: rebuild_BF}
\STATE Compute updated relative $k$-fold cross-validation errors $\varepsilon_{\ell}$ for all $\Theta_{{\Delta}}$ via \eqref{eq:kfold_err}\; \label{l: update_k_fold}
\STATE $\ell \gets \ell + 1$\; \label{l: update_al_stage}
\ENDWHILE
\end{algorithmic}
\end{algorithm}

\section{Numerical Experiments and Results}
\label{s:results}

In this section, we evaluate our bifidelity KLE surrogate framework and active learning strategy across three test problems of increasing complexity:
\begin{enumerate}
    \item \textbf{Example 1a and 1b: 1D pulse function.} A family of simple test cases with a known analytical form of the HF model, enabling direct comparison against ground truth and evaluation of bifidelity surrogate performance under a number of synthetic cost and correlation settings.

    \item \textbf{Example 2: 2D convection-diffusion.} A parametric PDE problem solved using different discretizations, with uncertainty in source strength and location.

    \item \textbf{Example 3: 3D turbulent jet flow.} QoIs extracted from computational fluid dynamics (CFD)
    simulations of flow around a turbulent round jet.
\end{enumerate}

\ref{app:deeponet_comparisons} reports additional comparisons of BF-KLE performance with a bifidelity DeepONet operator learning framework on an analytical 1D jump function benchmark from \cite{howard_multifidelity_2023}.

\subsection{Example 1a: 1D pulse function}
\label{ss: ex1-1d}

We begin with a benchmark example where the HF model is defined analytically as a product of a decaying exponential and a sine function:
\begin{align}
    y_{\textrm{HF}}({x}, \theta) = \exp(-a{x}) \sin(b{x}),
    \label{e:yhf_eqn}
\end{align}
where $\theta=[a,b]$ are uncertain input parameters. We adopt $a\sim\mathcal{U}(40,60)$ in all cases, and define two different LF models based on distinct approximations of the sine term. 
\begin{itemize}
\item \textbf{Case 1 (C1):} The sine function is approximated using a truncated Taylor series expansion:
\begin{align}
    y_{\textrm{LF},1}(x, \theta) = \exp (-ax) \left( bx - \frac{b^3 x^3}{3!} + \frac{b^5 x^5}{5!}\right),
    \label{e:ylf_case1_eqn}
\end{align}
with $b\sim\mathcal{U}(60,80)$ for both the HF and LF models and $x \in [0, 0.1]$. This approximation closely matches the true sine behavior for small values of $x$, but diverges significantly as $x$ increases.

\item \textbf{Case 2 (C2):} To induce smoother and more consistent correlation with the HF model, we use a modified empirical approximation inspired by \cite{stroethoff_bhaskaras_2014}:
\begin{align}
    y_{\textrm{LF},2}(x,\theta) = \exp(-ax)\left[\frac{3.5 bx\frac{180}{\pi}\left(180 - bx\frac{180}{\pi}\right)}{15000 - bx\frac{180}{\pi}\left(180 - bx\frac{180}{\pi}\right)}\right],
    \label{e:ylf_case2_eqn}
\end{align}
with $b\sim \mathcal{U}(30,50)$ for both the HF and LF models and $x \in [0, 0.1]$.
\end{itemize}

\begin{figure}[htbp]
    \centering
    \includegraphics[width=\linewidth]{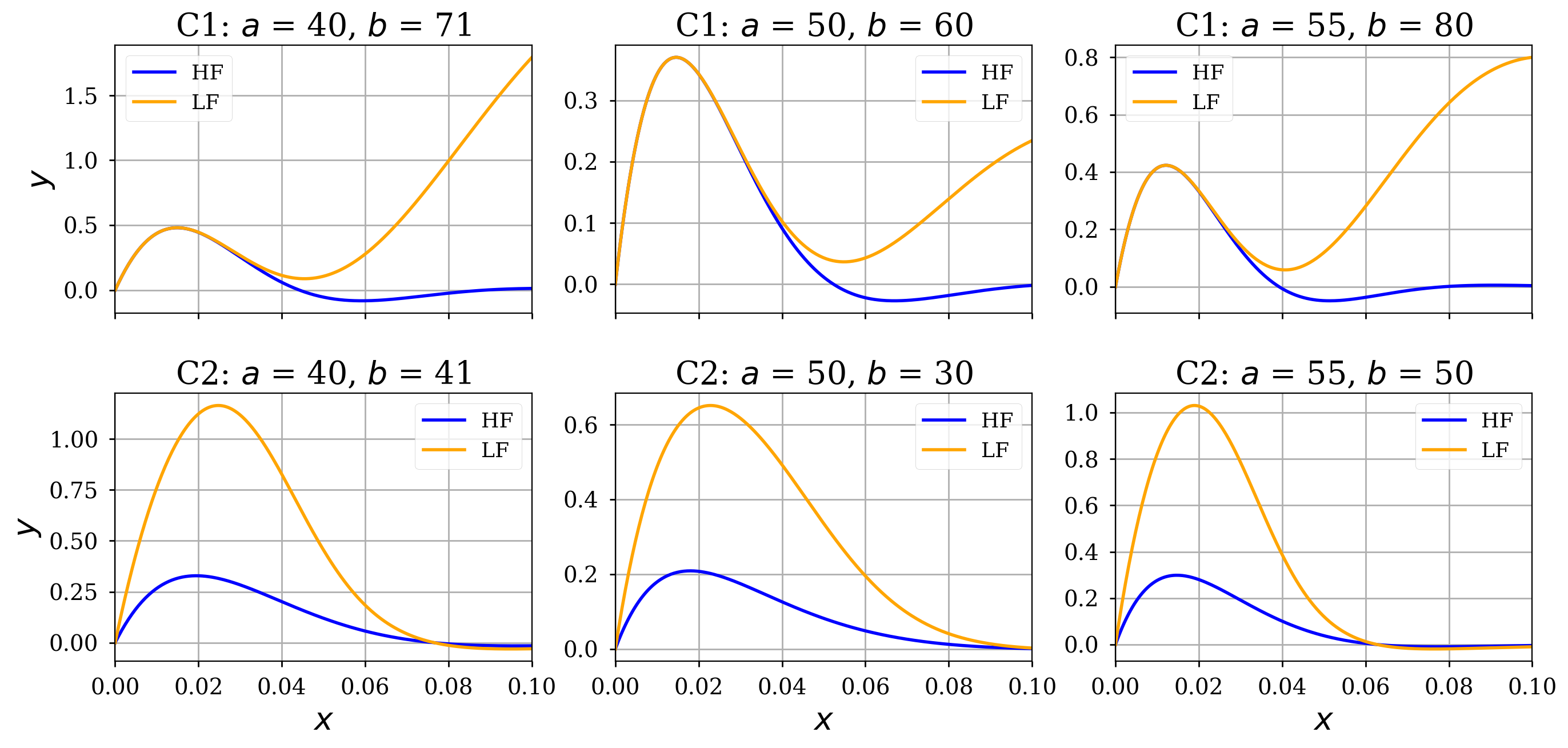}
    \caption{Evaluations for $y_{\textrm{LF}}$ and $y_{\textrm{HF}}$ at selected values of $a$ and $b$ for C1 (top row) and C2 (bottom row).}
    \label{fig:1d_samples_lf_hf_c1c2}
\end{figure}

Figure~\ref{fig:1d_samples_lf_hf_c1c2} shows representative realizations of the HF model alongside both LF models for selected values of $a$ and $b$. As expected, the C1 LF model yields good agreement with HF for small $x$, with visible divergence at larger values. In contrast, the C2 LF model exhibits greater absolute error.

\begin{figure}[htbp]
    \centering
    \includegraphics[width=\linewidth]{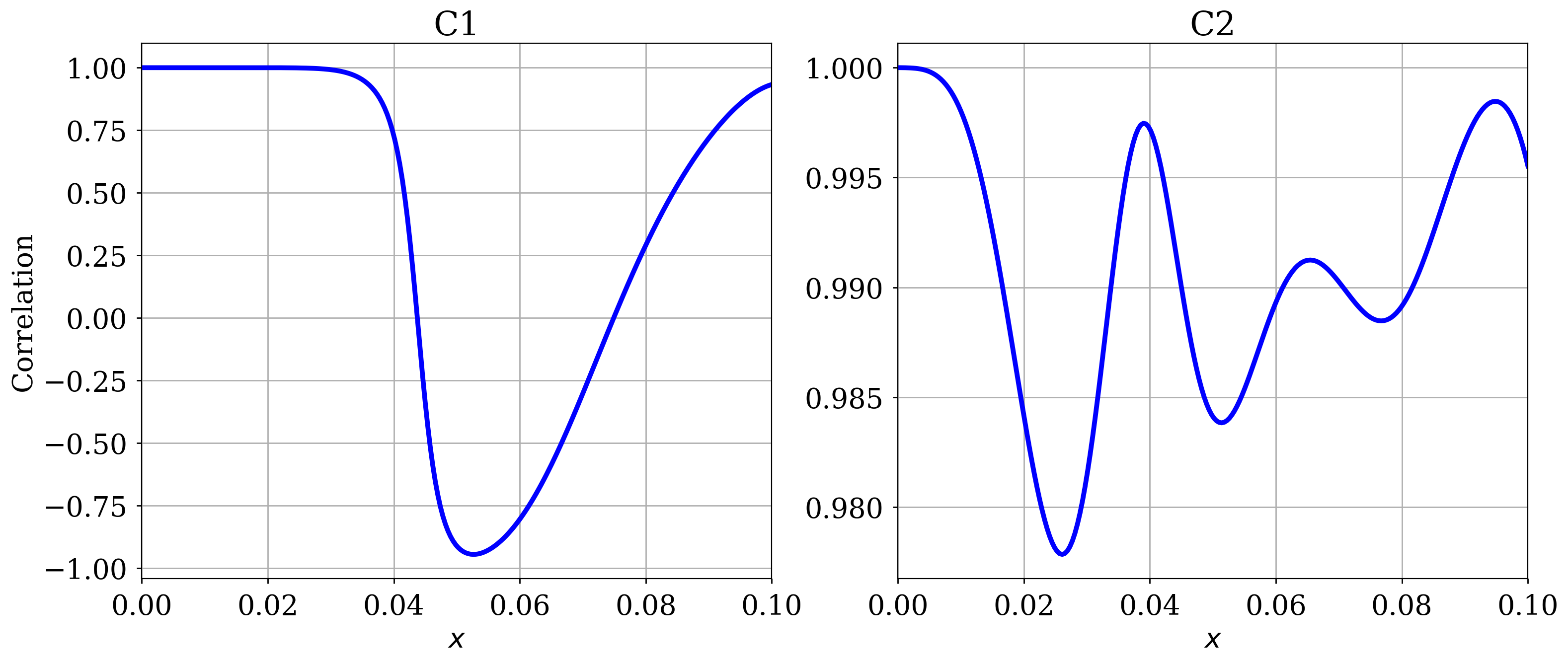}
    \caption{Pearson correlation coefficient between $y_{\textrm{HF}}$ and $y_{\textrm{LF}}$ over $x$ for C1 (left) and C2 (right), calculated using 1000 samples of $\theta$. C2 exhibits a more consistent correlation throughout the domain.}
    \label{fig: corr_1d_og}
\end{figure}

Figure~\ref{fig: corr_1d_og} presents the correlations between each LF model and the HF reference. The left panel illustrates that the C1 LF model maintains near-perfect positive correlation for small $x$, followed by a sharp decline around the midpoint and partial recovery near $x=0.1$.
On the other hand, the C2 LF model (right panel) shows stable correlation close to 1 throughout the domain.

\subsubsection{Bifidelity surrogate construction and performance}
\label{sss:ex1-1d}

\begin{figure}[htbp]
    \centering
    \includegraphics[width=0.7\linewidth]{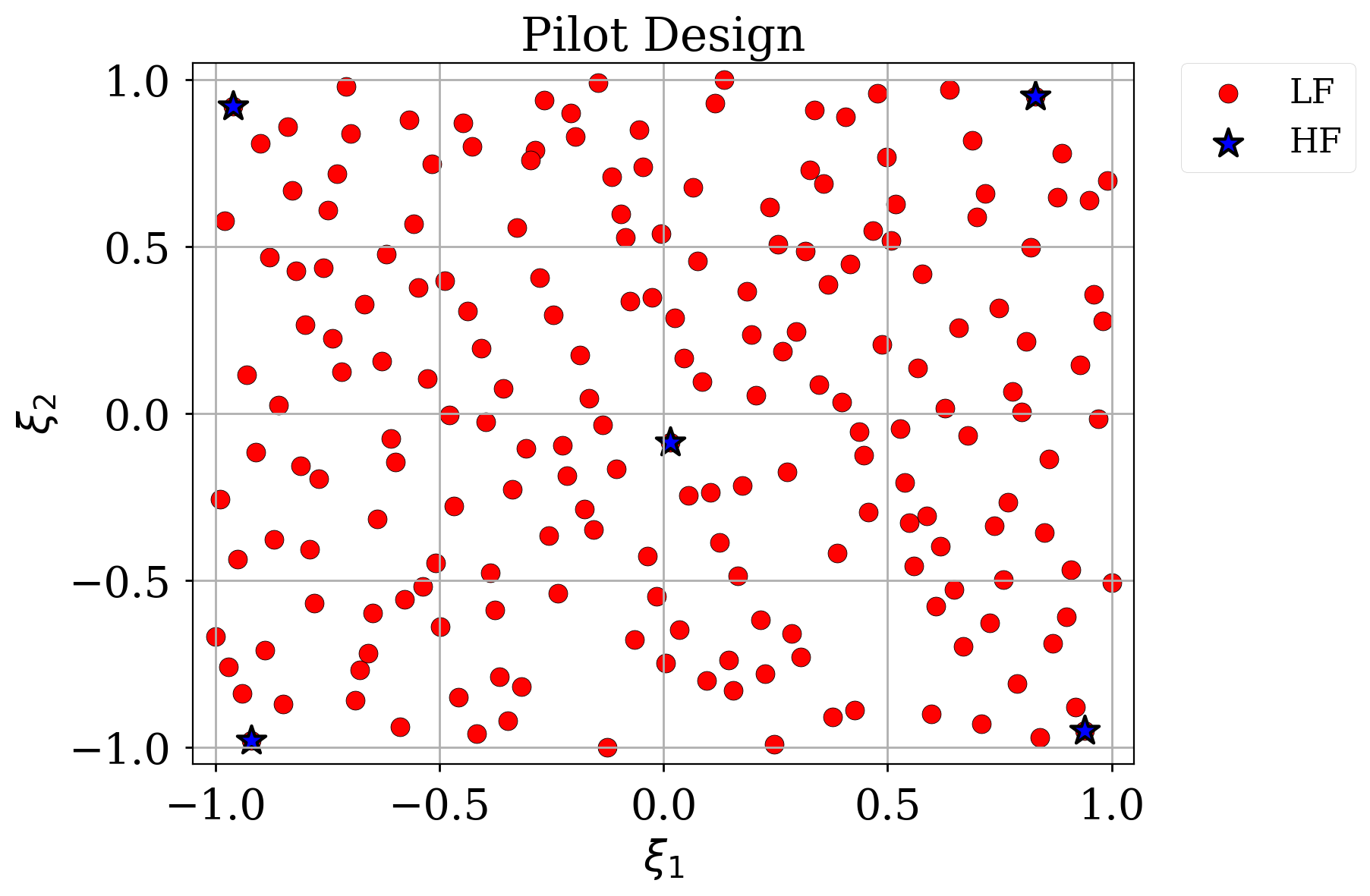}
    \caption{Pilot design for the 1D pulse problem with $(N_{\mathrm{LF}}^P, N_{\Delta}^P)=(200, 5)$. The design is mapped from the non-dimensional $(\xi_1,\xi_2)$ space to the corresponding $\theta$ parameters for C1 and C2 via affine scaling.}
    \label{fig:1d_lf_hf_pilot}
\end{figure}

We construct the initial bifidelity KLE surrogate using $N_{\mathrm{LF}}^P=200$ pilot LF simulations generated via a Latin hypercube (LH) space-filling design.
From these, a subset of $N_{\Delta}^P=5$ paired HF-LF samples, also selected to ensure good space-filling properties, are used as the pilot set for modeling the discrepancy field. The design sets $\Theta_{\mathrm{LF}}^P$ and $\Theta_{\Delta}^P$ in the non-dimensional space are shown in Figure~\ref{fig:1d_lf_hf_pilot}.

Each KLE is truncated to retain 99\% of the total variance, corresponding to approximately 1--3 modes, and the associated PCE employs a total-order expansion of degree 3. 
The computational budget for active learning is set to $B=65$ total HF evaluations. 
The GP model for generalization error is constructed using $k=5$-fold cross-validation, and the EI optimization is carried out in Python using routines from the BoTorch Bayesian optimization library~\citep{balandat2020botorch}. 
To account for stochasticity in $k$-fold partitioning and EI optimization, all experiments are repeated over 20 independent replicates. For comparison, in each replicate, an additional bifidelity KLE is constructed using the same total number of HF points selected via pseudo-random sampling; we call this BF-KLE-RS. In all examples, surrogate construction for BF-KLE-AL and BF-KLE-RS uses a common set of pilot samples prior to active learning or random sampling.

To assess the effectiveness of the proposed BF-KLE-AL strategy, we evaluate the surrogate's predictive accuracy against ground-truth HF solutions $y_{\mathrm{HF}}$ at multiple acquisition stages. Specifically, we compute an integrated relative error for each acquisition stage $\ell$:
\begin{align}
    \mu_{\varepsilon,\ell} = \int \frac{\|y_{\mathrm{HF}}({x}, \theta) -\tilde{y}_{\mathrm{BF}, \ell}({x}, \theta)\|_{2}}{\|y_{\mathrm{HF}}({x}, \theta)\|_{2}}\, p(\theta)  \, \mathrm{d}\theta,
    \label{eq:oracle_err_defn}
\end{align}
which we estimate by discretizing the integral using a $200 \times 200$ uniform grid over $\theta$. 

\begin{figure}[htbp]
    \centering
        \includegraphics[width=\linewidth]{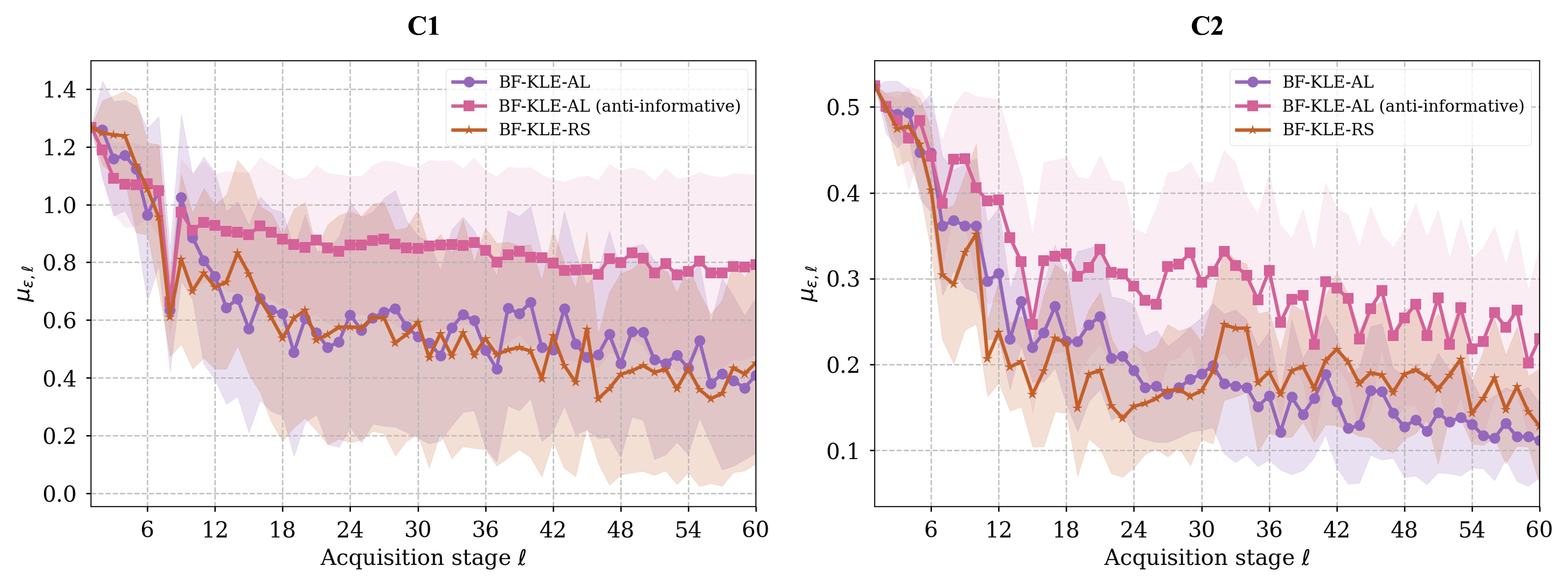}
    \caption{Comparison of integrated relative error for BF-KLE-AL, BF-KLE-RS, and anti-informative sampling using 60 additional HF points for C1 and C2. Results are averaged over 20 replicates.}
    \label{f: oracle_multiple_reps_LF_c1_c2_with_EI_50batch}
\end{figure}

Figure~\ref{f: oracle_multiple_reps_LF_c1_c2_with_EI_50batch} summarizes the average results over all 20 replicates. All methods share identical hyperparameter settings for KLE truncation, PCE mapping, and active learning parameters. In addition to BF--KLE--AL and BF--KLE--RS, we also test an intentionally anti-informative sampling policy that prioritizes low-error regions (i.e., minimizes EI). This provides a lower-bound reference (and sanity check) for the effect of poor sampling decisions. The distinction between minimizing upward EI and maximizing downward EI is discussed in Section~\ref{ss: al-bf-kle}.

\begin{figure}[htbp]
    \centering
    \includegraphics[width=\textwidth]{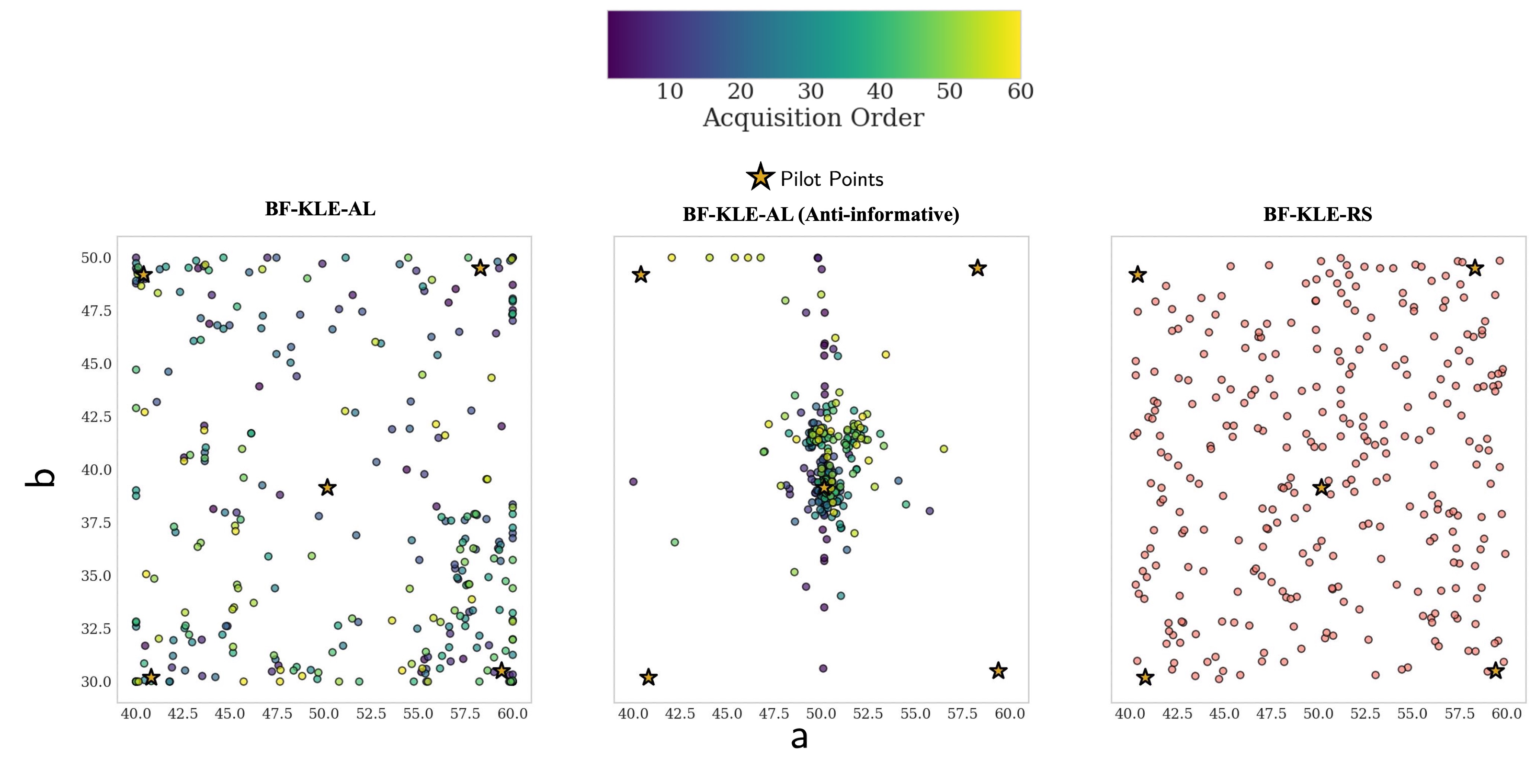}
    \caption{Pilot HF samples and additional points acquired via BF-KLE-AL (left), anti-informative sampling (middle), and BF-KLE-RS (right), aggregated over 5 replicates for C2. Points are colored by acquisition order. 
        }
    \label{fig: hf_acquired_1d}
\end{figure}

The results reveal several clear trends. First, the reported anti-informative comparator has the largest errors in this experiment, exceeding those of random sampling. Second, while surrogate accuracy generally improves as more HF points are added, the relative benefit of active learning depends strongly on the LF-HF correlation structure. In C1, where correlation varies sharply across the domain, active learning and random sampling perform comparably up to a point, with little to no advantage for EI-based myopic selection. After approximately 45 HF points, random sampling has better accuracy gains.
In contrast, for C2 where the LF and HF response fields exhibit stronger and more spatially uniform correlation, active learning yields more noticeable accuracy gains (though still modest) once approximately 30 HF samples have been acquired. Figure \ref{fig: hf_acquired_1d} illustrates the batchwise acquisition locations for C2, aggregated over five replicates.
The relatively small performance gap between BF-KLE-AL and BF-KLE-RS in this example may stem from the fact that the active learning heuristic, despite targeting regions of higher estimated error (e.g., with some clustering at the corners), still produces a broadly dispersed sampling pattern. As a result, the benefit over random sampling is diminished, especially given the already strong LF signal. 

\begin{figure}[htbp]
    \centering
        \includegraphics[width=\textwidth]{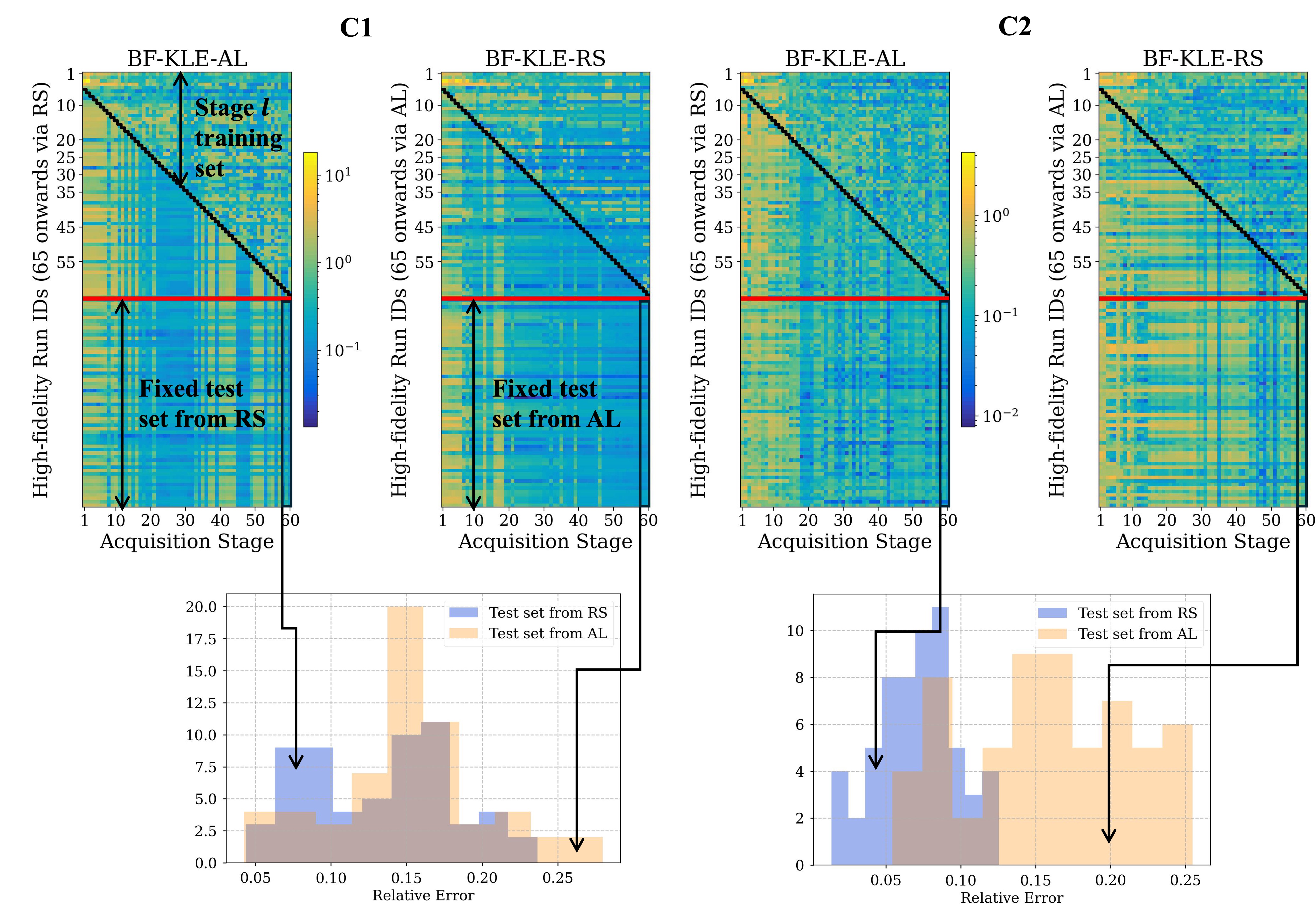}
    \caption{Relative errors for all surrogate training points in a single replicate for C1 (top left) and C2 (top right). Entries below the red separator indicate predictive performance on a small test set comprising points from the competing strategy. Histograms (bottom) at the last data acquisition stage show the corresponding distributions of relative errors.}
    \label{fig:err_1d_single_rep_updated_L1}
\end{figure}

To further examine how new acquisitions influence model performance, Figure~\ref{fig:err_1d_single_rep_updated_L1} presents the evolution of the relative errors for a representative replicate for both BF-KLE-AL and BF-KLE-RS. Two types of error trends are analyzed:
\begin{enumerate}
    \item errors at the training points used to construct the bifidelity surrogate, both before and after each point is incorporated into the training set; and

    \item errors on a separate test set derived from the competing strategy (e.g., BF-KLE-RS samples held out for the BF-KLE-AL surrogate and vice versa), providing a complementary generalization check.
\end{enumerate}
Each subfigure in Figure~\ref{fig:err_1d_single_rep_updated_L1} can be interpreted as follows. Each pixel row corresponds to a specific data point, and each column represents a data acquisition stage (either through active learning or random sampling). Pixels below the black line denote points that are currently serving as test samples, while pixels above correspond to points included in the training set. As the algorithm proceeds from left to right, the black line shifts downward, indicating that more points have been incorporated into the training set. Points below the red line represent a separate test set from the competing strategy (per item 2 above) that is not used for training that particular surrogate model. 

The histogram insets at the final stage summarize the distribution of relative errors across these test sets. Consistent with the observations in Figure~\ref{f: oracle_multiple_reps_LF_c1_c2_with_EI_50batch}, the average relative error on the random-sampling test set is lower for the BF-KLE-AL surrogate in C2, indicating improved generalization from active learning. In contrast, the BF-KLE-RS surrogate exhibits noticeably higher test errors. For C1, both strategies yield comparable performance, though the BF-KLE-RS surrogate performs marginally worse for this specific replication. Additional discussion of the relative error evolution follows in \ref{app:ex-1b-hf-lf-families}.

\subsubsection{Forward UQ using KLE surrogates}

\begin{figure}[htbp]
    \centering
        \includegraphics[width=0.99\linewidth]{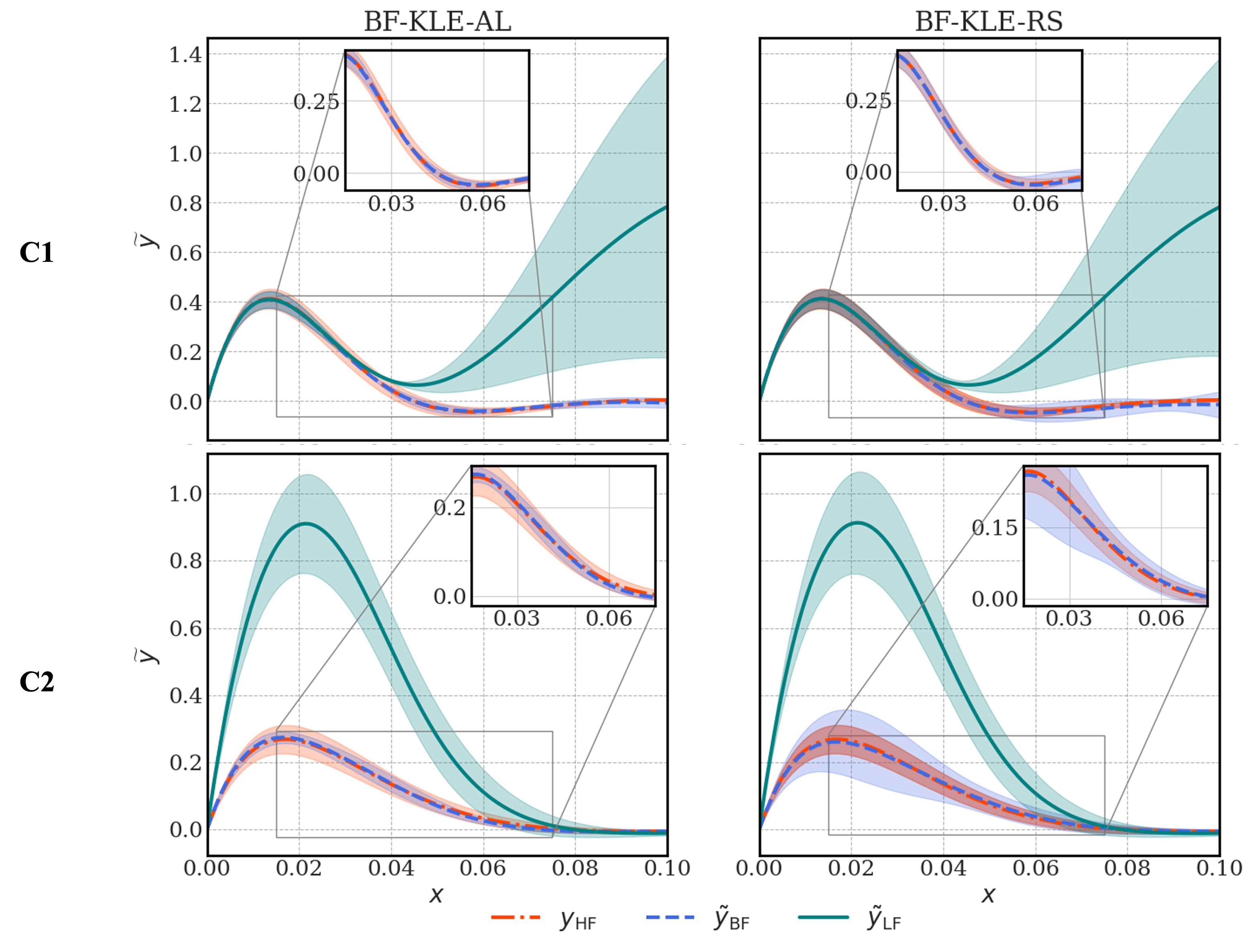}
    \caption{Comparison of forward UQ results: QoI prediction mean $\pm1$ standard deviation for $y_{\mathrm{HF}}$, $\tilde{y}_{\mathrm{BF}}$, and $\tilde{y}_{\mathrm{LF}}$. The bifidelity surrogates $\tilde{y}_{\mathrm{BF}}$ are constructed via active learning (left) and random sampling (right) strategies for C1 (top) and C2 (bottom) for a single replication. Those from active learning successfully correct the divergence in the LF predictions, and better represent the uncertainty in $y_{\mathrm{HF}}$. The insets provide a closer view of $y_{\mathrm{HF}}$ and $\tilde{y}_{\mathrm{BF}}$ in the overlapping regions.}
    \label{fig:surr_uncertainty}
\end{figure}

We now compare forward UQ results obtained using the bifidelity KLE surrogates. The displayed mean and standard deviation summarize the propagation of the prescribed input distribution through each model. 
Figure~\ref{fig:surr_uncertainty} compares the mean and $\pm1$ standard deviation bounds propagated directly from the true HF model $y_{\mathrm{HF}}$ with the same predictive statistics from the bifidelity and LF-KLE surrogates. The LF surrogate is constructed solely from LF data obtained in the bifidelity case.

Several observations can be made. First, due to the bias in the pilot LF snapshots, the LF-KLE surrogate fails to accurately reproduce the mean and variability of the HF model. In C1, the divergence of the LF approximation with increasing $x$ leads to inflated uncertainty in the LF-KLE predictions, a trend that also persists in the surrogate constructed via random sampling. In contrast, the BF-KLE-AL surrogate more accurately captures the mean response and successfully corrects for the systematic bias present in the LF approximations.

Notably, BF-KLE-AL somewhat underpredicts the uncertainty in each case as seen from the inset plots. On the other hand, BF-KLE-RS uncertainty estimates are generally inflated, particularly in regions of high LF approximation bias.

\subsection{Example 1b: Extended cases for 1D pulse function}
\label{ss: ex1-1d-part2}
The previous results showed setting-dependent differences between active learning and random sampling for particular paired HF-LF functions at fixed budgets. Next, we conduct ablation studies to determine the impact of key parameters on the BF-KLE performance relative to a single-fidelity HF-KLE counterpart built with approximately matched computational cost. We focus on parameters that commonly characterize multifidelity performance in various problem settings. These include:
\begin{enumerate}
    \item The HF-LF correlation trend,
    \item The per-sample HF-LF cost ratio, and
    \item The budget allocation for LF and discrepancy KLE terms.
\end{enumerate}

We describe the setup and findings from these experiments next.

\subsubsection{Surrogate Construction}
\label{sss: ex1-part2-surr-cons}

\begin{figure}[htbp]
    \centering
    \includegraphics[width=\linewidth]{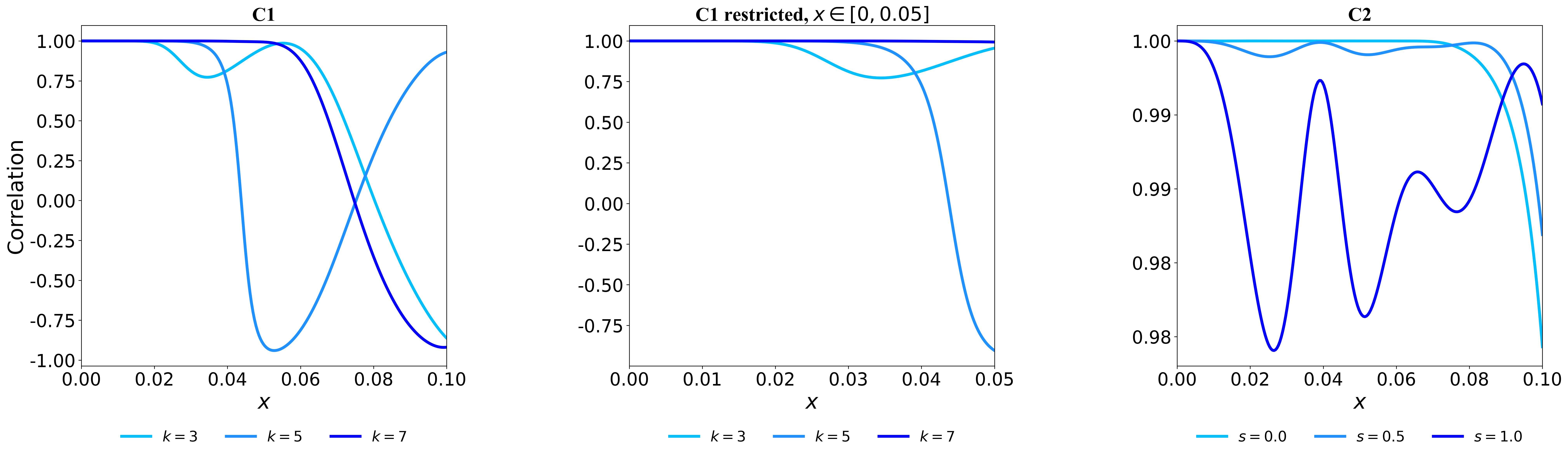}
    \caption{HF-LF correlation trends for cases C1 (left), C1 restricted (middle) and C2 (right), induced by variation of truncation order $k$ for C1 and the constants in the sine approximation formula \citep{stroethoff_bhaskaras_2014} through $s$ for C2. The restricted C1 LF model better approximates the true sine behavior over $x$.}
    \label{fig:surr_corrs_ex1}
\end{figure}

In Section~\ref{ss: ex1-1d}, the choice of LF approximation influenced the trend of the correlation with the HF reference. C1 correlations declined sharply before recovering partially, and the C2 model maintained a high, stable correlation up to $x=0.1$. Here, we extend these cases by varying their respective hyperparameters. For C1, we modify the order $k$ of the truncated Taylor expansion of the sine term and study $k=3, 5, 7$, where $k=5$ corresponds to the original C1 setup. We also report results for C1 with HF and LF models defined over $x \in [0, 0.05]$, where the Taylor approximation does not diverge significantly over this range of $x$. The C2 extension makes use of the expression:
\begin{align}
    y_{\textrm{LF},2}(x,\theta; s) = \exp(-ax)\left[\frac{(4 - 0.5s) bx\frac{180}{\pi}\left(180 - bx\frac{180}{\pi}\right)}{(40500 - 25500s) - bx\frac{180}{\pi}\left(180 - bx\frac{180}{\pi}\right)}\right],
    \label{eq:stroethoff_c2_modified_s_tuning}
\end{align}
where $s=0,0.5,1$ are studied. The value $s=0$ gives the sine approximation in \citet{stroethoff_bhaskaras_2014}, while $s=1$ recovers the original C2 expression in Section~\ref{ss: ex1-1d}. The respective HF-LF correlations are presented in Figure~\ref{fig:surr_corrs_ex1},
while Figure~\ref{fig:1d_samples_lf_hf_c1_additional} and Figure~\ref{fig:1d_samples_lf_hf_c2_additional} show evaluations of the models at selected values of $\theta$.

Each ablation setting is repeated over 20 replications. Within a single replication, a large space-filling Latin hypercube (LH) design is generated over the two-dimensional input domain. Smaller designs (the LF pilot set, the discrepancy correction set augmented at each stage of acquisition, and the single-fidelity HF-only set) are drawn as nested sub-designs with good space-filling properties.
This choice simplifies the design and construction of the surrogates and ensures better coverage of the $\theta$-space for all surrogates in the absence of AL. The LF-KLE is built from the LF pilot set consisting of $N_{\textrm{LF}}$ simulations, and not updated further during acquisition of additional HF simulations. The discrepancy set is additionally constrained to be a subset of the LF pilot set to construct paired HF-LF evaluations.

To compare BF-KLE and single-fidelity HF-KLE at rounded equivalent computational budgets, the number of HF simulations required for the latter at acquisition stage $\ell$ is determined via:
\begin{align}
    N_{\textrm{HF}}^{(\ell)} = \lceil N_{\textrm{LF}} / r \rceil + N_{\Delta}^{(\ell)}, 
    \label{eq:n_hf_equiv_1d}
\end{align}
where $r \in \{4,8,16\}$ is the per-sample HF-LF cost ratio. Because of the ceiling, the HF-only cost can exceed the BF cost by less than one HF evaluation when $N_{\mathrm{LF}}/r$ is nonintegral. This rounding favors the HF-only baseline slightly; the budgets are exactly equal when that ratio is integral. The surrogates are compared at varying budgets for the LF-KLE, namely $N_{\textrm{LF}}\in\{100, 125, 150\}$. We set $ N_{\textrm{HF}}^{(\ell)} = N_{\Delta}^{(\ell)}$ for the special case where the cost of LF simulations is neglected in the sampling budget (``free-LF'' in the results).

\begin{figure}[htbp]
    \centering
    \includegraphics[width=0.6\linewidth]{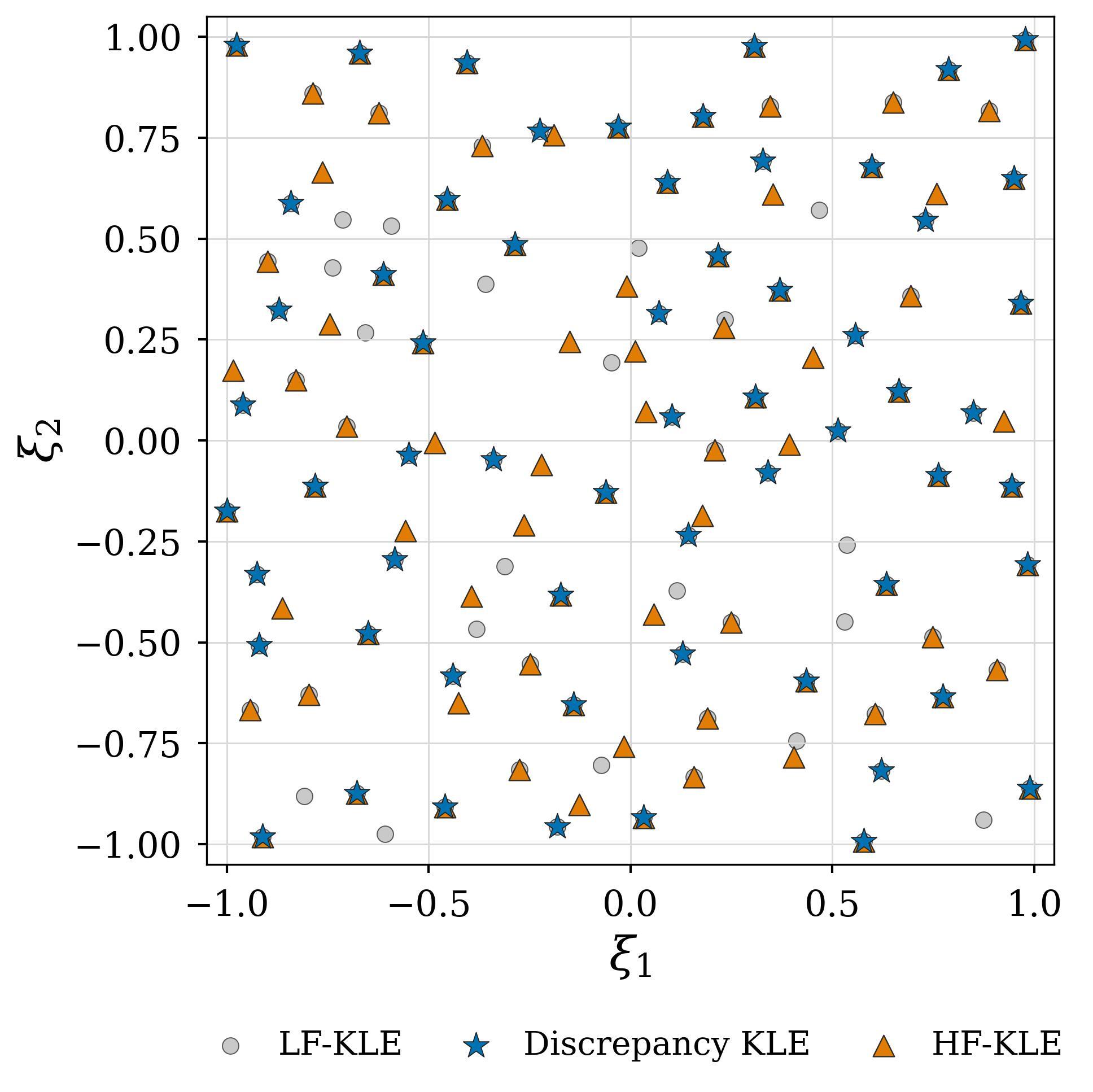}
    \caption{Design for the 1D pulse problem in the non-dimensional $(\xi_1, \xi_2)$ space for comparing BF-KLE and HF-KLE, with $B=85, r=4$.}
    \label{fig:pilot_design_ex1_part2}
\end{figure}

As an illustration, for cost ratio $r=4$, a 185-point LH design is drawn initially in the two-dimensional $\theta$-input space. A subset of $N_{\textrm{LF}}=100$ realizations is used to construct the LF-KLE. Setting batch size $q=15$ and $L=4$ acquisition stages, nested subsets of 15, 30, 45 and 60 points are drawn to build the discrepancy surrogate and subsequently the BF-KLE. The equivalent HF-KLE budget is set via Equation~\ref{eq:n_hf_equiv_1d} to 40, 55, 70 and 85 HF simulations, with total computational budget $B= N_{\textrm{HF}}^{(L)} = 85$ once acquisition is complete. This process is repeated for 20 independent replicates. The design sets for a particular replicate ($\Theta_{\textrm{LF}}, \Theta_{\Delta}$ and $\Theta_{\textrm{HF}}$) are shown in Figure~\ref{fig:pilot_design_ex1_part2}.

\subsubsection{Results}
\label{sss:pulse_function_family_results}

\begin{figure}[htbp]
    \centering
    \includegraphics[width=\linewidth]{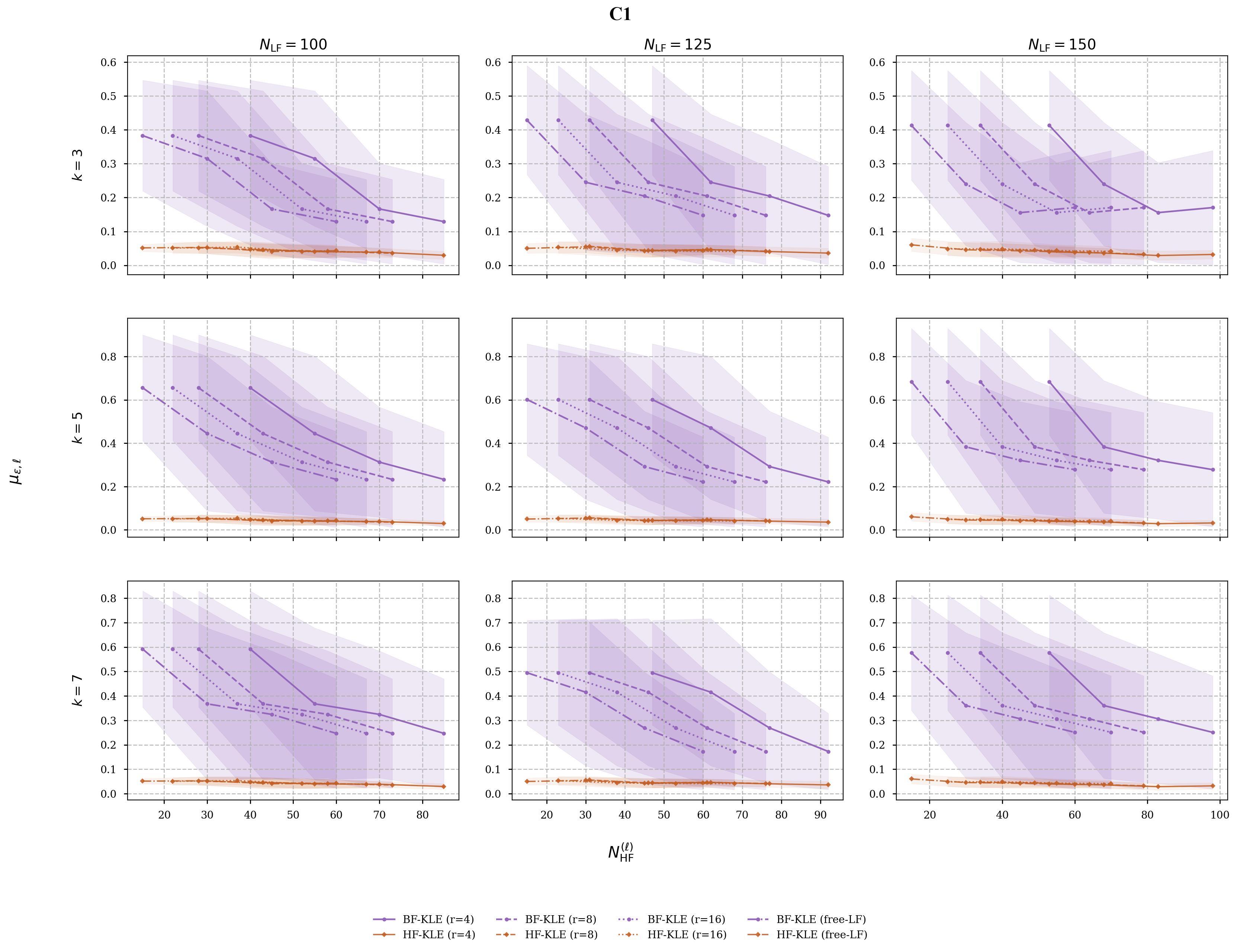}
    \caption{Summary of integrated relative error  $\mu_{\varepsilon,\ell}$ for BF-KLE and matched-cost HF-KLE in C1 on $x\in[0,0.1]$, over 20 replicates. Rows correspond to Taylor truncation orders $k=3,5,7$, and columns to LF-KLE dataset sizes $N_{\mathrm{LF}}=100,125,150$. Line styles distinguish the cost ratios $r=4,8,16$ and the free-LF case. The horizontal axis shows the rounded equivalent HF budget.}
    \label{fig:rel_err_trends_c1}
\end{figure}

\begin{figure}[htbp]
    \centering
    \includegraphics[width=\linewidth]{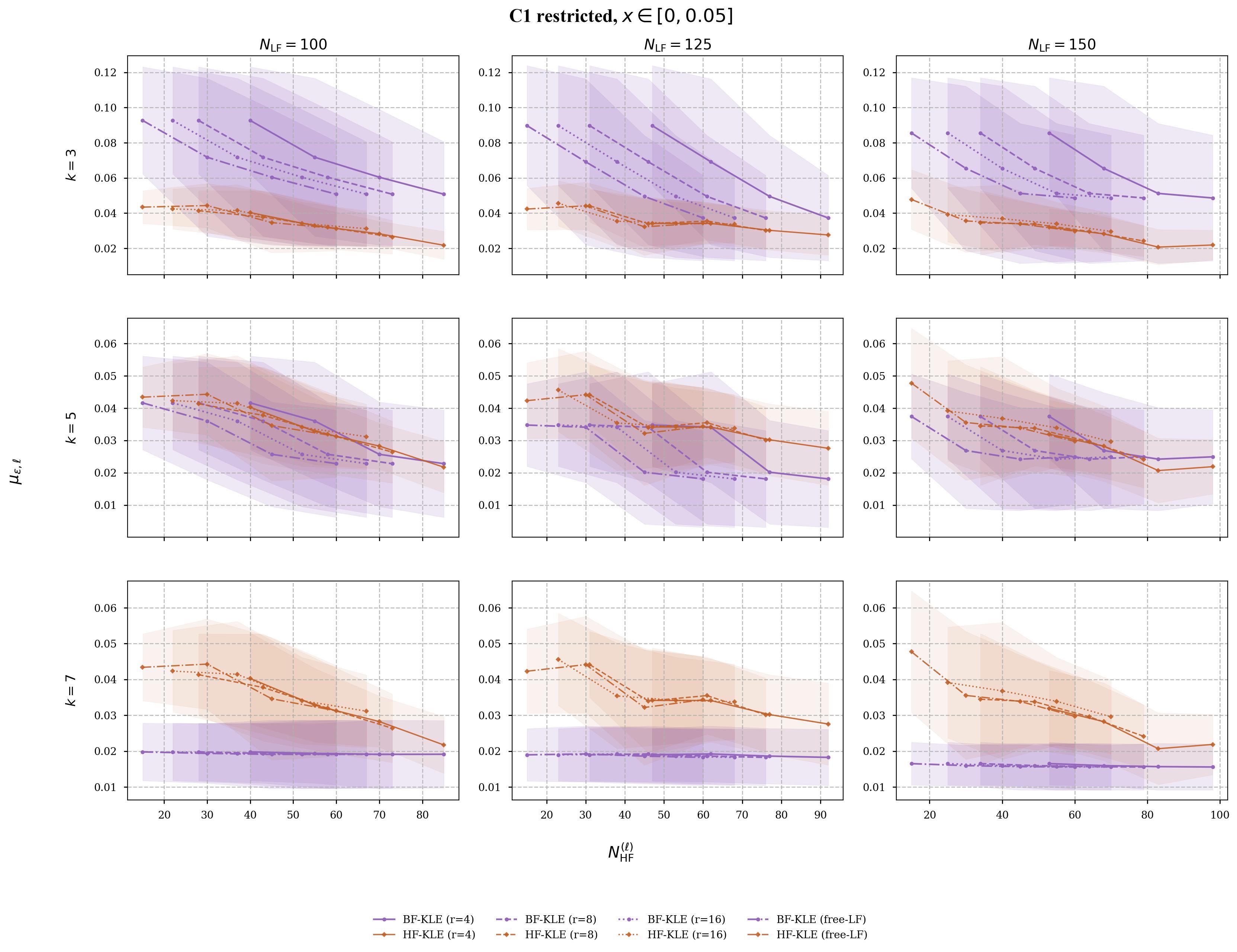}
    \caption{Summary of integrated relative error  $\mu_{\varepsilon,\ell}$ for BF-KLE and matched-cost HF-KLE in the restricted C1 cases, $x\in[0,0.05]$, over 20 replicates. Rows correspond to Taylor truncation orders $k=3,5,7$, and columns to LF-KLE dataset sizes $N_{\mathrm{LF}}=100,125,150$. Line styles distinguish the cost ratios $r=4,8,16$ and the free-LF case. The horizontal axis shows the rounded equivalent HF budget.}
    \label{fig:rel_err_trends_c1_restricted}
\end{figure}

\begin{figure}[htbp]
    \centering
    \includegraphics[width=\linewidth]{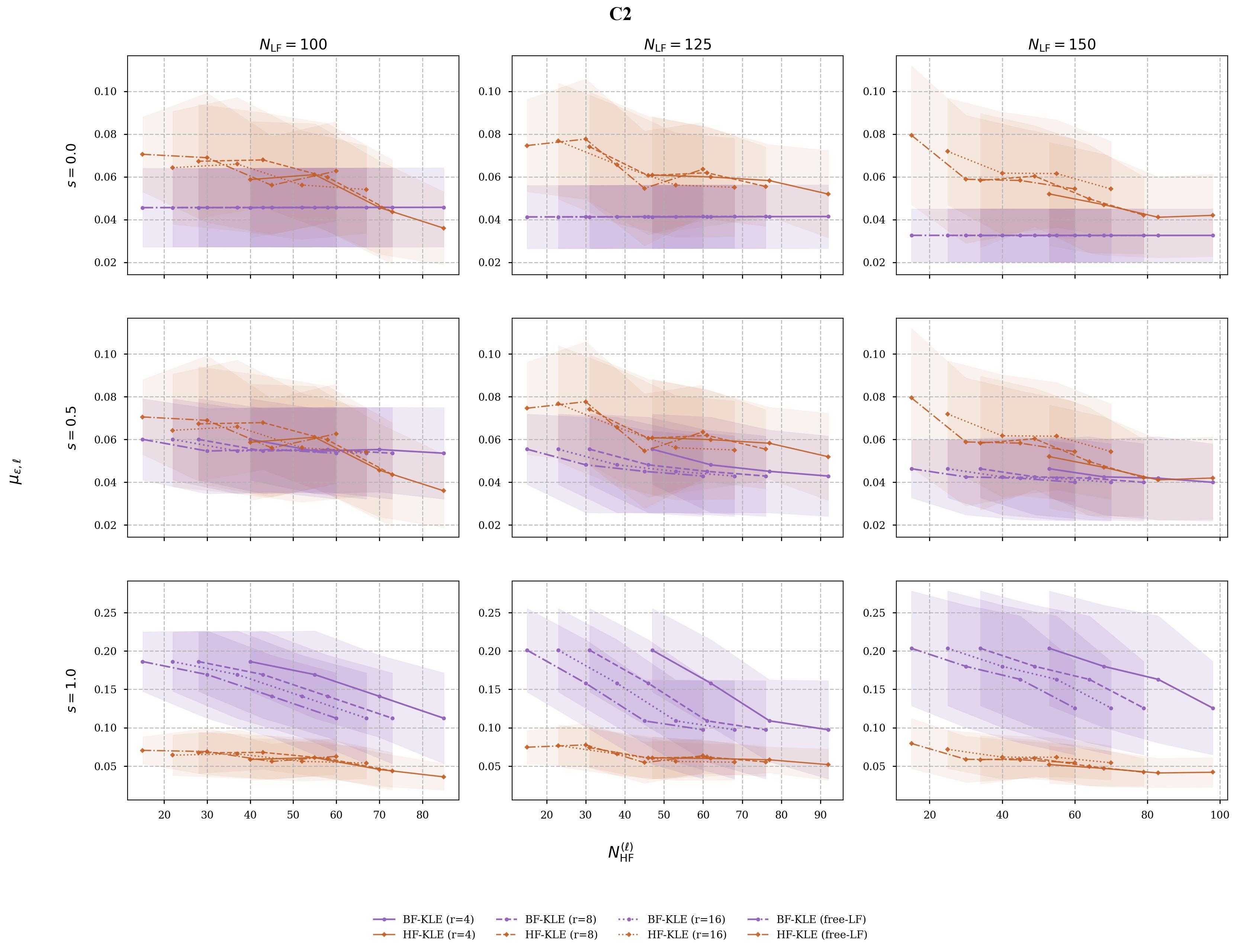}
    \caption{Summary of integrated relative error  $\mu_{\varepsilon,\ell}$ for BF-KLE and matched-cost HF-KLE in C2, over 20 replicates. Rows correspond to sine-approximation parameters $s=0,0.5,1$, and columns to LF-KLE dataset sizes $N_{\mathrm{LF}}=100,125,150$. Line styles distinguish the cost ratios $r=4,8,16$ and the free-LF case. The horizontal axis shows the rounded equivalent HF budget.}
    \label{fig:rel_err_trends_c2}
\end{figure}

The integrated relative error metric (estimated via Equation~\ref{eq:oracle_err_defn}) measures each surrogate's predictive accuracy against the analytical HF model. Figures~\ref{fig:rel_err_trends_c1}, \ref{fig:rel_err_trends_c1_restricted}, and \ref{fig:rel_err_trends_c2} report the mean error and $\pm 1$ standard deviation bounds over 20 independent replicates as the paired discrepancy set grows from $N_{\Delta}^{(\ell)}=15$ to $60$. The horizontal axis represents the equivalent HF budget (Equation~\ref{eq:n_hf_equiv_1d}). Alternate visualizations reporting only the final stage BF-KLE advantage ($N_{\Delta}^{(\ell)}= 60$) are available in \ref{app:ex-1b-hf-lf-families}.

For C1 on the full domain $x\in[0,0.1]$, BF-KLE is consistently less accurate than the matched-cost HF-KLE (Figure~\ref{fig:rel_err_trends_c1}). The Taylor approximation deteriorates rapidly for larger values of $x$ and $b\in[60,80]$. In this range, retaining additional terms in the expansion does not produce a uniformly better LF model (Figure~\ref{fig:1d_bsine_all_s}). The $k=3$ BF-KLE has lower average error than the $k=5$ and $k=7$ variants, although it remains substantially less accurate than HF-KLE. While the BF-KLE error decreases as the discrepancy set grows, it is unable to offset the LF bias for the tested cost-ratio and sampling budget settings.

Restriction of the spatial domain to $x\in[0,0.05]$ ensures that the Taylor approximations remain closer to $\sin(bx)$ and reverses these trends (Figure~\ref{fig:rel_err_trends_c1_restricted}). 
While the $k=3$ BF-KLE remains generally less accurate than HF-KLE, the two surrogates are comparable for $k=5$, with a modest BF-KLE advantage for $N_{\mathrm{LF}}=125$ but no consistent advantage at $N_{\mathrm{LF}}=100$ or 150. When the truncation order is increased to $k=7$, the resulting BF-KLE  is significantly better than the HF-KLE. It shows little to no change in performance trends as additional paired evaluations are introduced and small improvements with increasing LF-KLE budget.

For C2, the BF-KLE advantage over HF-KLE degrades as $s$ increases (Figure~\ref{fig:rel_err_trends_c2}). At $s=0$, the sine approximation of \citet{stroethoff_bhaskaras_2014} is very accurate  (Figure~\ref{fig:1d_bsine_all_s}), resulting in similar trends to those observed for the seventh-order Taylor expansion for C1 ($x \in [0, 0.05]$). At $s=0.5$, the final-stage margins (Figure~\ref{fig:heatmaps_all_c2}) are negative for $N_{\textrm{LF}}=100$ and negligible for $N_{\mathrm{LF}}=125$ and 150. The latter results do not suggest a strong preference for BF-KLE over HF-KLE in the intermediate case. Finally, at $s=1$, BF-KLE is consistently less accurate in the reported comparison. Since the discrepancy budget and the order-three PCE representation are unchanged across $s$, the available discrepancy surrogate is not able to fully correct the biased LF predictions, though the BF-KLE performance improves steadily as additional paired evaluations become available at higher values of $r$.

These ablations identify favorable conditions for BF-KLE: an LF model that captures relevant HF features, a sufficiently accurate LF surrogate, and enough paired data to approximate the discrepancy at the available cost. Pointwise correlation alone does not determine performance. LF bias, discrepancy complexity, and approximation choices vary together in these studies, so the observed trends do not establish necessary or sufficient conditions for a bifidelity advantage.

Extensions of the methodology to permit more flexible representations of the discrepancy field, training and budget allocation strategies and additional hyperparameter tuning (for instance, the PCE truncation scheme) may improve performance; however, these are not evaluated in the current ablation. \ref{app:deeponet_comparisons} compares the BF-KLE setup with a neural-network-based bifidelity DeepONet on a benchmark problem from \cite{howard_multifidelity_2023}, where the DeepONet representation has better predictive accuracy, albeit at the cost of higher offline training time.

In practice, when constrained by lower sampling budgets and non-negligible LF costs, the BF-KLE advantage increases when the chosen LF model can reproduce the important features of the HF response accurately over the spatial or temporal domain (as seen for $s=0$ in C2 and $k=7$ in C1 for small $x$) and is much cheaper to evaluate compared to the HF model. Such characteristics allow a larger fraction of the budget to be used to better represent the discrepancy field and BF-KLE surrogate refinement through AL. Some of the QoIs modeled in the turbulent jet flow problem (Section~\ref{ss:jetflow}) exhibit these features for the chosen HF and LF models, and the subsequent results indicate the potential of bifidelity modeling and active learning beyond synthetic test cases.

\subsection{Example 2: 2D Convection-Diffusion}
\label{ss: ex2-2d}

Next, we demonstrate the proposed method on a more complex problem: a two-dimensional scalar convection-diffusion equation. Let $\phi(\mathbf{x}, t)$ denote a scalar field defined on the spatial domain $\mathbf{x}=(x, y)\in[0,1]^2$. 
The governing PDE is:
\begin{align}
\frac{\partial\phi(\mathbf{x}, t)}{\partial t}+\nabla\cdot(\mathbf{u}\phi(\mathbf{x}, t))=\nabla\cdot(\alpha\nabla\phi(\mathbf{x}, t))+\dot{\omega}_\phi, \qquad t > 0,
\label{eq:pde_2d}
\end{align}
where $\mathbf{u}=(u,v)$ is the spatially varying advection velocity, $\alpha=0.01$ is the constant diffusion coefficient, and $\dot{\omega}_\phi(x, y)$ is a source (reaction) term. 
The initial condition is $\phi(\mathbf{x}, t=0) = 0$
and boundary conditions are periodic.

The components of the advection velocity are defined as:
\begin{align}
u(x,y)&=\frac{1}{10}-\sin{(\pi x)^2}\Big[\sin{(\pi(y-0.05))}\cos{(\pi(y-0.05))} \nonumber\\
& \hspace{7em}-\sin{(\pi(y+0.05))}\cos{(\pi(y+0.05))}\Big],\label{eq:u_vel} \\
v(x,y)&=\sin{(\pi x)}\cos{(\pi x)}\left[\sin{(\pi(y-0.05))}^2-\sin{(\pi(y+0.05))}^2\right].\label{eq:v_vel}
\end{align}
The source term is:
\begin{align}
\dot{\omega}_\phi(x,y)&= \frac{\theta_s}{2\pi \theta_h^2} \Bigg[\exp\left(-\frac{(x-\theta_x)^2+(y-\theta_y)^2}{2\theta_h^2}\right)\nonumber\\
&\hspace{3.5em} -\exp\left(-\frac{(x-\theta_x + 0.05)^2+(y-\theta_y + 0.05)^2}{2 \theta_h^2}\right)\Bigg],
\label{eq:source_fn}
\end{align}
where $\theta = [\theta_s, \theta_h, \theta_x, \theta_y]$ comprises four uncertain parameters representing the source strength, width, and spatial coordinates. The independent distributions are: $\theta_s \sim \CU(0.01, 0.05)$, $\theta_h \sim \CU(0.05, 0.08)$, $\theta_x \sim \CU(0.3, 0.7)$, and $\theta_y \sim \CU(0.55, 0.85)$. 

\begin{figure}[htbp]
    \centering
    \includegraphics[width=\linewidth]{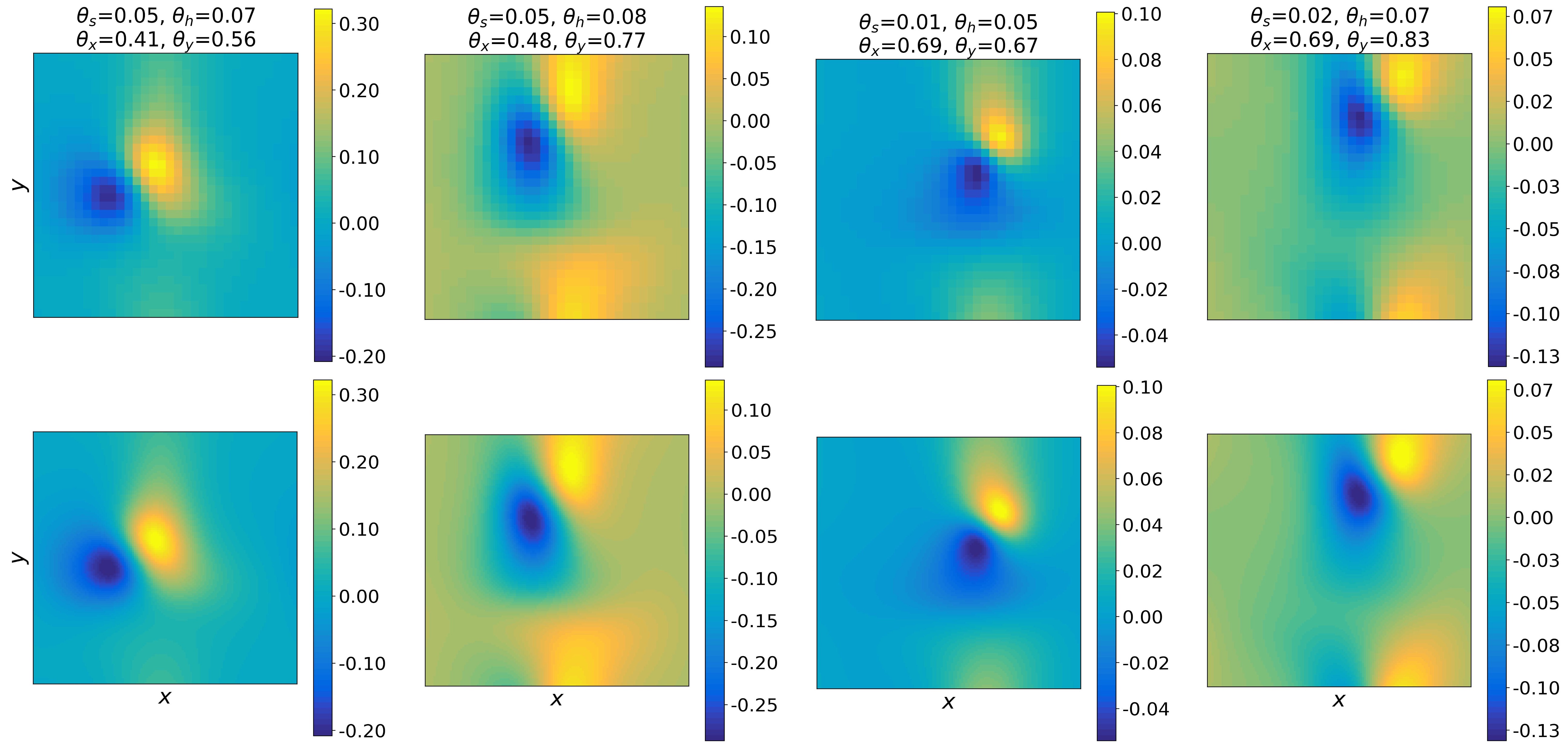}
    \caption{Representative snapshots of $\phi(\mathbf{x}, t=2.5)$ at selected $\theta$ parameter values, for LF (top row) and HF (bottom row) simulations.
    }
    \label{fig:2d_lf_hf_samples}
\end{figure}

Two model fidelities are defined using different grid resolutions,
HF: $128 \times 128$ uniform grid, and LF: $32 \times 32$ uniform grid. 
The HF field solutions are restricted to the LF grid for constructing the bifidelity surrogate.
The PDE is integrated in time up to $t=2.5$ using first-order explicit time stepping. Convection terms are discretized using a first-order upwind scheme, and diffusion terms using second-order central differencing. The Courant--Friedrichs--Lewy (CFL) condition determines the time step for each fidelity, resulting in $\Delta t_{\mathrm{LF}}=0.02$ and $\Delta t_{\mathrm{HF}}=0.0012$ for $\mathrm{CFL}=0.8$. 
The QoI is defined as $y = \phi(\mathbf{x}, t=2.5)$. 
Representative LF and HF realizations (Figure~\ref{fig:2d_lf_hf_samples}) show strong correlation across the spatial domain. 

Pilot designs of size $(N_{\mathrm{LF}}^P, N_{\Delta}^P) = (300, 10)$ are generated using the same space-filling design procedure as in Example 1.
Each KLE is truncated to retain 99\% of the total variance, corresponding to approximately 4--12 modes, and the associated PCE employs a total-order expansion of degree 3. 

The computational budget for active learning is set to $B=50$ total HF evaluations. 
The GP model for the generalization error is constructed using 10-fold cross-validation.
All experiments are repeated over 20 independent replicates. 
Relative errors are computed using a 1000-sample Monte Carlo estimation of the integral in \eqref{eq:oracle_err_defn}.

\begin{figure}[htbp]
    \centering
    \includegraphics[width=0.7\textwidth]{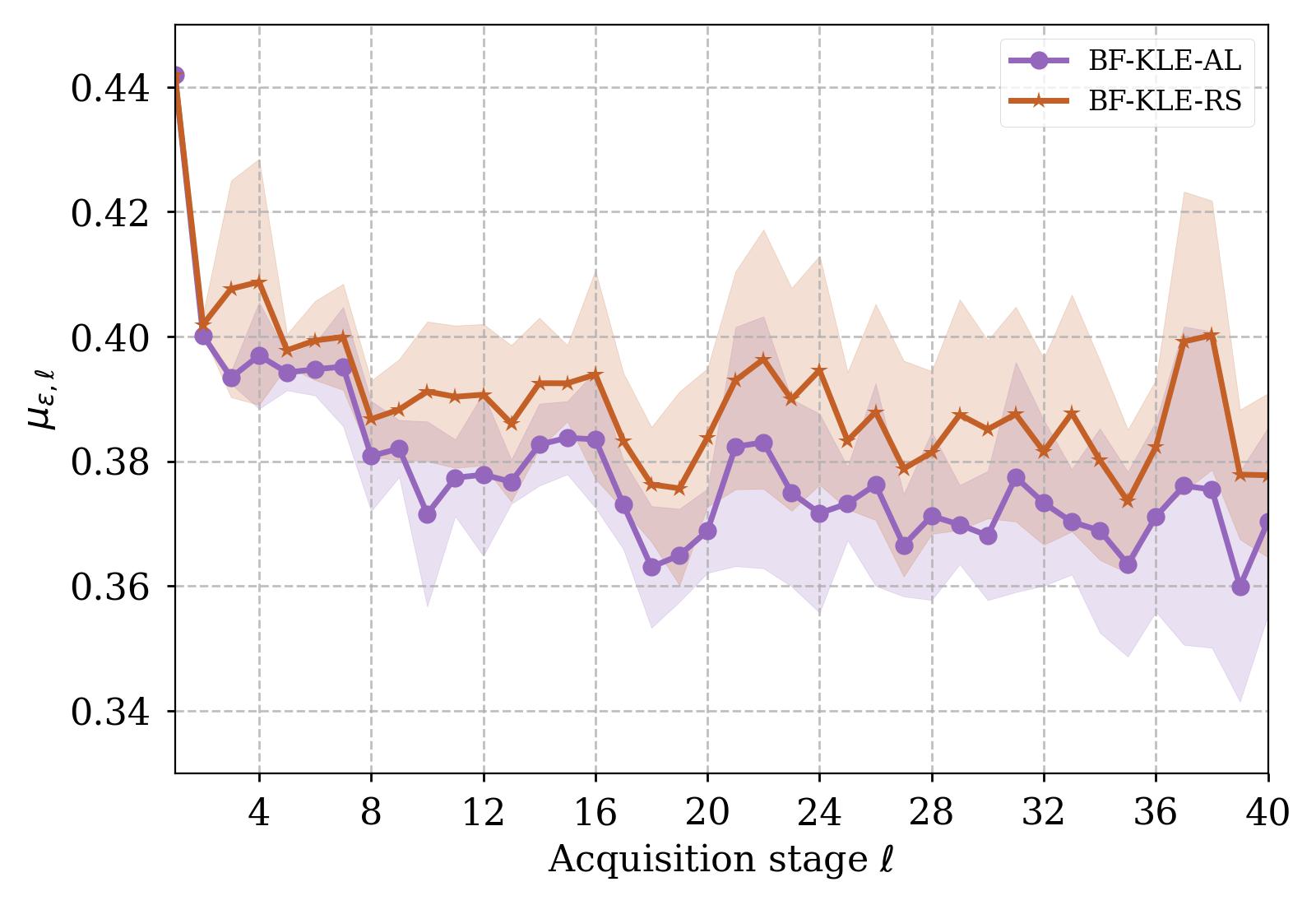}
    \caption{Comparison of performance against ground-truth HF solutions for BF-KLE-AL and BF-KLE-RS surrogates, averaged over 20 replicates.}
    \label{fig:oracle_2d_all_reps}
\end{figure}

Figure~\ref{fig:oracle_2d_all_reps} presents the evolution of the relative error as a function of data acquisition stage. Although the overall reduction in error is moderate (primarily because the discrepancy term contributes less to the total surrogate variance in this problem), the BF-KLE-AL surrogate consistently outperforms BF-KLE-RS on average.

\begin{figure}[htbp]
    \centering
    \includegraphics[width=\textwidth]{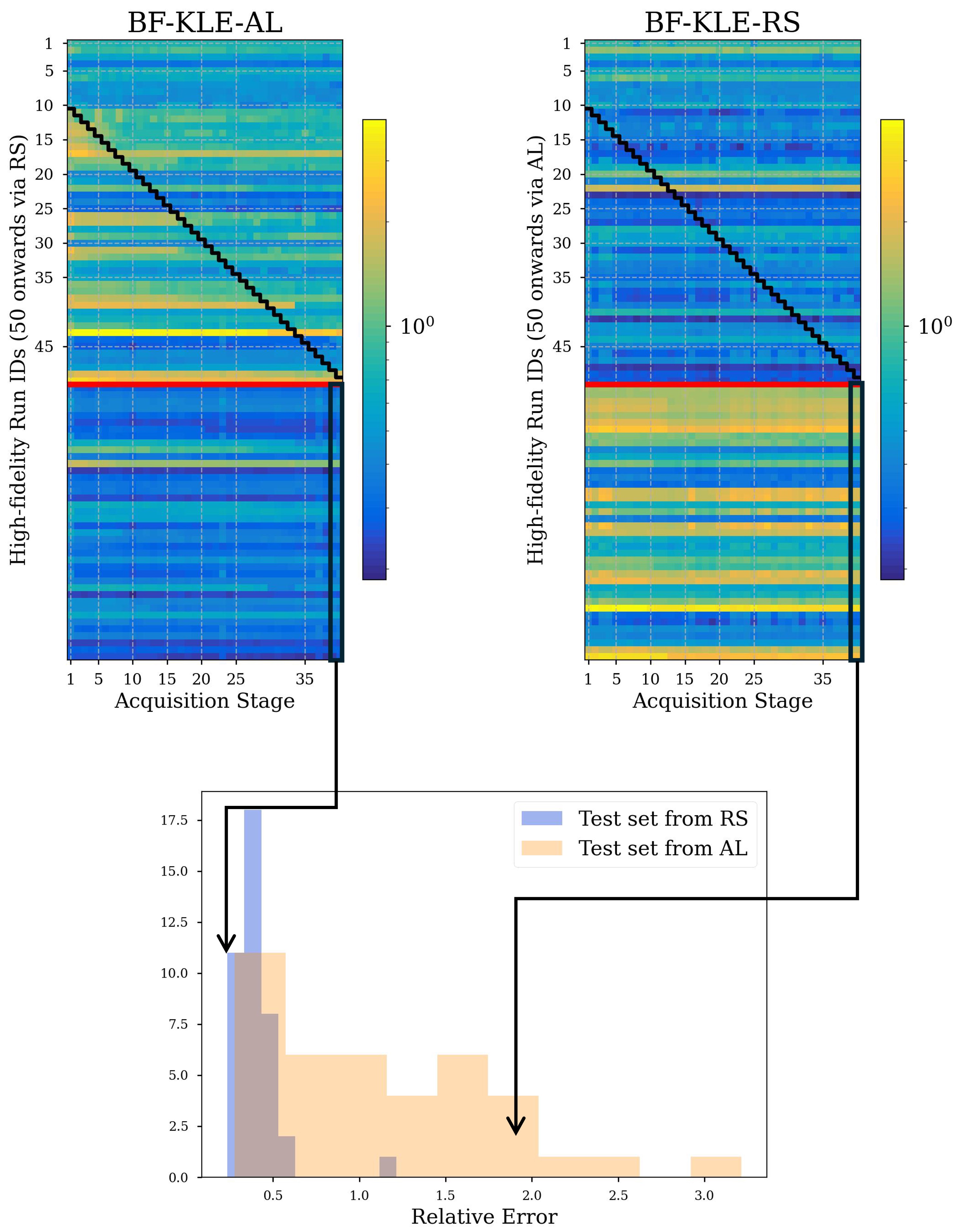}
    \caption{Comparison of relative errors for all surrogate training points (pilot and acquired) for a single replicate (top), similar to the previous 1D example with the red separator showing performance on a small test set from the competing strategy. The histogram (bottom) at the last data acquisition stage shows the corresponding distributions of relative errors.}
    \label{fig:samples_2d_1rep_heatmap}
\end{figure}

Figure~\ref{fig:samples_2d_1rep_heatmap} illustrates the detailed progression of relative errors for a representative replicate, following the same visualization format as Figure~\ref{fig:err_1d_single_rep_updated_L1}. 
For this 2D problem, the performance difference between the two strategies is evident across all stages: surrogates trained via active learning exhibit markedly lower relative errors on the random sampling test set compared with those trained by random selection.

\subsection{Example 3: 3D turbulent jet flow}\label{ss:jetflow}

In this example, we adopt a 3D turbulent jet flow from our earlier study~\citep{Jivani2021} focusing on the effects of parametric uncertainty on round jet behavior. The system involves an isothermal, axisymmetric jet exhausting into quiescent air at a jet Mach number $M_j = 0.9$. 
In the earlier study, uncertainty was introduced through two parameters: the stagnation pressure $p_0$ and the modified eddy viscosity ratio (MEVR) $\tilde{\nu}/\nu$, while a uniform velocity $u_e$ was imposed at the nozzle exit. In this work, we adopt a more physically realistic velocity profile and parameterize it to influence the downstream jet evolution.

Specifically, we model the nozzle exit velocity using a top-hat profile that peaks at the centerline and decays toward the nozzle wall:
\begin{align}
    u_e(r) = U_c \tanh \left(\frac{D / 2-r}{\kappa D}\right),
    \label{e: topHat}
\end{align}
where $u_e(r)$ is the axisymmetric velocity profile at radial location $r$, $D$ is the nozzle diameter, $U_c$ is a velocity scale, and $\kappa D$ controls the radial transition width The profile gives $u_e(0)=U_c\tanh(1/(2\kappa))$ at the centerline. 

We treat the following three input parameters as uncertain: $\theta = [U_c, \kappa, \log(\tilde{\nu}/\nu)]$, where 
$U_c\sim \CU(293.24,312.94)$ m/s, selected for subsonic flow near $M_j=0.9$; $\kappa\sim \CU(0.1,0.3)$; and $\log(\tilde{\nu}/\nu)\sim \CU(1.53,4.6)$
\footnote{The interval is chosen by mapping the inlet eddy-viscosity ratio $\mu_t/\mu\in[1,100]$ to the logarithm of the MEVR using the Spalart--Allmaras relation \citep{spalart_one-equation_1992}. Uniformity is then specified directly for the log-MEVR variable. A nonlinear transformation of a uniform eddy-viscosity ratio would generally produce a different distribution.
}.
The parameters $U_c$ and $\kappa$ shape the nozzle velocity profile, while the MEVR specifies a turbulence-model input. Together with the remaining flow boundary conditions, these parameters influence the local Mach number, stagnation conditions, and downstream evolution.  Stagnation pressure is not an independently sampled input in this study.

All simulations are performed using the SU2 open-source multiphysics solver \citep{Economon2016}. 
HF simulations are conducted using the enhanced delayed detached-eddy simulation (EDDES) approach, which resolves large-scale turbulent structures. LF simulations use the Reynolds-averaged Navier--Stokes (RANS) formulation. Both solvers employ the Spalart--Allmaras (SA) turbulence model for consistency. 
All runs are performed on the G1 mesh, which consists of approximately 1.6 million cells and is shared across partners of the EU Go4Hybrid (G4H) project consortium \citep{mockett_single-stream_2018_a}.

This HF-LF pairing introduces a notable computational cost difference: each EDDES simulation requires approximately 6.25 CPU hours, compared to about 1.5 hours for a corresponding RANS run, a roughly $4\times$ increase. The added cost stems from the temporal resolution needed to capture unsteady turbulent dynamics in EDDES, in contrast to the steady-state nature of RANS simulations. 

\begin{figure}[htbp]
    \centering
    \includegraphics[width=0.75\linewidth]{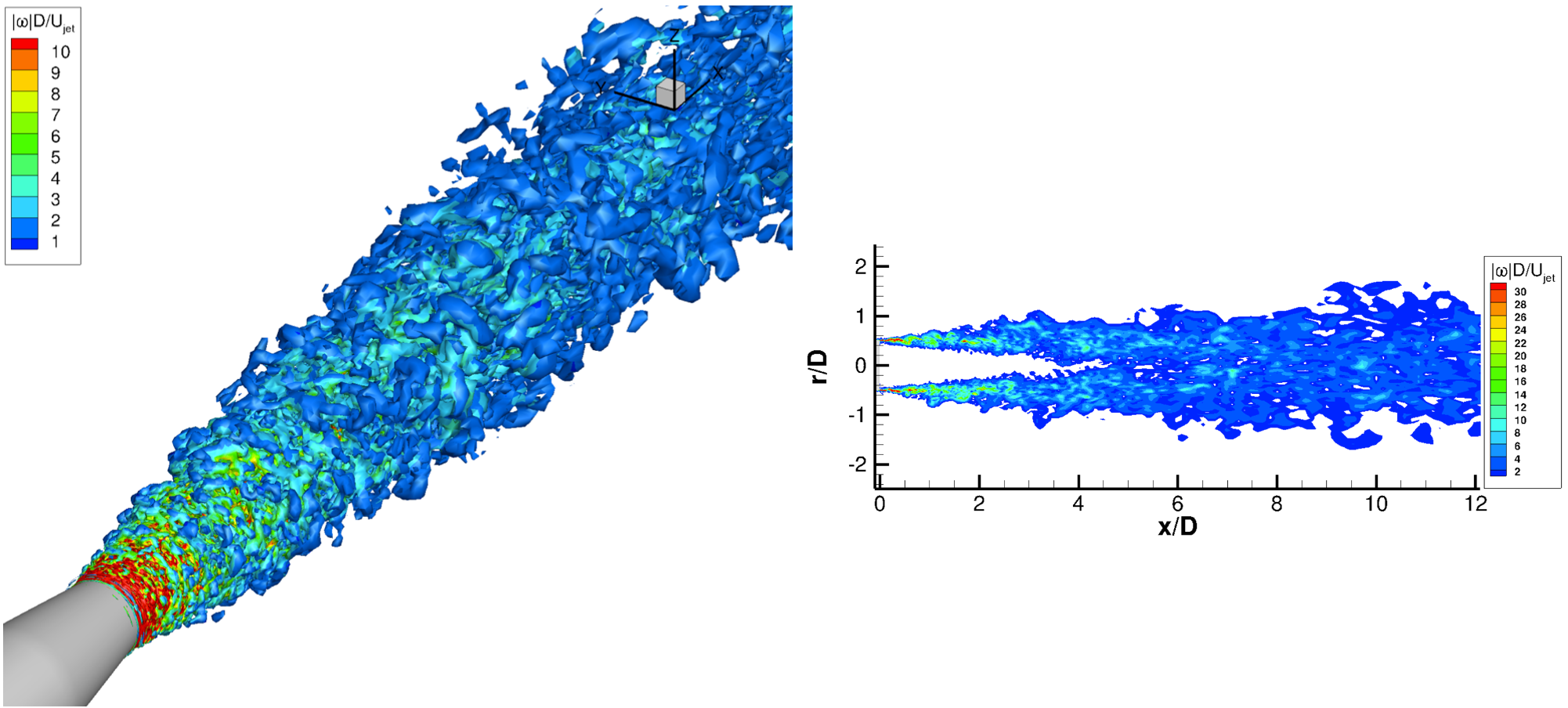}
    \caption{3D flow features in a turbulent round jet. Coherent turbulence structures effusing from the nozzle, visualized via Q-criterion isosurfaces colored by vorticity magnitude (left). Contour plot of vorticity magnitude on the jet meridian plane, highlighting shear layers and mixing regions (right).}
    \label{f:jet_flowfield}
\end{figure}

Figure~\ref{f:jet_flowfield} illustrates complex turbulent structures in a representative HF simulation using a 3D Q-criterion isosurface (colored by vorticity magnitude) and a meridional slice contouring vorticity magnitude. The Q-criterion is defined as
$Q = \frac{1}{2} \left( \lVert \Omega \rVert^2 - \lVert S \rVert^2 \right)$,
where
$\Omega = \tfrac{1}{2}(\nabla \mathbf{u} - \nabla \mathbf{u}^{\top})$ and $
S = \tfrac{1}{2}(\nabla \mathbf{u} + \nabla \mathbf{u}^{\top})$
are the antisymmetric (rotation-rate or vorticity) and symmetric (strain-rate) components of the velocity-gradient tensor $\nabla \mathbf{u}$, respectively, and $\mathbf{u}$ denotes the velocity field.

\begin{figure}[htbp]
    \centering
        \includegraphics[width=\textwidth]{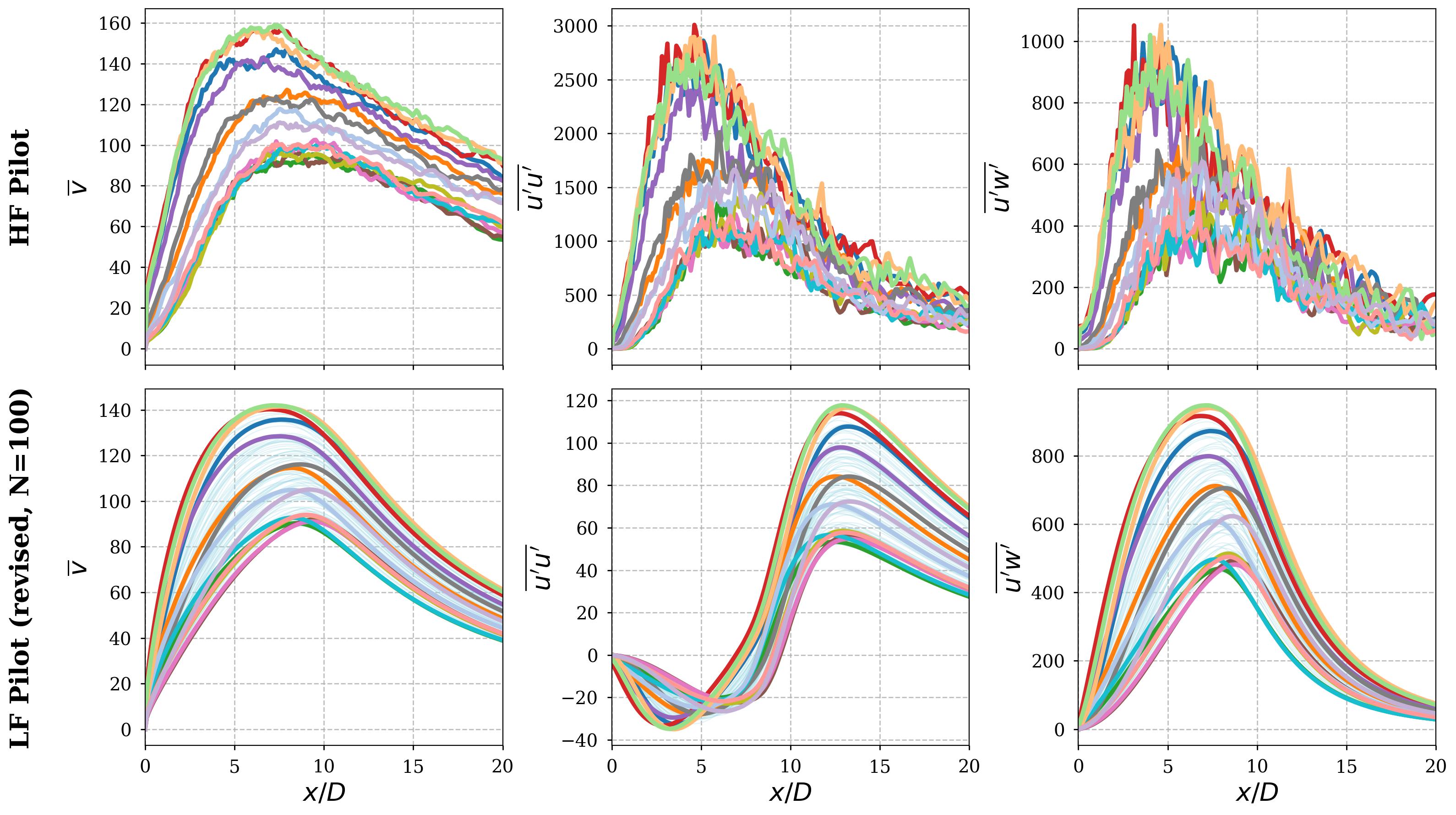}
    \caption{QoI profiles along the jet lipline from the 15 HF (EDDES) pilot simulations (top) and 100 LF (RANS) pilot simulations (bottom). From left to right: $\bar{v}$, $\overline{u'u'}$, $\overline{u'w'}$. Common LF-HF simulation pairs used to construct the discrepancy term are highlighted in the same color set.}
    \label{f: JetPilotQ}
\end{figure}

From these 3D solutions, we extract three key QoI profiles along the nozzle lipline: 
\begin{enumerate}
\item $\bar{v}$: the mean axial velocity; 
\item $\overline{u'u'}$: the normal component of Reynolds stress; and 
\item $\overline{u'w'}$: the shear component of Reynolds stress. 
\end{enumerate}
These QoIs are 1D fields, plotted in 
Figure~\ref{f: JetPilotQ} for a pilot dataset comprising 15 HF (EDDES) and 100 LF (RANS) simulations. They are extracted from the 3D flow solutions using ParaView, an open-source data analysis and visualization tool \citep{ayachit_paraview_2015}.

Pilot samples of size $(N_{\mathrm{LF}}^P=100, N_{\Delta}^P=15)$ are used to construct bifidelity KLE surrogates for each QoI. Here, each KLE is truncated to retain 95\% of the total variance, corresponding to approximately 1--2 modes for the LF term and 5--20 modes for the discrepancy term, and the associated PCE employs a total-order expansion of degree 3.

To assess the evolution of surrogate performance, we compare results of BF-KLE-AL against two reference surrogates: 
\begin{enumerate}
    \item BF-KLE-RS: built using the pilot set plus additional randomly selected HF-LF pairs.

    \item HF-KLE: Single-fidelity surrogate built from HF simulations alone.
\end{enumerate}

The total computational budget for BF-KLE-AL, BF-KLE-RS and their single-fidelity HF-KLE counterpart is set to 437.5 CPU hours overall. Of these, the pilot set for the bifidelity case makes up 243.75 CPU hours, while 25 additional paired HF-LF evaluations make up the rest. The HF-KLE is built using 70 HF evaluations (including the 15 simulations from the pilot set) in a one-shot manner. Finally, 5 held-out ground-truth HF simulations serve as an additional measure of predictive accuracy of each surrogate..
To guide the active learning process, the GP model for the generalization error is constructed using leave-one-out (LOO) cross-validation, averaged over all three QoIs, in place of \eqref{eq:kfold_err}: 
\begin{align}
        \varepsilon^{(i)} &= \frac{1}{3}
        \left(\frac{\|\bar{v}_{\mathrm{HF}}({x}, \theta_i) -\tilde{\bar{v}}_{\mathrm{BF}}^{-(i)}({x}, \theta_i)\|_{2}}{\|\bar{v}_{\mathrm{HF}}({x}, \theta_i)\|_{2}} + \frac{\|\overline{u'u'}_{\mathrm{HF}}({x}, \theta_i) -\widetilde{\overline{u'u'}}^{-(i)}_{\mathrm{BF}}({x}, \theta_i)\|_{2}}{\|\overline{u'u'}_{\mathrm{HF}}({x}, \theta_i)\|_{2}} \right. \nonumber\\
        & \hspace{3em} \left. + \frac{\|\overline{u'w'}_{\mathrm{HF}}({x}, \theta_i) -\widetilde{\overline{u'w'}}^{-(i)}_{\mathrm{BF}}({x}, \theta_i)\|_{2}}{\|\overline{u'w'}_{\mathrm{HF}}({x}, \theta_i)\|_{2}}\right).
    \label{eq:loo_err_metric}
\end{align}
In order to leverage available computational resources and enable parallel HF simulations, we employ batch active learning with a batch size of $q = 5$. This allows up to five HF evaluations to be executed simultaneously during each acquisition stage.

\begin{figure}[htbp]
    \centering
    \includegraphics[width=\linewidth]{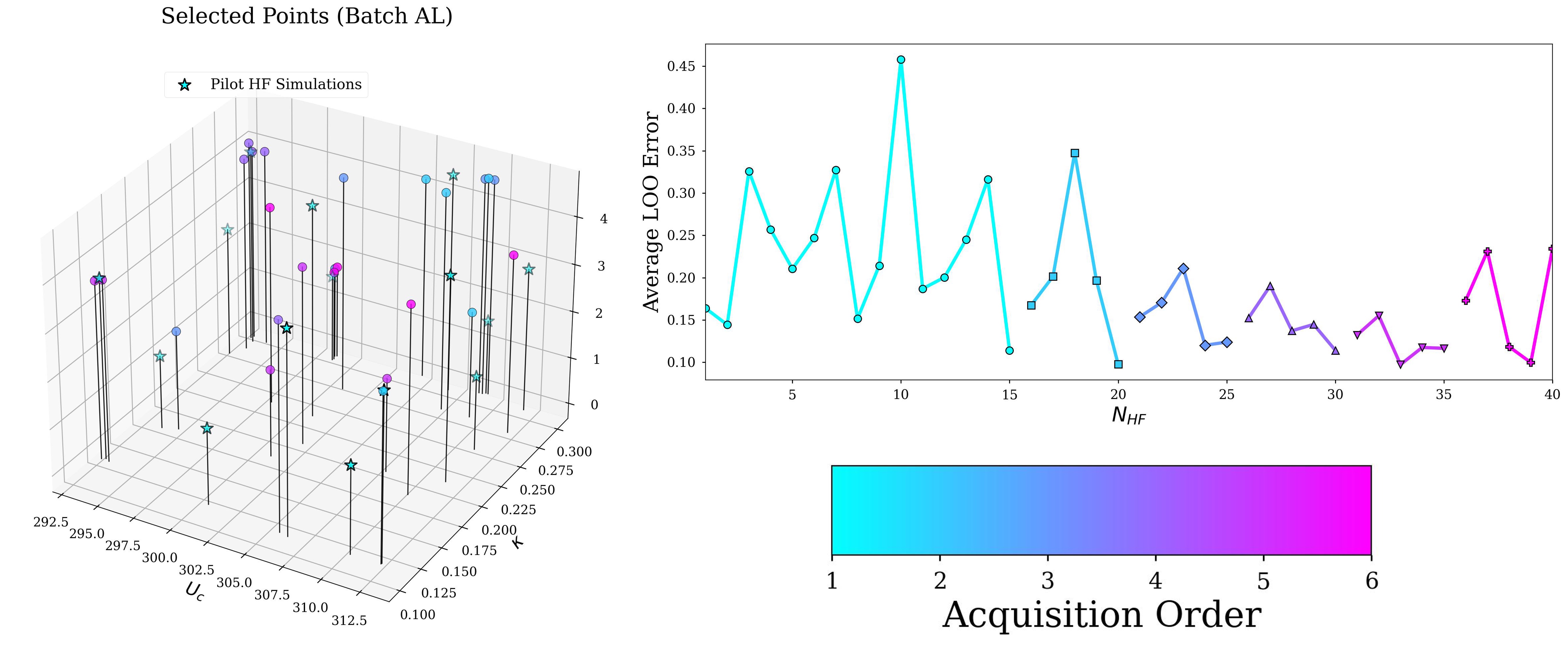}
    \caption{Pilot and newly acquired HF-LF sample locations obtained through active learning (left), with corresponding evolution of the QoI-averaged LOO cross-validation error across acquisition stages (right). The bifidelity surrogate is initialized with 15 pilot HF simulations at stage 1, followed by five batches of 5 new HF simulations at stages 2--6.}
    \label{f:AcqBatchWiseSelected02}
\end{figure}

Figure~\ref{f:AcqBatchWiseSelected02} shows the pilot set and the additional HF-LF simulation pairs acquired during successive active learning batches, color-coded by acquisition stage. The accompanying plot tracks the history of QoI-averaged LOO cross-validation error. The error levels vary across selected points and acquisition stages. The displayed history suggests improvement for many points, with occasional increases; it does not establish monotone improvement or a reduction in variability across the entire input distribution.

\begin{figure}[htbp]
    \centering
    \includegraphics[width=\linewidth,height=0.85\textheight,keepaspectratio]{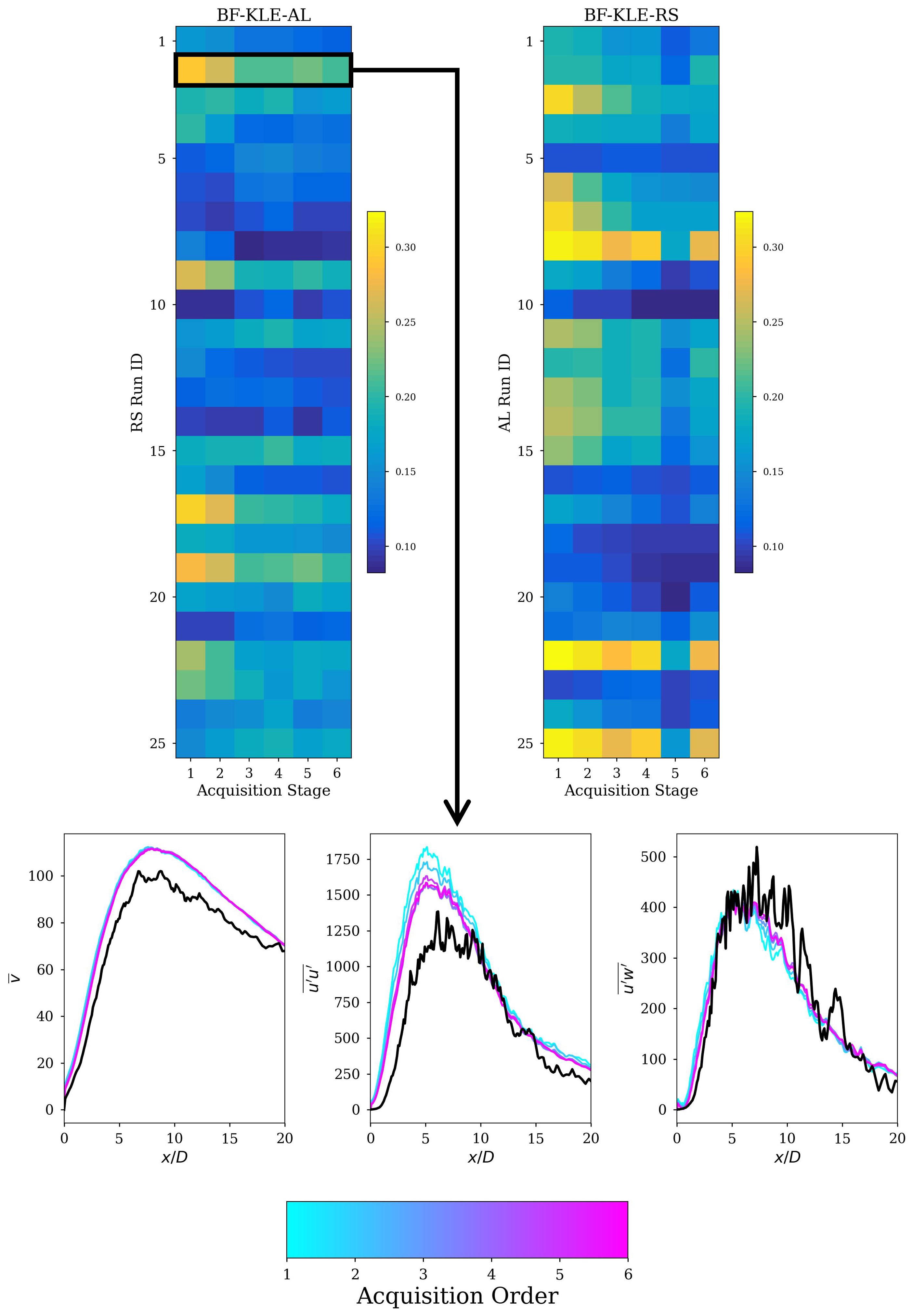}
    \caption{Comparison of relative errors across surrogate training stages (top). Each test set contains the 25 additional HF runs from the competing sampling strategy. Columns 1--6 show performance for the pilot model and five subsequent acquisition stages. BF-KLE-AL surrogate predictions (bottom) are shown for all three QoIs at a representative random parameter setting $\theta$.}
    \label{f:ChangeRelError}
\end{figure}

Figure~\ref{f:ChangeRelError} compares the BF-KLE-AL and BF-KLE-RS surrogates across acquisition stages. Each surrogate is evaluated on the 25 additional HF runs selected by the competing strategy; the 15 shared pilot runs are excluded from these test sets. Stage 1 denotes the pilot surrogate, and stages 2--6 follow the five acquisition batches. The figure uses the same heatmap format as the earlier examples. Errors decrease for many test points as the surrogates are updated, although some points show increases. The AL and RS test sets occupy different input locations, so this diagnostic does not by itself provide a comparison on a common test distribution.

In the bottom part of Figure~\ref{f:ChangeRelError},
we further visualize the QoI predictions from the BF-KLE-AL surrogate for a representative run. Among the three QoIs, the average error driving the EI criterion is most strongly influenced by the Reynolds stress components, particularly $\overline{u'u'}$. This dominance is reflected in the results: BF-KLE-AL rapidly converges to the true HF field for $\overline{u'u'}$ and $\overline{u'w'}$, while improvements in the mean axial velocity $\bar{v}$ are comparatively smaller.

\begin{figure}[htbp]
    \centering
        \includegraphics[width=\linewidth]{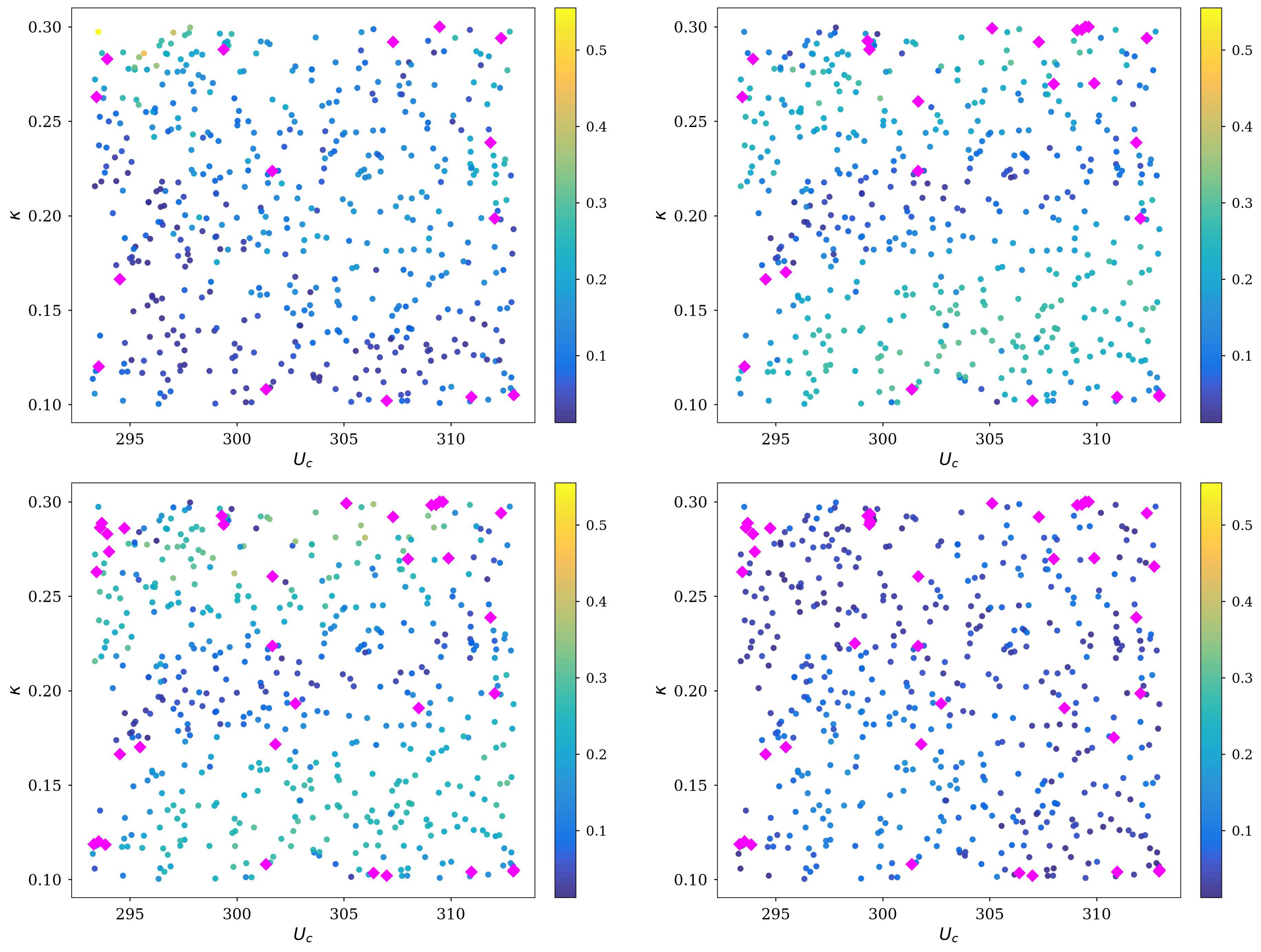}
    \caption{Relative differences in $\overline{u'u'}$ between HF-KLE and BF-KLE-AL surrogate predictions at randomly selected test points, shown at stages 1, 3, 5, and 6. Stage 1 is the pilot model; the other panels report results after 2, 4, and 5 acquisition batches, respectively. The \textcolor{mymagenta}{$\blacklozenge$} markers denote the training points used to build the BF-KLE-AL and HF-KLE at the respective stages (not including the LF-KLE pilot set). Results are visualized in the projected input parameter space of  $U_c$ and $\kappa$.}
    \label{f: relDiffRandomTestPts}
\end{figure}

\begin{figure}[htbp]
    \centering
    \includegraphics[width=\textwidth]{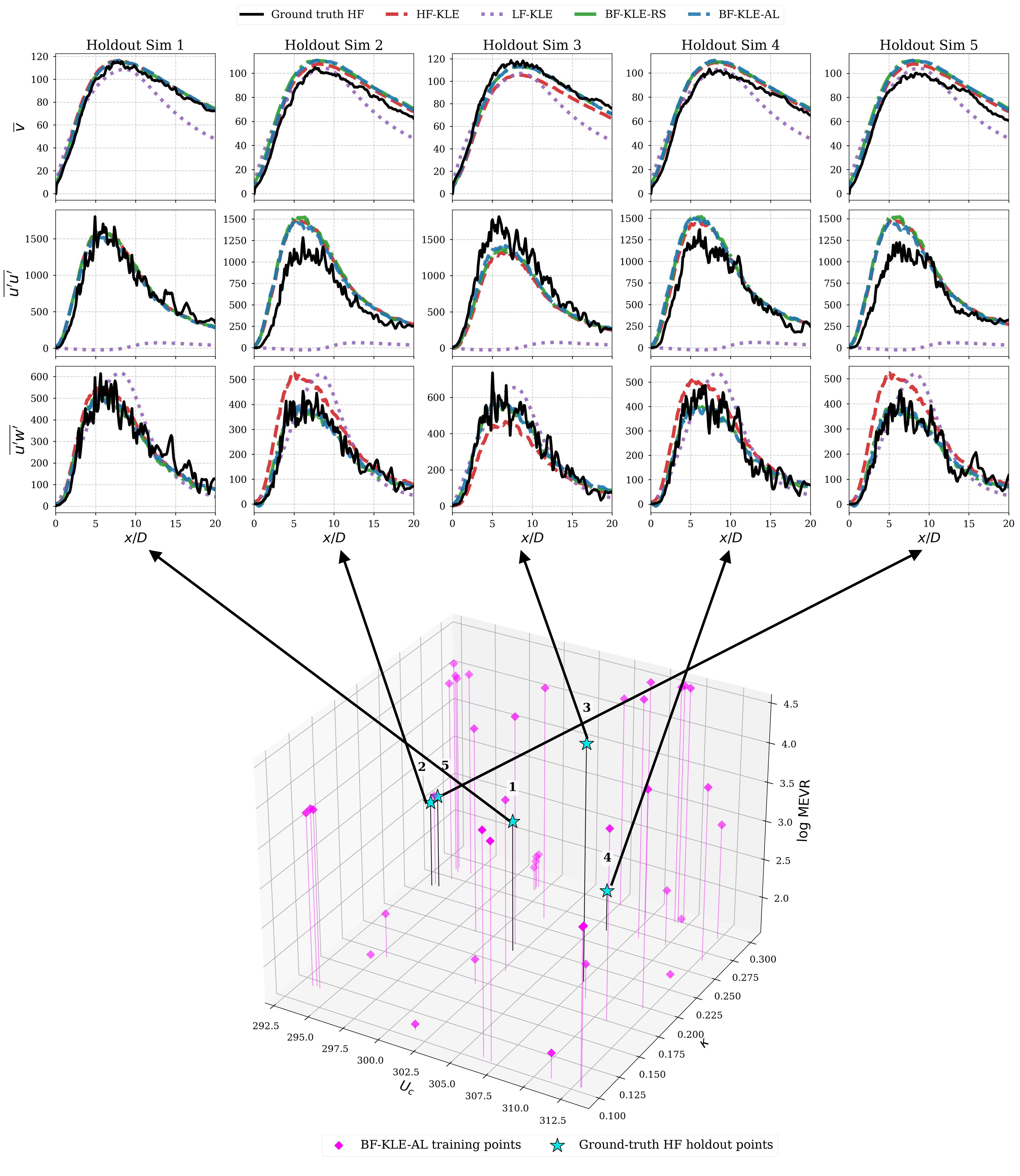}
    \caption{Comparison of bifidelity surrogate and single-fidelity KLE performance against ground-truth HF solutions for  $\bar{v}, \overline{u'u'}$ and $\overline{u'w'}$ visualized in the input parameter space of $U_c$, $\kappa$ and $\log(\tilde{\nu}/\nu)$. \textcolor{mymagenta}{$\blacklozenge$} denote the BF-KLE-AL discrepancy term training points while \textcolor{tabcyan}{$\bigstar$} denote common test points  for all surrogates. The BF-KLE predictions are notably better compared to HF-KLE for $\overline{u'w'}$ and are comparable for the remaining QoIs. The bifidelity surrogates and the HF-KLE are built with the same computational budget.
    }
    \label{f:AL_Random_Jet}
\end{figure}

Figure~\ref{f: relDiffRandomTestPts} reports the relative difference between the BF-KLE-AL and HF-KLE predictions of $\overline{u'u'}$ at 500 random parameter points. The differences generally decrease across the displayed stages, indicating closer agreement between the two surrogates. 
The independent HF comparisons in Figure~\ref{f:AL_Random_Jet} and Table~\ref{t:groundtruthJet} provide the available direct accuracy assessments of BF-KLE-AL and HF-KLE relative to ground truth.

Figure~\ref{f:AL_Random_Jet} and Table~\ref{t:groundtruthJet} compare predictions against five held-out HF simulations. HF-KLE has the lowest reported mean-velocity error (0.0834, versus 0.0856 for BF-KLE-AL and 0.0888 for BF-KLE-RS). For normal Reynolds stress, BF-KLE-AL is slightly better than HF-KLE (0.2490 versus 0.2563), while BF-KLE-RS is worse (0.2689). Both bifidelity surrogates improve the shear-stress prediction (0.1494 and 0.1514, versus 0.2801 for HF-KLE). BF-KLE-AL has smaller reported errors than BF-KLE-RS for all three QoIs, but this AL advantage is modest relative to the bifidelity shear-stress gain. All three HF-informed surrogates outperform LF-KLE on the reported errors. These results suggest some practical benefits of building a bifidelity surrogate and more informative HF sample selection via AL. 

\begin{figure}[htbp]
    \centering
    \includegraphics[width=\linewidth]{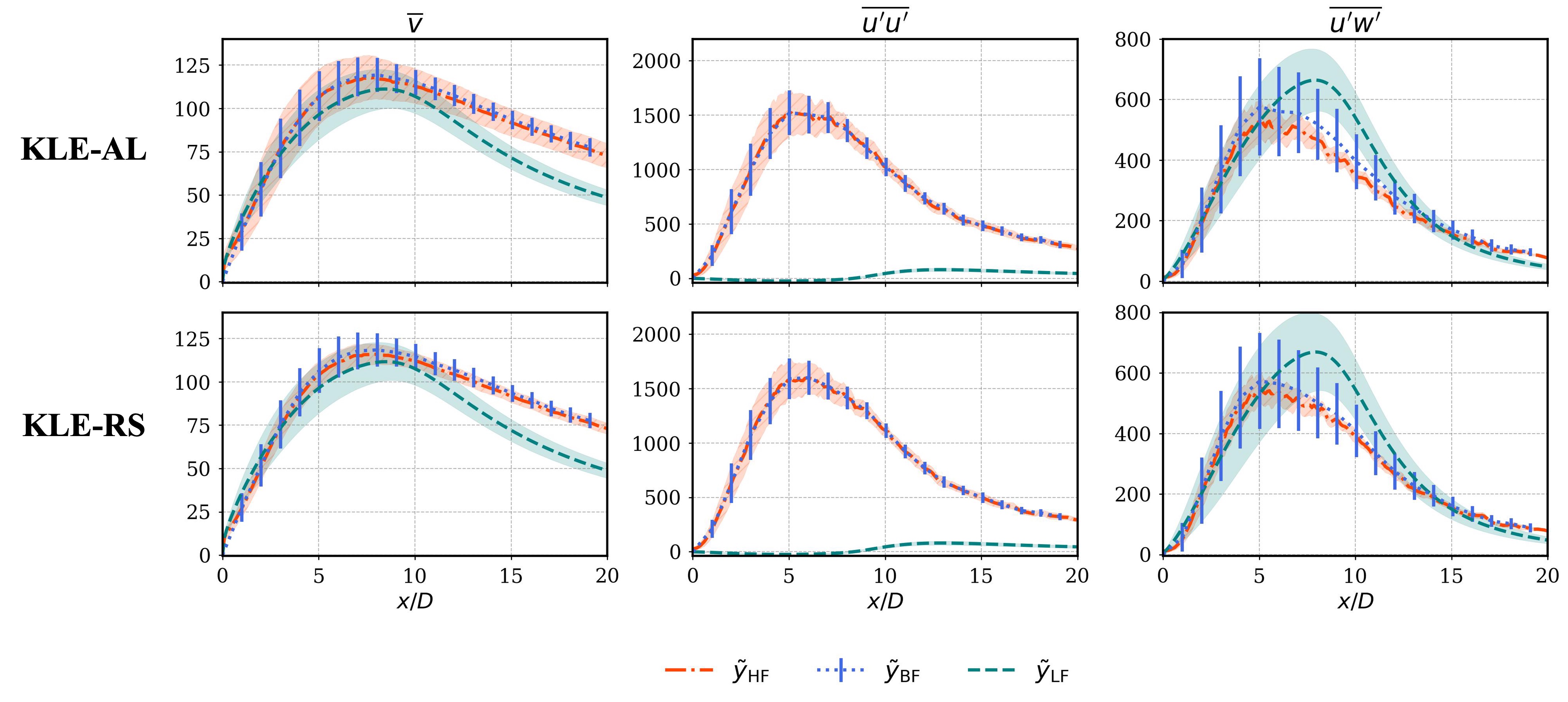}
    \caption{Comparison of forward UQ results: QoI prediction mean $\pm1$ standard deviation for $\tilde{y}_{\mathrm{HF}}$, $\tilde{y}_{\mathrm{BF}}$, and $\tilde{y}_{\mathrm{LF}}$. The bifidelity surrogates $\tilde{y}_{\mathrm{BF}}$ are constructed via active learning (top) and random sampling (bottom) strategies for the three QoIs (columns). 
The BF-KLE predictive statistics (shown via error bars) are closer to those of HF-KLE (intervals marked in orange) than the LF-KLE statistics. 
These intervals describe propagated input variability, but do not independently validate HF distributional statistics.    
}
    \label{f: uq_jet}
\end{figure}

Finally, we show the forward UQ results, comparing propagated-input statistics from the fitted surrogates. 
Figure~\ref{f: uq_jet} presents the comparison between the predictive statistics (mean and $\pm 1$ standard deviation bounds) from BF-KLE, LF-KLE and HF-KLE surrogates built via active learning and random sampling (the LF-KLE and HF-KLE surrogates are constructed solely from the respective single-fidelity simulations obtained when applying each strategy to BF-KLE). As expected from the bias observed in the pilot LF snapshots, especially for $\overline{u'u'}$, the LF-KLE surrogate does not accurately reproduce the mean and variability in the HF-KLE surrogate uncertainty estimates. On the other hand, the mean response and variability from BF-KLE-AL and BF-KLE-RS closely overlap with the respective statistics for HF-KLE-AL and HF-KLE-RS. The BF-KLE-AL standard deviation for $\overline{v}$ is somewhat larger than the HF-KLE-AL result, while the Reynolds-stress bands are closer. 

\begin{table}[htbp]
\centering
\caption{Mean relative error for BF-KLE-AL, BF-KLE-RS, HF-KLE, and LF-KLE, reported over five held-out HF test cases separately for each QoI (lower is better).}\label{t:groundtruthJet}

\begin{tabular}{lllll}
\hline
QoI & BF-KLE-AL & BF-KLE-RS & HF-KLE & LF-KLE \\
\hline
$\overline{v}$      & 0.0856 & 0.0888 & \textbf{0.0834} & 0.1529 \\
$\overline{u'u'}$   & \textbf{0.2490} & 0.2689 & 0.2563 & 0.9887 \\
$\overline{u'w'}$   & \textbf{0.1494} & 0.1514 & 0.2801 & 0.2828 \\
\hline
\end{tabular}
\end{table}

\section{Conclusions}
\label{s:conclusions}

In this work, we introduced a bifidelity KLE surrogate model for field-valued QoIs. The approach combines the expressive power of the KLE, commonly used to represent spatially or temporally correlated fields, with PCEs to maintain an explicit mapping between input uncertainties and output responses. By leveraging inexpensive and well-correlated LF simulations to capture dominant trends with a limited number of HF evaluations to correct systematic bias, the method provides a way to trade LF information and HF corrections against simulation cost. Its benefit depends on the response pair and on how accurately the two surrogate terms can be learned.

To further enhance performance, we developed an active learning strategy that adaptively selects new HF simulations based on the surrogate's estimated generalization error. This error is quantified through cross-validation, modeled using GP regression, and used to drive an EI acquisition function that targets regions of poor surrogate performance. The resulting BF-KLE-AL method was demonstrated on a sequence of test problems of increasing complexity, from ablations on an analytical benchmark to a parametric convection-diffusion equation and finally to a realistic turbulent jet flow simulation, identifying setting-dependent gains and limitations. Active learning improves on random sampling in several tests, but the effect is small or unfavorable in others. The ablations show that a cost-matched HF-only surrogate can be more accurate when LF bias or discrepancy-approximation error is large. In the jet tests, the clearest bifidelity advantage is for shear Reynolds stress, while the mean-velocity comparison favors HF-KLE slightly.

Several limitations and opportunities for future work remain. First, our present investigation focuses on the KLE surrogate structure, and it would be of interest to explore and benchmark bifidelity/multifidelity behavior under alternative architectures such as neural networks or reduced-order models, tailoring the model composition and training strategies accordingly. Within the KLE-PCE representation, analyzing the sources of surrogate error, particularly the approximation of the sample covariance matrix in adaptive settings, would be valuable, as this can guide methods for co-allocation of limited computational resources across different approximations and fidelities. Second, the current framework has been demonstrated for low- to moderate-dimensional uncertainty spaces; extending it to higher dimensions will benefit from dimensionality-reduction techniques such as active subspaces, global sensitivity-based screening, or sparse PCEs to identify and prioritize the most influential parameters. Joint bifidelity modeling with vector-valued KLEs based on cross-component covariances \citep{Cho2013} remains future work.

Finally, the proposed active learning algorithm is myopic, selecting points greedily without accounting for the long-term impact of each acquisition. Developing non-myopic or multi-step lookahead acquisition strategies, potentially informed by Bayesian optimal experimental design principles in lieu of the heuristic used here, could further improve efficiency. Moreover, extending the framework to multifidelity hierarchies with more than two levels would enable adaptive selection of both the sampling location and the fidelity level, thereby balancing information gain against computational cost in a principled, cost-aware manner.

\section*{Acknowledgements}
AJ and XH were partially supported by the National Science Foundation under Grant 2027555.
XH was also partially supported by the Department of Navy awards
N000142312735 and N000142512411 issued by the Office of Naval Research.
CS was supported by the Scientific Discovery through Advanced Computing (SciDAC) program through the FASTMath Institute.

The computational resources provided by the ARCHER2 UK National Supercomputing Service (\url{https://www.archer2.ac.uk}) are gratefully acknowledged. 

Sandia National Laboratories is a multimission laboratory managed and operated by National Technology and Engineering Solutions of Sandia, LLC, a wholly owned subsidiary of Honeywell International, Inc., for the U.S. Department of Energy's National Nuclear Security Administration under contract DE-NA-0003525. This paper describes objective technical results and analysis. Any subjective views or opinions that might be expressed in the paper do not necessarily represent the views of the U.S. Department of Energy or the United States Government.

This article is an expanded version of a paper~\cite{Jivani2021} presented at AIAA SciTech virtually on 10 January 2021. The conference paper is available at \url{https://doi.org/10.2514/6.2021-1367}.

\section*{Declaration of generative AI and AI-assisted technologies in the writing process}
During the preparation of this work, the authors used Gemini and ChatGPT to assist with occasional language editing, organization, and clarity of presentation, as well as refactoring of the research code for public release. All scientific ideas, analyses and conclusions are the authors' own. The authors reviewed and revised all AI-assisted text and take full responsibility for the content of the manuscript.

\bibliographystyle{unsrtnat}

\bibliography{references_v2}

@book{ayachit_paraview_2015,
 address = {Clifton Park, NY},
 author = {Ayachit, Utkarsh},
 edition = {Full color version},
 isbn = {9781930934306},
 language = {eng},
 publisher = {Kitware Inc},
 shorttitle = {The {ParaView} guide},
 title = {The paraview guide: updated for paraview version 4.3},
 year = {2015}
}

@article{azimi_hybrid_2012,
 archiveprefix = {arXiv},
 author = {Azimi, Javad and Jalali, Ali and Fern, Xiaoli},
 doi = {10.48550/arXiv.1202.5597},
 eprint = {1202.5597},
 month = {April},
 note = {arXiv:1202.5597 [cs]},
 publisher = {arXiv},
 title = {Hybrid batch bayesian optimization},
 year = {2012}
}

@inproceedings{balandat2020botorch,
 author = {Balandat, Maximilian and Karrer, Brian and Jiang, Daniel and Daulton, Samuel and Letham, Ben and Wilson, Andrew G and Bakshy, Eytan},
 booktitle = {Advances in Neural Information Processing Systems},
 editor = {H. Larochelle and M. Ranzato and R. Hadsell and M.F. Balcan and H. Lin},
 pages = {21524--21538},
 publisher = {Curran Associates, Inc.},
 title = {Botorch: a framework for efficient monte-carlo bayesian optimization},
 volume = {33},
 year = {2020}
}

@article{bomarito_optimization_2022,
 author = {Bomarito, G.F. and Leser, P.E. and Warner, J.E. and Leser, W.P.},
 doi = {10.1016/j.jcp.2021.110882},
 issn = {00219991},
 journal = {Journal of Computational Physics},
 language = {en},
 title = {On the optimization of approximate control variates with parametrically defined estimators},
 volume = {451},
 year = {2022}
}

@article{Cohn1996,
 author = {Cohn, David A. and Ghahramani, Zoubin and Jordan, Michael I.},
 doi = {10.1613/jair.295},
 eprint = {9603104},
 isbn = {1076-9757},
 issn = {10769757},
 journal = {Journal Artificial Intelligence Research},
 pages = {129--145},
 primaryclass = {cs},
 title = {Active learning with statistical models},
 volume = {4},
 year = {1996}
}

@article{Economon2016,
 author = {Economon, Thomas D. and Palacios, Francisco and Copeland, Sean R. and Lukaczyk, Trent W. and Alonso, Juan J.},
 doi = {10.2514/1.J053813},
 isbn = {0001-1452},
 issn = {0001-1452},
 journal = {AIAA Journal},
 number = {3},
 pages = {828--846},
 title = {Su2: an open-source suite for multiphysics simulation and design},
 volume = {54},
 year = {2016}
}

@article{Ernst2012,
 author = {Ernst, Oliver G. and Mugler, Antje and Starkloff, Hans-J{\"{o}}rg and Ullmann, Elisabeth},
 doi = {10.1051/m2an/2011045},
 issn = {0764-583X},
 journal = {ESAIM: Mathematical Modelling and Numerical Analysis},
 number = {2},
 pages = {317--339},
 title = {On the convergence of generalized polynomial chaos expansions},
 volume = {46},
 year = {2012}
}

@article{fernandez-godino_review_2023,
 author = {Fernández-Godino, M. Giselle},
 doi = {10.3934/acse.2023015},
 journal = {Advances in Computational Science and Engineering},
 language = {en},
 number = {4},
 pages = {351--400},
 title = {Review of multi-fidelity models},
 volume = {1},
 year = {2023}
}

@inproceedings{geraci2015multifidelity,
 author = {Geraci, Gianluca and Eldred, Michael S. and Iaccarino, Gianluca},
 booktitle = {Annual Research Briefs 2015},
 organization = {Center for Turbulence Research},
 pages = {169--181},
 title = {A multifidelity control variate approach for the multilevel monte carlo technique},
 year = {2015}
}

@inproceedings{geraci_2018_asd,
 address = {Glasgow, UK},
 author = {Geraci, Gianluca and Eldred, Michael S. and Gorodetsky, Alex A. and Jakeman, John D.},
 booktitle = {7th {European} {Conference} on {Computational} {Fluid} {Dynamics} ({ECFD} 7)},
 title = {Leveraging active directions for efficient multifidelity uncertainty quantification},
 url = {http://congress.cimne.com/eccm_ecfd2018/admin/files/filePaper/p1481.pdf},
 year = {2018}
}

@techreport{geraci_leveraging_2018,
 author = {Geraci, Gianluca and Eldred, Michael S.},
 doi = {10.2172/1475254},
 institution = {Sandia National Laboratories},
 language = {en},
 month = {September},
 number = {SAND2018-10817, 1475254},
 pages = {SAND2018--10817, 1475254},
 title = {Leveraging intrinsic principal directions for multifidelity uncertainty quantification.},
 urldate = {2022-10-10},
 year = {2018}
}

@book{Ghanem1991,
 address = {New York, NY},
 author = {Ghanem, Roger and Spanos, Pol D.},
 doi = {10.1007/978-1-4612-3094-6},
 edition = {1st},
 publisher = {Springer New York},
 title = {Stochastic finite elements: a spectral approach},
 year = {1991}
}

@incollection{ghanem_multifidelity_2017,
 address = {Cham},
 author = {Eldred, Michael S. and Ng, Leo W. T. and Barone, Matthew F. and Domino, Stefan P.},
 booktitle = {Handbook of {Uncertainty} {Quantification}},
 copyright = {http://www.springer.com/tdm},
 doi = {10.1007/978-3-319-12385-1_25},
 editor = {Ghanem, Roger and Higdon, David and Owhadi, Houman},
 isbn = {9783319123844 9783319123851},
 language = {en},
 pages = {991--1036},
 publisher = {Springer International Publishing},
 title = {Multifidelity uncertainty quantification using spectral stochastic discrepancy models},
 year = {2017}
}

@article{Giles2008,
 author = {Giles, Michael B.},
 doi = {10.1287/opre.1070.0496},
 issn = {0030364X},
 journal = {Operations Research},
 number = {3},
 pages = {607--617},
 title = {Multilevel monte carlo path simulation},
 volume = {56},
 year = {2008}
}

@incollection{ginsbourger_kriging_2010,
 address = {Berlin, Heidelberg},
 author = {Ginsbourger, David and Le Riche, Rodolphe and Carraro, Laurent},
 booktitle = {Computational {Intelligence} in {Expensive} {Optimization} {Problems}},
 doi = {10.1007/978-3-642-10701-6_6},
 editor = {Tenne, Yoel and Goh, Chi-Keong},
 isbn = {9783642107016},
 language = {en},
 pages = {131--162},
 publisher = {Springer},
 series = {Adaptation {Learning} and {Optimization}},
 title = {Kriging is well-suited to parallelize optimization},
 year = {2010}
}

@inproceedings{gonzalez_batch_2016,
 author = {Gonzalez, Javier and Dai, Zhenwen and Hennig, Philipp and Lawrence, Neil},
 booktitle = {Proceedings of the 19th {International} {Conference} on {Artificial} {Intelligence} and {Statistics}},
 language = {en},
 month = {May},
 pages = {648--657},
 publisher = {PMLR},
 title = {Batch bayesian optimization via local penalization},
 year = {2016}
}

@article{Gorodetsky2020,
 author = {Gorodetsky, Alex A. and Geraci, Gianluca and Eldred, Michael S. and Jakeman, John D.},
 doi = {10.1016/j.jcp.2020.109257},
 issn = {10902716},
 journal = {Journal of Computational Physics},
 title = {A generalized approximate control variate framework for multifidelity uncertainty quantification},
 volume = {408},
 year = {2020}
}

@article{gorodetsky_mfnets:_2020,
 author = {Gorodetsky, Alex A. and Jakeman, John D. and Geraci, Gianluca and Eldred, Michael S.},
 doi = {10.1615/Int.J.UncertaintyQuantification.2020032978},
 issn = {2152-5080},
 journal = {International Journal for Uncertainty Quantification},
 language = {en},
 number = {6},
 pages = {595--622},
 shorttitle = {{MFNets}},
 title = {Mfnets: multi-fidelity data-driven networks for bayesian learning and prediction},
 volume = {10},
 year = {2020}
}

@book{gramacy2020surrogates,
 author = {Gramacy, Robert B},
 publisher = {CRC Press},
 title = {Surrogates: gaussian process modeling, design, and optimization for the applied sciences},
 year = {2020}
}

@article{Huan2018b,
 author = {Huan, Xun and Safta, Cosmin and Sargsyan, Khachik and Vane, Zachary P. and Lacaze, Guilhem and Oefelein, Joseph C. and Najm, Habib N.},
 doi = {10.1137/17M1141096},
 issn = {2166-2525},
 journal = {SIAM/ASA Journal on Uncertainty Quantification},
 number = {2},
 pages = {907--936},
 title = {Compressive sensing with cross-validation and stop-sampling for sparse polynomial chaos expansions},
 volume = {6},
 year = {2018}
}

@inproceedings{Jivani2021,
 address = {VIRTUAL EVENT},
 author = {Jivani, Aniket and Huan, Xun and Safta, Cosmin and Zhou, Beckett Y. and Gauger, Nicolas R.},
 booktitle = {{AIAA} {Scitech} 2021 {Forum}},
 doi = {10.2514/6.2021-1367},
 isbn = {9781624106095},
 language = {en},
 publisher = {American Institute of Aeronautics and Astronautics},
 title = {Uncertainty quantification for a turbulent round jet using multifidelity {Karhunen}-{Loève} expansions},
 year = {2021}
}

@article{Jones1998,
 author = {Jones, Donald R. and Schonlau, Matthias and Welch, William J.},
 journal = {Journal of Global Optimization},
 pages = {455--492},
 title = {Efficient global optimization of expensive black-box functions},
 year = {1998}
}

@article{joy_batch_2020,
 author = {Joy, Tinu Theckel and Rana, Santu and Gupta, Sunil and Venkatesh, Svetha},
 doi = {10.1016/j.knosys.2019.06.026},
 issn = {09507051},
 journal = {Knowledge-Based Systems},
 language = {en},
 month = {January},
 pages = {104818},
 title = {Batch bayesian optimization using multi-scale search},
 volume = {187},
 year = {2020}
}

@article{kennedy_predicting_2000,
 author = {Kennedy, M. C. and O'Hagan, A.},
 doi = {10.1093/biomet/87.1.1},
 issn = {00063444, 14643510},
 journal = {Biometrika},
 number = {1},
 pages = {1--13},
 publisher = {[Oxford University Press, Biometrika Trust]},
 title = {Predicting the output from a complex computer code when fast approximations are available},
 volume = {87},
 year = {2000}
}

@book{LeMaitre2010,
 address = {Houten, Netherlands},
 author = {{Le Ma{\^{i}}tre}, Olivier P. and Knio, Omar M.},
 doi = {10.1007/978-90-481-3520-2},
 publisher = {Springer Netherlands},
 title = {Spectral methods for uncertainty quantification: with applications to computational fluid dynamics},
 year = {2010}
}

@incollection{mockett_single-stream_2018_a,
 address = {Cham},
 author = {Fuchs, M. and Mockett, C. and Shur, M. and Strelets, M. and Kok, J. C.},
 booktitle = {{Go4Hybrid}: {Grey} {Area} {Mitigation} for {Hybrid} {RANS}-{LES} {Methods}},
 doi = {10.1007/978-3-319-52995-0_6},
 editor = {Mockett, Charles and Haase, Werner and Schwamborn, Dieter},
 isbn = {9783319529943 9783319529950},
 pages = {125--137},
 publisher = {Springer International Publishing},
 title = {Single-stream round jet at m = 0.9},
 volume = {134},
 year = {2018}
}

@inproceedings{Mockus1974,
 address = {Novosibirsk, Russia},
 author = {Mockus, J.},
 booktitle = {IFIP Technical Conference on Optimization Techniques},
 doi = {10.1007/3-540-07165-2_55},
 isbn = {978-3-540-07165-5},
 pages = {400--404},
 title = {On bayesian methods for seeking the extremum},
 year = {1974}
}

@article{Mueller:2025,
 author = {Mueller, Joy N. and Sargsyan, Khachik and Daniels, Craig J. and Najm, Habib N.},
 doi = {10.1137/23M1613505},
 journal = {SIAM/ASA Journal on Uncertainty Quantification},
 number = {1},
 pages = {1-29},
 title = {Polynomial chaos surrogate construction for random fields with parametric uncertainty},
 volume = {13},
 year = {2025}
}

@article{Najm2009,
 author = {Najm, Habib N.},
 doi = {10.1146/annurev.fluid.010908.165248},
 issn = {0066-4189},
 journal = {Annual Review of Fluid Mechanics},
 number = {1},
 pages = {35--52},
 publisher = {Annual Reviews},
 title = {Uncertainty quantification and polynomial chaos techniques in computational fluid dynamics},
 volume = {41},
 year = {2009}
}

@inproceedings{ng_multifidelity_2012,
 address = {Honolulu, Hawaii},
 author = {Ng, Leo Wai-Tsun and Eldred, Michael},
 booktitle = {53rd {AIAA}/{ASME}/{ASCE}/{AHS}/{ASC} {Structures}, {Structural} {Dynamics} and {Materials} {Conference}; 20th {AIAA}/{ASME}/{AHS} {Adaptive} {Structures} {Conference}; 14th {AIAA}},
 doi = {10.2514/6.2012-1852},
 isbn = {9781600869372},
 language = {en},
 publisher = {American Institute of Aeronautics and Astronautics},
 title = {Multifidelity uncertainty quantification using non-intrusive polynomial chaos and stochastic collocation},
 year = {2012}
}

@article{peherstorfer_optimal_2016,
 author = {Peherstorfer, Benjamin and Willcox, Karen and Gunzburger, Max},
 doi = {10.1137/15M1046472},
 issn = {1064-8275, 1095-7197},
 journal = {SIAM Journal on Scientific Computing},
 language = {en},
 number = {5},
 pages = {A3163--A3194},
 title = {Optimal model management for multifidelity monte carlo estimation},
 volume = {38},
 year = {2016}
}

@article{peherstorfer_survey_2018,
 author = {Peherstorfer, Benjamin and Willcox, Karen and Gunzburger, Max},
 doi = {10.1137/16M1082469},
 issn = {0036-1445, 1095-7200},
 journal = {SIAM Review},
 language = {en},
 number = {3},
 pages = {550--591},
 title = {Survey of multifidelity methods in uncertainty propagation, inference, and optimization},
 volume = {60},
 year = {2018}
}

@book{Rasmussen2006,
 address = {Cambridge, MA},
 author = {Rasmussen, Carl Edward and Williams, Christopher K. I.},
 publisher = {The MIT Press},
 title = {Gaussian processes for machine learning},
 year = {2006}
}

@book{Settles2012,
 address = {San Rafael, CA},
 author = {Settles, Burr},
 doi = {10.2200/S00429ED1V01Y201207AIM018},
 publisher = {Morgan {\&} Claypool Publishers},
 title = {Active learning},
 year = {2012}
}

@article{stroethoff_bhaskaras_2014,
 author = {Stroethoff, Karel},
 doi = {10.54870/1551-3440.1313},
 issn = {1551-3440},
 journal = {The Mathematics Enthusiast},
 language = {en},
 month = {December},
 number = {3},
 pages = {485--494},
 title = {Bhaskara’s approximation for the sine},
 volume = {11},
 year = {2014}
}

@article{Xiu2002,
 author = {Xiu, Dongbin and Karniadakis, George Em},
 doi = {10.1137/S1064827501387826},
 issn = {1064-8275},
 journal = {SIAM Journal on Scientific Computing},
 number = {2},
 pages = {619--644},
 title = {The wiener-askey polynomial chaos for stochastic differential equations},
 volume = {24},
 year = {2002}
}

@article{Xiu2009,
 author = {Xiu, Dongbin},
 issn = {1815-2406},
 journal = {Communications in Computational Physics},
 number = {2-4},
 pages = {242--272},
 title = {Fast numerical methods for stochastic computations: a review},
 volume = {5},
 year = {2009}
}

@article{zhang_multifidelity_2018,
 author = {Zhang, Yiming and Kim, Nam H. and Park, Chanyoung and Haftka, Raphael T.},
 doi = {10.2514/1.J057299},
 issn = {0001-1452, 1533-385X},
 journal = {AIAA Journal},
 language = {en},
 number = {12},
 pages = {4944--4952},
 title = {Multifidelity surrogate based on single linear regression},
 volume = {56},
 year = {2018}
}

@misc{safta_numerical_2026,
	title = {Numerical {Considerations} for the {Construction} of {Karhunen}-{Loève} {Expansions}},
	doi = {10.48550/arXiv.2603.19108},
	publisher = {arXiv},
	author = {Safta, Cosmin and Najm, Habib N.},
	year = {2026},
}

@inproceedings{sella_projection-based_2023,
	address = {National Harbor, MD \& Online},
	title = {Projection-based multifidelity linear regression for data-poor applications},
	isbn = {9781624106996},
	doi = {10.2514/6.2023-0916},
	language = {en},
	booktitle = {{AIAA} {SCITECH} 2023 {Forum}},
	publisher = {American Institute of Aeronautics and Astronautics},
	author = {Sella, Vignesh and Pham, Julie and Chaudhuri, Anirban and Willcox, Karen E.},
	month = jan,
	year = {2023},
}

@inproceedings{spalart_one-equation_1992,
	address = {Reno,NV,U.S.A.},
	title = {A one-equation turbulence model for aerodynamic flows},
	doi = {10.2514/6.1992-439},
	language = {en},
	booktitle = {30th {Aerospace} {Sciences} {Meeting} and {Exhibit}},
	publisher = {American Institute of Aeronautics and Astronautics},
	author = {Spalart, P. and Allmaras, S.},
	year = {1992},
}

@InProceedings{pmlr-v51-shahriari16,
  title = 	 {Unbounded Bayesian Optimization via Regularization},
  author = 	 {Shahriari, Bobak and Bouchard-Cote, Alexandre and Freitas, Nando},
  booktitle = 	 {Proceedings of the 19th International Conference on Artificial Intelligence and Statistics},
  pages = 	 {1168--1176},
  year = 	 {2016},
  volume = 	 {51},
  series = 	 {Proceedings of Machine Learning Research},
  address = 	 {Cadiz, Spain},
  month = 	 {09--11 May},
  publisher =    {PMLR}
}

@inproceedings{Rainforth2016,
 author = {Rainforth, Tom and Le, Tuan Anh and van de Meent, Jan-Willem and Osborne, Michael A and Wood, Frank},
 booktitle = {Advances in Neural Information Processing Systems},
 pages = {},
 publisher = {Curran Associates, Inc.},
 title = {Bayesian Optimization for Probabilistic Programs},
 volume = {29},
 year = {2016}
}

@article{berk_bayesian_2020,
	title = {Bayesian optimisation in unknown bounded search domains},
	volume = {195},
	issn = {09507051},
	doi = {10.1016/j.knosys.2020.105645},
	language = {en},
	journal = {Knowledge-Based Systems},
	author = {Berk, Julian and Gupta, Sunil and Rana, Santu and Nguyen, Vu and Venkatesh, Svetha},
	month = may,
	year = {2020},
	pages = {105645},
}

@article{howard_multifidelity_2023,
	title = {Multifidelity deep operator networks for data-driven and physics-informed problems},
	volume = {493},
	issn = {00219991},
	doi = {10.1016/j.jcp.2023.112462},
	language = {en},
	journal = {Journal of Computational Physics},
	author = {Howard, Amanda A. and Perego, Mauro and Karniadakis, George Em and Stinis, Panos},
	month = nov,
	year = {2023},
	pages = {112462},
}

@article{lu_learning_2021,
	title = {Learning nonlinear operators via {DeepONet} based on the universal approximation theorem of operators},
	volume = {3},
	issn = {2522-5839},
	doi = {10.1038/s42256-021-00302-5},
	language = {en},
	number = {3},
	journal = {Nature Machine Intelligence},
	author = {Lu, Lu and Jin, Pengzhan and Pang, Guofei and Zhang, Zhongqiang and Karniadakis, George Em},
	year = {2021},
	pages = {218--229},
}

@incollection{BOULLE202483,
title = {Chapter 3 - A mathematical guide to operator learning},
series = {Handbook of Numerical Analysis},
publisher = {Elsevier},
volume = {25},
pages = {83-125},
year = {2024},
booktitle = {Numerical Analysis Meets Machine Learning},
issn = {1570-8659},
doi = {10.1016/bs.hna.2024.05.003},
author = {Nicolas Boullé and Alex Townsend}
}

@article{betz_numerical_2014,
	title = {Numerical methods for the discretization of random fields by means of the {K}arhunen-{L}o\`{e}ve expansion},
	volume = {271},
	issn = {00457825},
	doi = {10.1016/j.cma.2013.12.010},
	language = {en},
	journal = {Computer Methods in Applied Mechanics and Engineering},
	author = {Betz, Wolfgang and Papaioannou, Iason and Straub, Daniel},
	year = {2014},
	pages = {109--129},
}

@article{narayan_stochastic_2014,
	title = {A {Stochastic} {Collocation} {Algorithm} with {Multifidelity} {Models}},
	volume = {36},
	issn = {1064-8275, 1095-7197},
	doi = {10.1137/130929461},
	language = {en},
	number = {2},
	journal = {SIAM Journal on Scientific Computing},
	author = {Narayan, Akil and Gittelson, Claude and Xiu, Dongbin},
	year = {2014},
	pages = {A495--A521},
}

@article{Cho2013,
  author = {Cho, H. and Venturi, D. and Karniadakis, G. E.},
  title = {{Karhunen--Lo{\`e}ve} expansion for multi-correlated stochastic processes},
  journal = {Probabilistic Engineering Mechanics},
  volume = {34},
  pages = {157--167},
  year = {2013},
  doi = {10.1016/j.probengmech.2013.09.004}
}

\clearpage
\appendix
\section{Additional results for Examples 1a and 1b (known analytical form of the HF model)}
\label{app:ex-1b-hf-lf-families}

\begin{figure}[htbp]
    \centering
    \includegraphics[width=\linewidth]{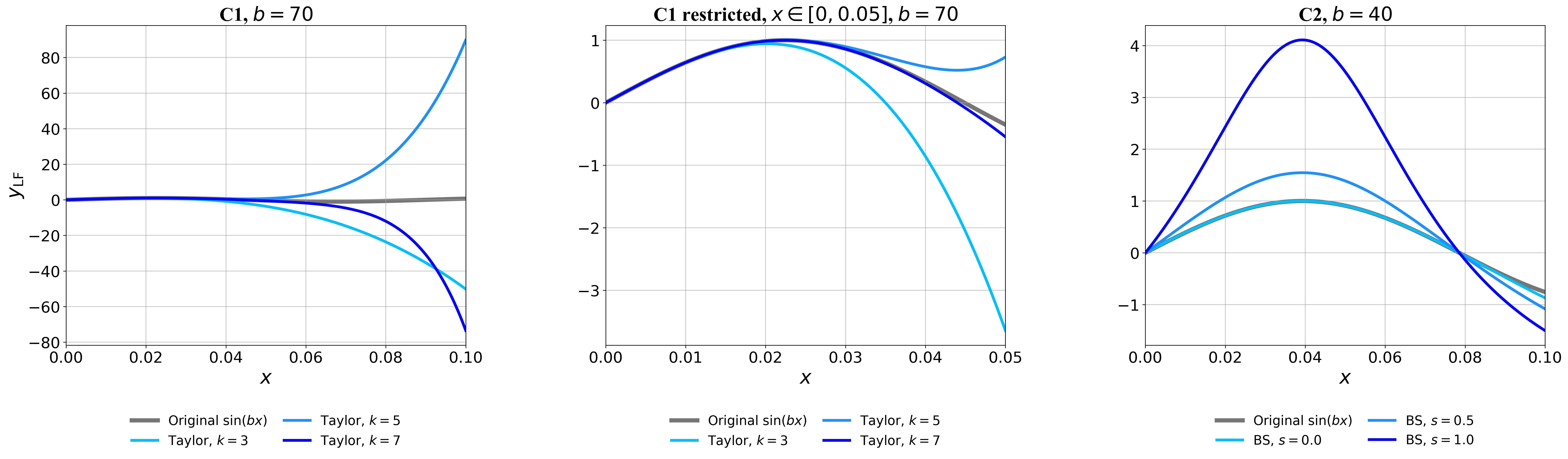}
    \caption{Approximation of $y_{\textrm{LF}}$ for C1 (varying Taylor expansion order $k$) and C2 (varying $s$ via Equation~\ref{eq:stroethoff_c2_modified_s_tuning}).}
    \label{fig:1d_bsine_all_s}
\end{figure}

Figure~\ref{fig:1d_bsine_all_s} shows various LF approximations to the sine term for the cases reported in Section~\ref{ss: ex1-1d-part2}.

\begin{figure}[htbp]
    \centering
    \includegraphics[width=0.9\linewidth]{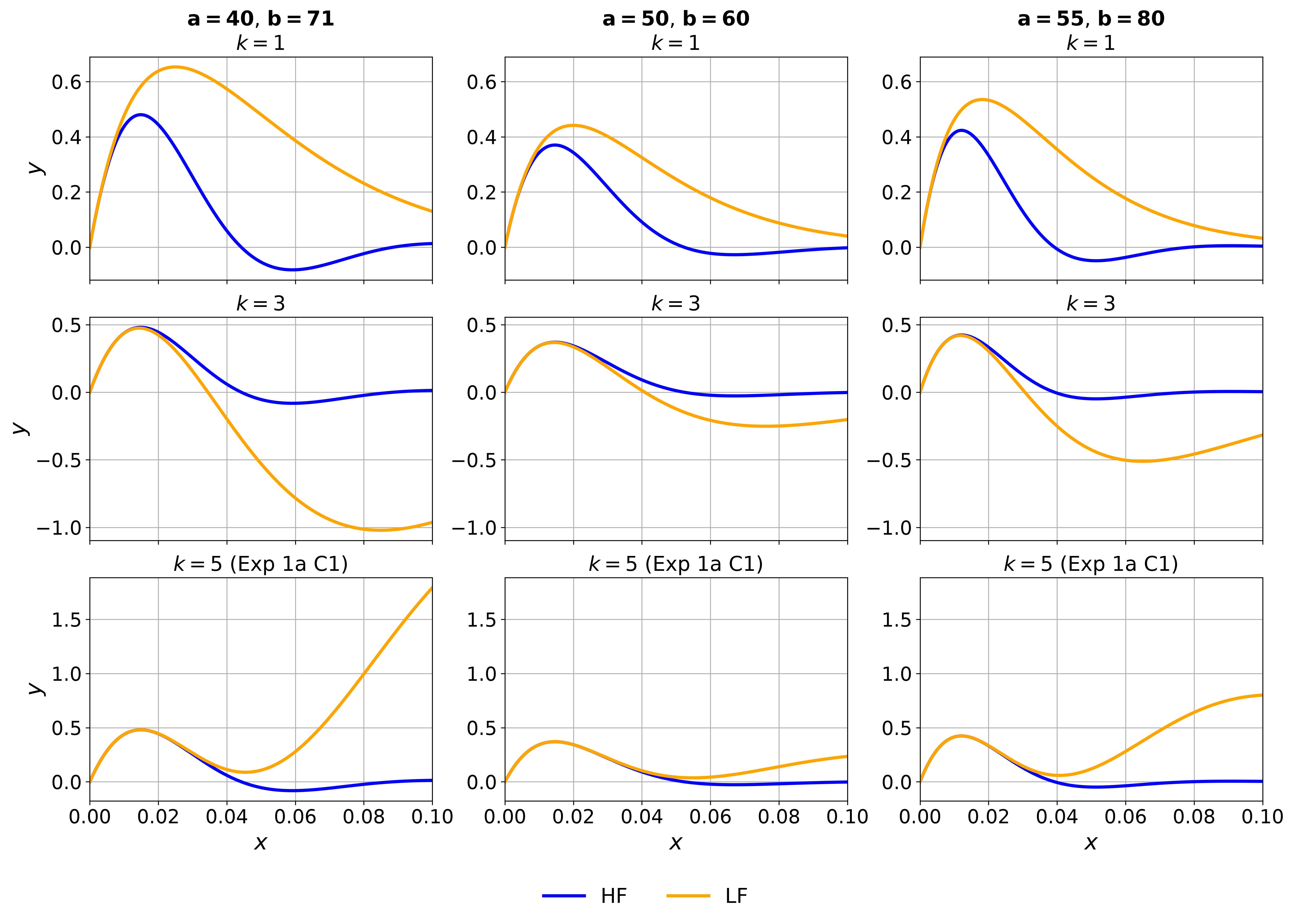}
    \caption{Evaluations for $y_{\textrm{LF}}$ and $y_{\textrm{HF}}$ at selected values of $a$ and $b$ for C1 variants.}
    \label{fig:1d_samples_lf_hf_c1_additional}
\end{figure}

\begin{figure}[htbp]
    \centering
    \includegraphics[width=0.9\linewidth]{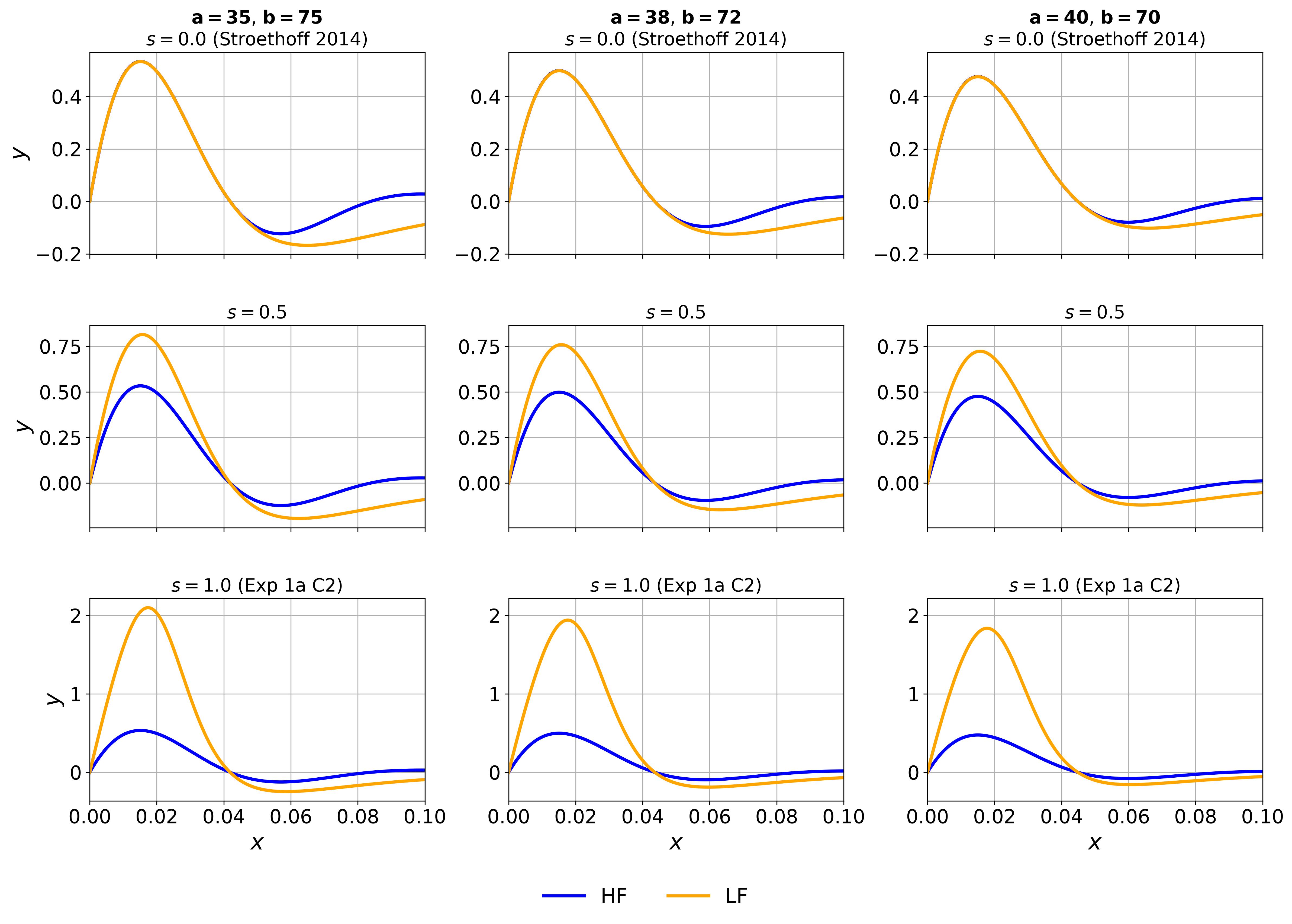}
    \caption{Evaluations for $y_{\textrm{LF}}$ and $y_{\textrm{HF}}$ at selected values of $a$ and $b$ for C2 variants.}
    \label{fig:1d_samples_lf_hf_c2_additional}
\end{figure}

Figures~\ref{fig:1d_samples_lf_hf_c1_additional} and~\ref{fig:1d_samples_lf_hf_c2_additional} display the evaluations for the HF and LF models at selected values of $\theta$ for the above cases.

\begin{figure}[htbp]
    \centering
    \includegraphics[width=0.9\linewidth]{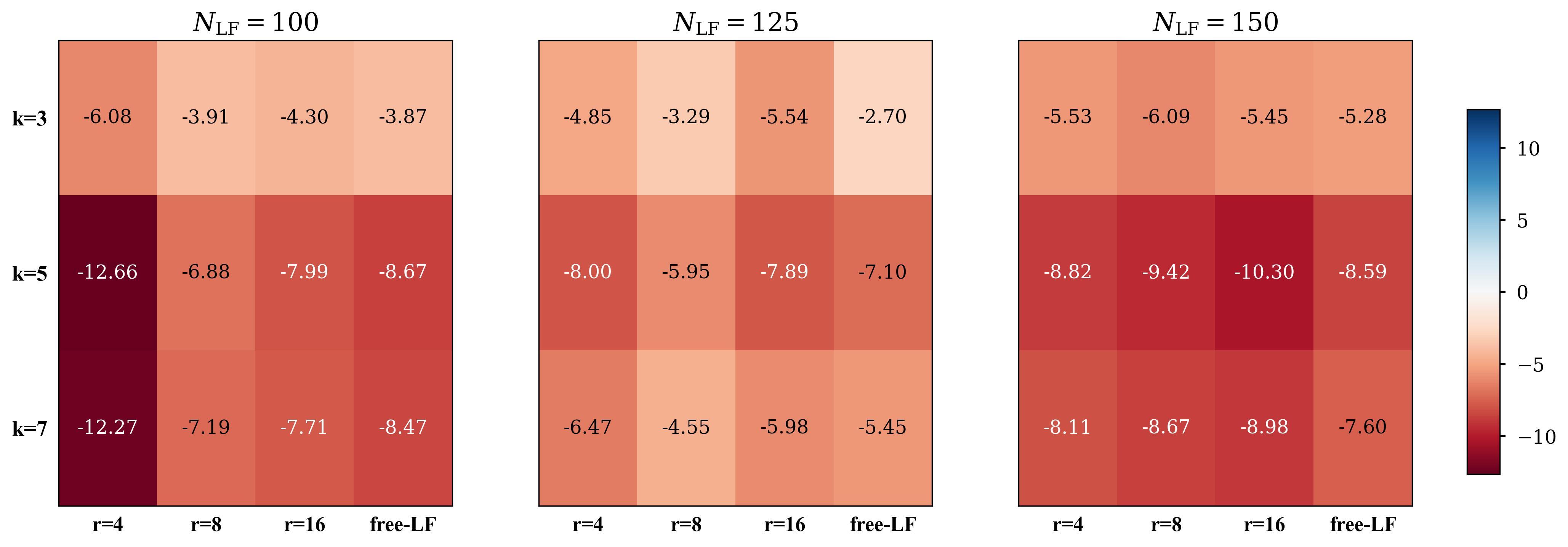}
    \caption{Mean final-stage relative BF-KLE advantage for C1, computed over 20 independent replicates at various ablation settings.}
    \label{fig:heatmaps_all_c1}
\end{figure}

\begin{figure}[htbp]
    \centering
    \includegraphics[width=0.9\linewidth]{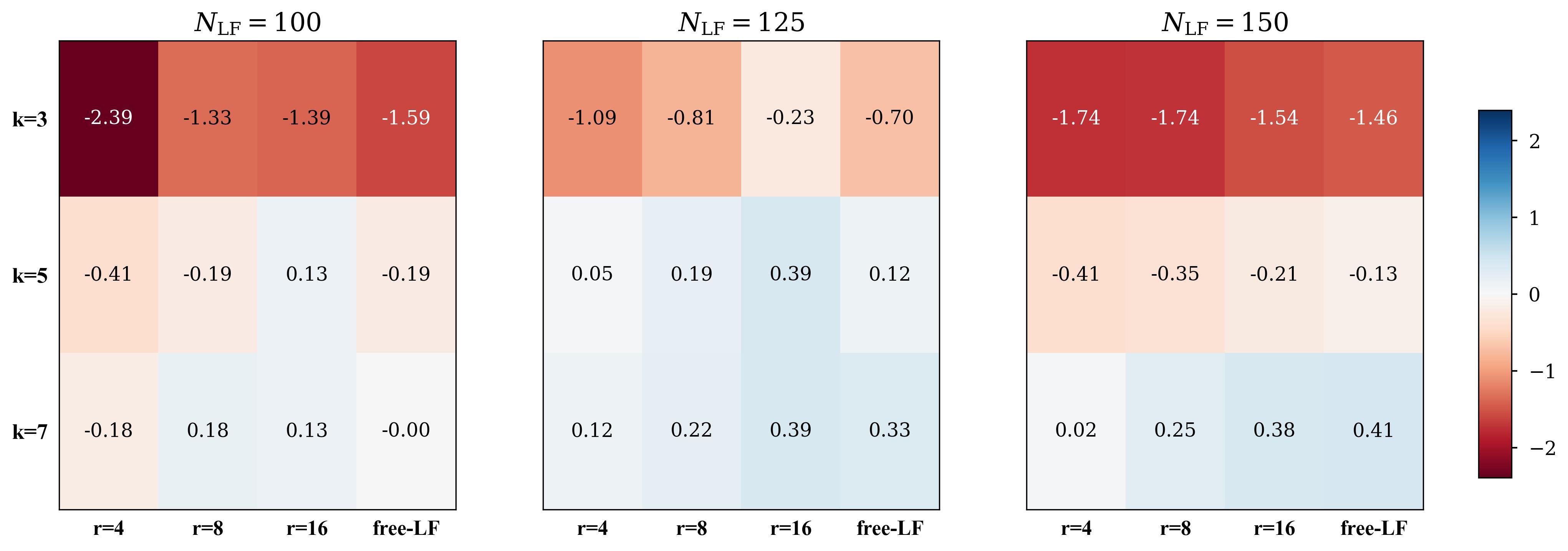}
    \caption{Mean final-stage relative BF-KLE advantage for C1 with HF and LF models defined over $x \in [0, 0.05]$, computed over 20 independent replicates at various ablation settings.}
    \label{fig:heatmaps_all_c1_restricted}
\end{figure}

\begin{figure}[htbp]
    \centering
    \includegraphics[width=0.9\linewidth]{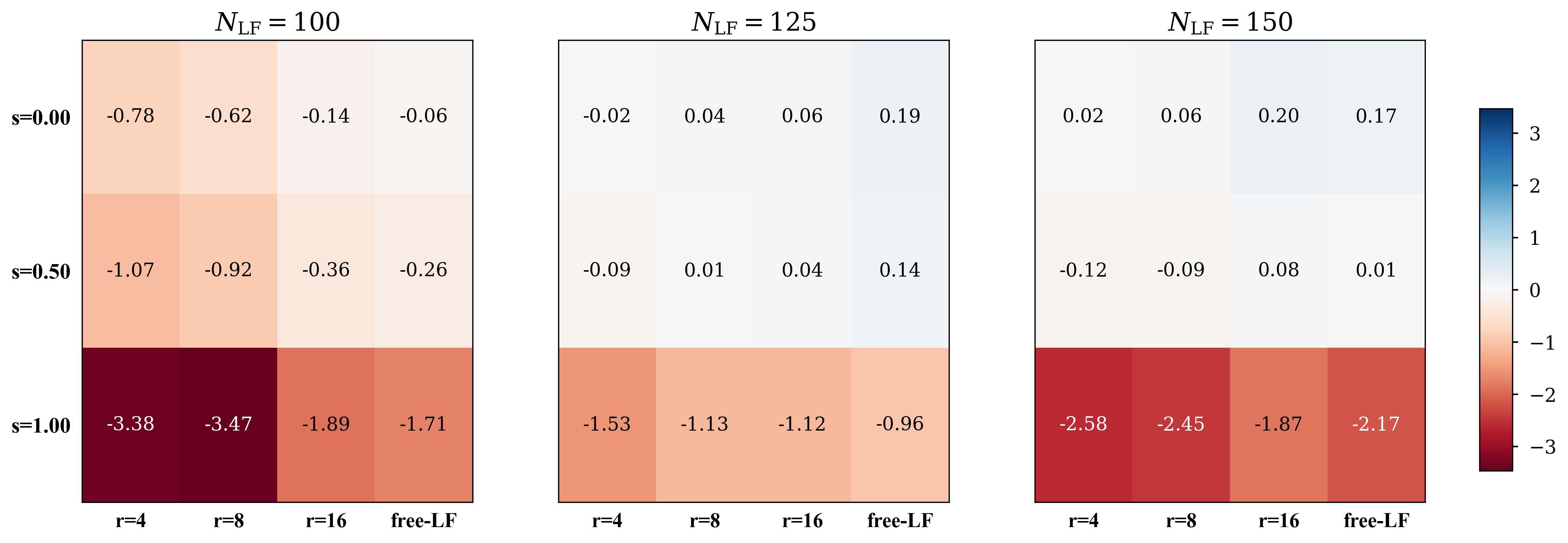}
    \caption{Mean final-stage relative BF-KLE advantage for C2, computed over 20 independent replicates at various ablation settings.}
    \label{fig:heatmaps_all_c2}
\end{figure}

Figures~\ref{fig:heatmaps_all_c1}, \ref{fig:heatmaps_all_c1_restricted}, and \ref{fig:heatmaps_all_c2} show the mean (over 20 replicates) final-stage relative BF-KLE advantage $A_{\textrm{BF}}$ for C1, C1 (restricted), and C2, respectively. Each cell is colored by $A_{\textrm{BF}}=\frac{1}{20}\sum_{m=1}^{20} (E_{\textrm{HF}}^{(m)}-E_{\textrm{BF}}^{(m)})/E_{\textrm{HF}}^{(m)}$ where $E_{\textrm{BF}}^{(m)}$ and $E_{\textrm{HF}}^{(m)}$ are the integrated BF-KLE and HF-KLE relative errors at $N_{\Delta}=60$. Positive values (in blue) indicate regimes where building BF-KLE is advantageous, while negative ones (in red) favor the single-fidelity surrogate.

\begin{figure}[htbp]
    \centering
    \includegraphics[width=\linewidth]{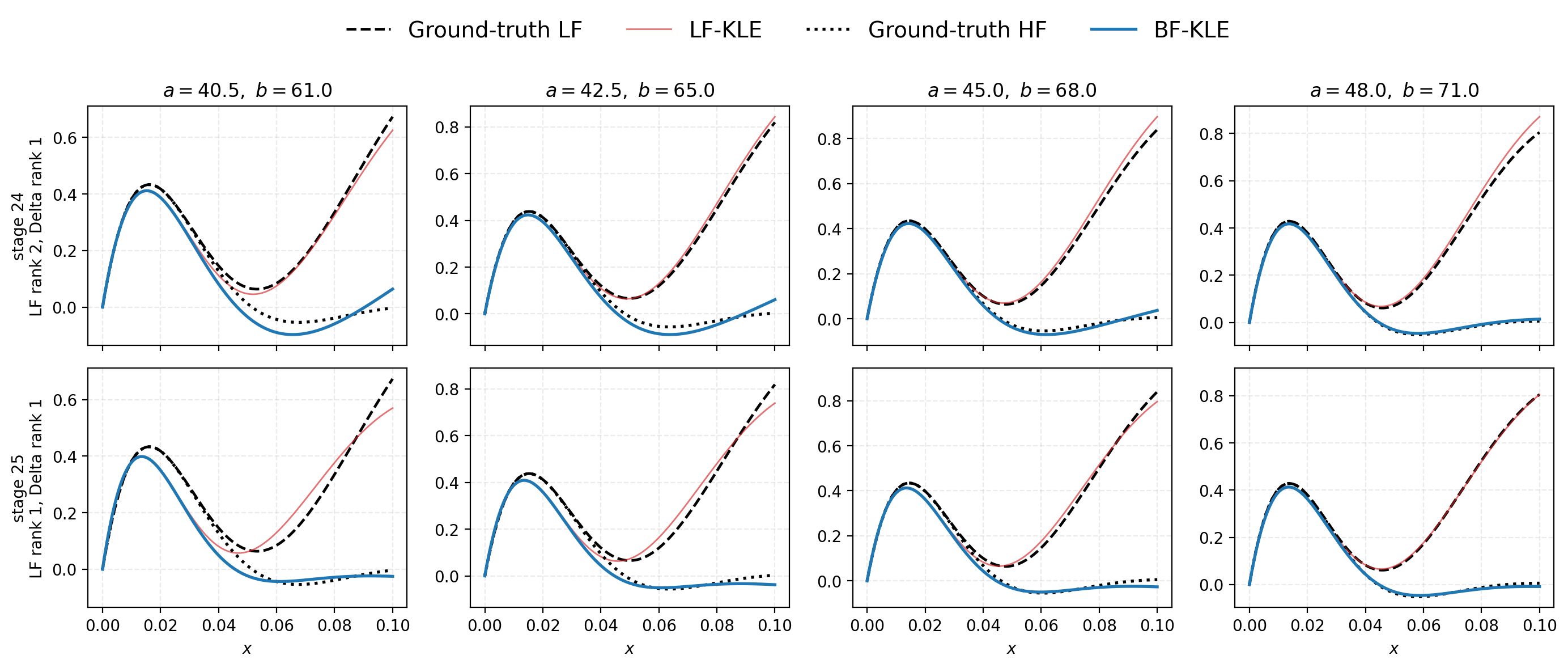}
    \caption{Illustration of fluctuations in LF-KLE and BF-KLE accuracy for some realizations of $\theta=(a, b)$ (leftmost panel, $a=40.5, b=61$) as the LF rank changes between stages.}
    \label{fig:rank_fluctuations_lf_bf_predictions}
\end{figure}

\begin{figure}[htbp]
    \centering
    \includegraphics[width=\linewidth]{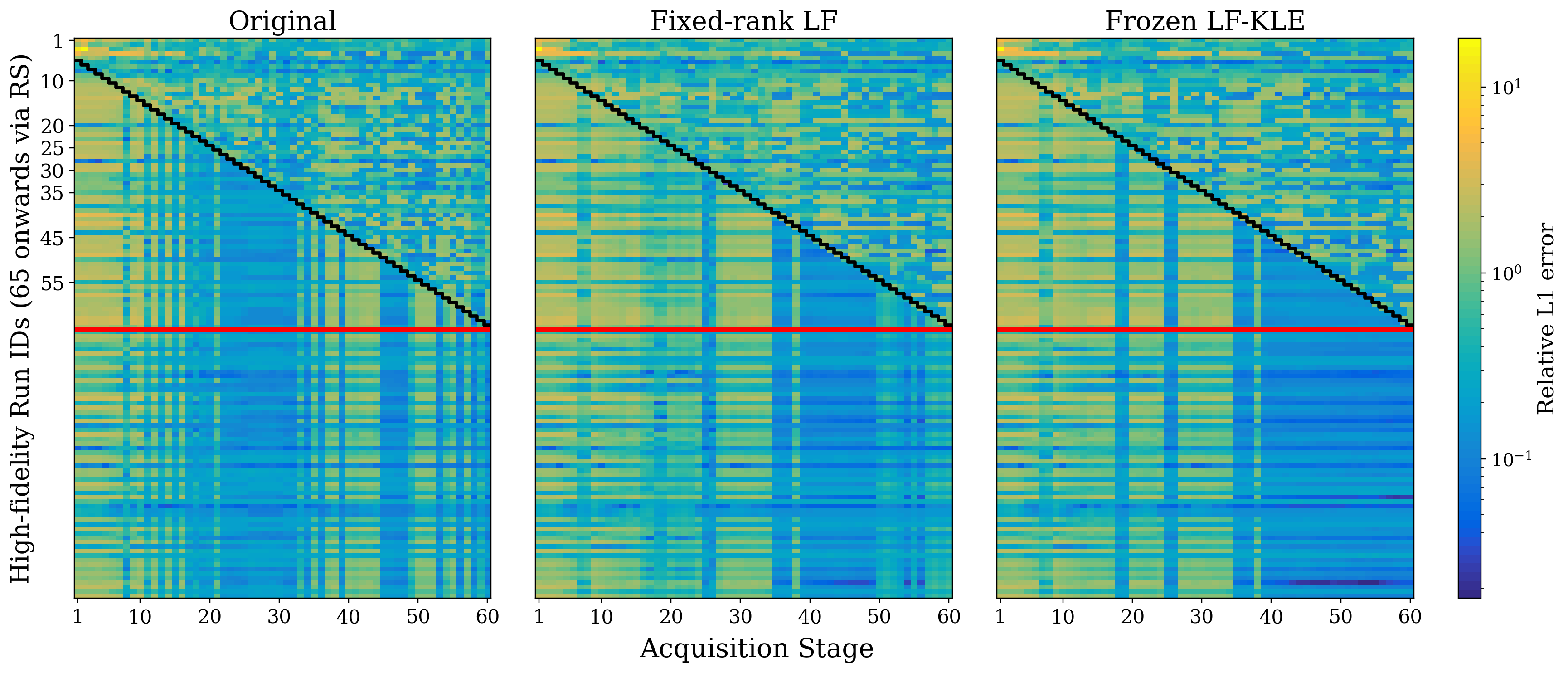}
    \caption{Relative errors for all surrogate training points in a single replicate for C1 using BF-KLE-AL. (left) reproduced from Figure~\ref{fig:err_1d_single_rep_updated_L1}, (middle) LF rank set to 2 for all acquisition stages and (right) LF-KLE is built once from the pilot dataset (rank 2) and not updated as additional paired HF-LF simulations are acquired.  Entries below the red separator indicate predictive performance on a small test set comprising points from the competing strategy (BF-KLE-RS).}
    \label{fig:stab_improvements}
\end{figure}

In Figure~\ref{fig:err_1d_single_rep_updated_L1} showing the evolution of relative error for a fixed test set, the errors on some realizations of $\theta$ in the held-out test set can increase noticeably from one stage to the next, especially for C1. A potential cause of this instability may be fluctuations in the number of retained modes in the truncated LF-KLE. An example of this phenomenon can be seen for one of the selected test predictions in Figure~\ref{fig:rank_fluctuations_lf_bf_predictions}. Simple changes to the BF-KLE-AL setup, such as adopting a fixed value of $k_t$ for all acquisition stages based on a reference value computed from the pilot dataset, or keeping the LF-KLE unchanged during the AL-based surrogate refinement, both appear to be reasonable heuristics to alleviate the instabilities (Figure~\ref{fig:stab_improvements}). Hyperparameters such as the PC expansion order, truncation scheme, and batch size for AL may also influence the overall effectiveness of BF-KLE-AL. These single-replicate diagnostics suggest practical mitigations, but do not guarantee monotone prediction accuracy. Future extensions of this work could instead consider more principled early-stage stopping criteria based on a separate validation set, as described in Section~\ref{ss: al-bf-kle}.

\clearpage
\section{Comparison with Multifidelity Operator Networks}
\label{app:deeponet_comparisons}

In this section, we compare the predictive accuracy of BF-KLE and a multifidelity DeepONet surrogate model (referred to as BF-DeepONet in results) on a one-dimensional jump function problem with a ground-truth HF solution available \citep{howard_multifidelity_2023}.

Briefly, let $\CU$ and $\CS$ be infinite-dimensional input and output function spaces, respectively, and $\CG^{\dagger}: \CU \to \CS$ denote the unknown, possibly nonlinear operator mapping between these. Given input-output training pairs $(u, s)$, where $u \in \CU$, $s = \CG^\dagger(u) \in \CS$, the goal of operator learning is to learn an accurate finite-dimensional approximation $\CG_\phi$ parameterized by $\phi$. A variety of linear and nonlinear surrogate representations, including neural networks, can be employed for this task. Operator learning and related surrogate modeling techniques are surveyed in \cite{BOULLE202483}.

DeepONets \citep{lu_learning_2021} are a popular neural operator architecture applied to a wide range of applications. A single-fidelity DeepONet consists of two simultaneously trained neural networks (branch and trunk). The branch network accepts evaluations of the input function $u$ discretized at a fixed number of spatial and/or temporal locations $x \in \CX \subseteq \RR^n$ and returns $b \in \RR^Q$. The trunk network can accept arbitrary query locations $x^{\ast} \in \CX \subseteq \RR^n$, returning $t \in \RR^Q$. A trained DeepONet generates predictions via $ s(x^\ast) = \CG_\phi(u)(x^\ast) = \sum_{k} b_kt_k$.

The multifidelity version of this architecture trains three DeepONet subnets jointly. The low-fidelity surrogate component is captured through an LF subnet while the additive discrepancy term between low-fidelity and high-fidelity models is represented through a combination of linear and nonlinear correlation trends obtained from separate subnets, each accepting the LF subnet predictions as input. The architecture in \citet{howard_multifidelity_2023} can query the LF subnet at the HF observation locations and accommodate different observation grids. The BF-KLE implementation here aligns field data on a common grid before constructing the discrepancy. The benchmark below uses common-grid data for both methods, so it does not test their behavior with differing grids. \citet{howard_multifidelity_2023} provides additional details on the modeling of the discrepancy term,  training of the BF-DeepONet architecture and hyperparameter selection for various benchmarks.

The BF-KLE implementation studied here maps a finite-dimensional uncertain input to a scalar field. Its input representation therefore differs from the general function-to-function setting of operator learning. 
However, in special cases, where the input function $u$ can be further parameterized via a small number of uncertain physical parameters $\theta$, the DeepONet setup described above can be expressed as a bifidelity surrogate approximation to an HF random field, similar to the form of Equation~\ref{eq:hf_approx_lf_disc}. 
This enables limited comparisons of these techniques on suitable benchmark problems with data generated on a common grid.

The LF and HF jump function models are defined as follows:
\begin{align}
y_{\textrm{LF}}(x, \theta) &= \begin{cases}
0.5(6x - 2)^2 \sin(\theta x - 4) + 10(x - 0.5) - 5 & x \le 0.5 \\
0.5(6x - 2)^2 \sin(\theta x - 4) + 10(x - 0.5) - 2 & x > 0.5
\end{cases}
\label{e:ylf_eqn_deeponet}
\end{align}
and
\begin{align}
    y_{\textrm{HF}}({x}, \theta) = 2y_{\textrm{LF}}(x, \theta) -20x + 20,
\label{e:yhf_eqn_deeponet}
\end{align}
where $\theta \in [10, 14]$ is the uncertain input parameter and $x \in [0, 1]$.

We train both surrogates for 10 independent replicates. In a given replicate, the BF-KLE and the BF-DeepONet are built using the same training data: $N_{\textrm{LF}} = 100$ pilot LF simulations generated via a Latin hypercube space-filling design, with a subset of $N_{\Delta} = 10$ paired HF-LF evaluations to model the discrepancy field. Similar to the numerical experiments reported in the paper, each KLE is truncated to retain 99\% of total variance, corresponding to approximately 2--3 modes. The associated PCE employs a total-order expansion of degree 6.

\begin{figure}[htbp]
    \centering
    \includegraphics[width=\linewidth]{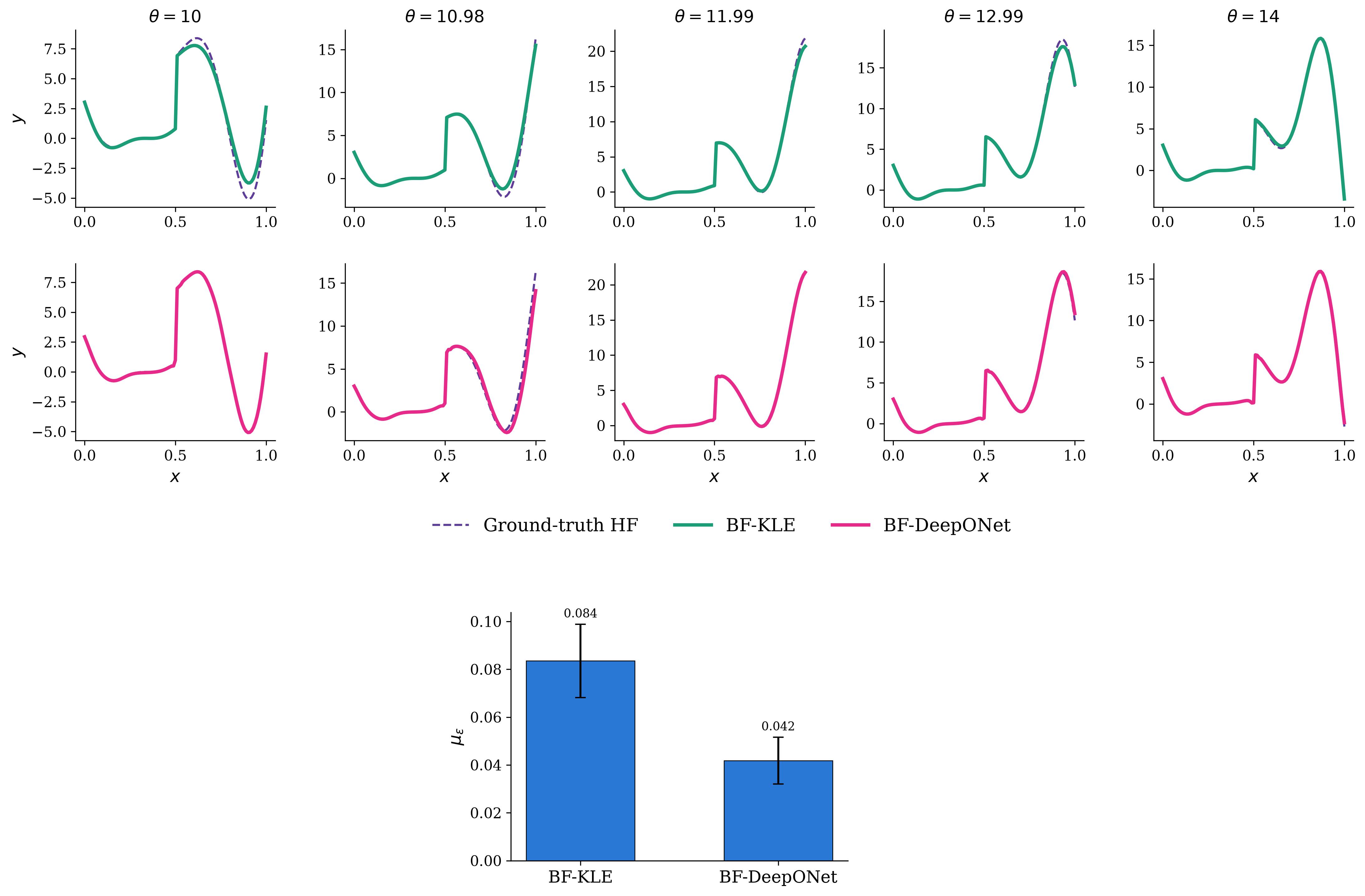}
    \caption{Comparison of bifidelity KLE and bifidelity DeepONet against ground-truth HF solutions at selected values of $\theta$ for a particular replicate (top), averaged results for integrated relative error over 10 replicates (bottom). Replicates differ in the space-filling design used to generate the training data for the surrogates.}
    \label{fig:test_predictions_deeponets_kle}
\end{figure}

Surrogate improvement via AL is not pursued in these experiments; the BF-KLE and BF-DeepONet surrogates are fitted to all available training data at once and compared to reference HF evaluations at held-out values of $\theta$. Selected predictions, as well as the integrated relative error $\mu_{\varepsilon}$ (Equation~\ref{eq:oracle_err_defn}) are visualized in Figure~\ref{fig:test_predictions_deeponets_kle}.

The BF-DeepONet has lower reported integrated relative error in this data-budget comparison and more closely follows the displayed HF curves. The result shows the value of a more flexible surrogate representation in this example, while a full computational comparison is dependent on training choices. A preliminary timing diagnostic for two replicates used 5000 DeepONet training iterations on an Intel Core i9-10920X system with 12 cores and 24 threads. It indicated substantially longer DeepONet fitting time than BF-KLE fitting for those runs. However, this observation is specific to the implementations, hardware, and training schedules used; it does not establish a general speedup or equal-accuracy cost advantage for BF-KLE. A reproducible cost-accuracy comparison also requires accounting for the DeepONet configuration, optimizer schedule and re-training strategy after AL acquisitions.

 \end{document}